\documentclass{article}

\usepackage[final, nonatbib]{neurips_2025}

\usepackage[utf8]{inputenc} % allow utf-8 input
\usepackage[T1]{fontenc}    % use 8-bit T1 fonts
\usepackage{hyperref}       % hyperlinks
\usepackage{url}            % simple URL typesetting
\usepackage{booktabs}       % professional-quality tables
\usepackage{amsfonts}       % blackboard math symbols
\usepackage{nicefrac}       % compact symbols for 1/2, etc.
\usepackage{microtype}      % microtypography
\usepackage{xcolor}         % colors

\title{NeSyPr: Neurosymbolic Proceduralization \\ For Efficient Embodied Reasoning}

\usepackage{kotex}
\usepackage{bm,bbm}
\usepackage{adjustbox}
\usepackage{multirow}
\usepackage{algorithm}

\usepackage{algpseudocode}

\usepackage{subfigure}
\usepackage{subcaption}
\usepackage{booktabs}
\usepackage{color}
\usepackage{enumerate}
\usepackage{enumitem}
\usepackage{amsmath}
\usepackage{amssymb}
\usepackage{mathtools}
\usepackage{amsthm}
\usepackage{wrapfig}
\usepackage{dsfont}
\usepackage{arydshln}
\usepackage{comment}
\usepackage[table]{xcolor}  % color for tables

\usepackage{tcolorbox}
\tcbuselibrary{most}
\usepackage{lipsum}
\newtcolorbox{question}{breakable,colframe=cyan,colback=cyan!20}
\usepackage{listings}

\usepackage[style=numeric, sorting=none]{biblatex}
\newcommand{\ourmodel}{\textsc{NeSyPr}}
\newcommand{\WMem}{\bm{M}}
\newcommand{\PMem}{\bm{R}}
\newcommand{\PMB}{\mathcal{M}}
\newcommand{\HS}{\bm{H}}
\newcommand{\Emb}{\bm{E}}
\newcommand{\Wei}{\bm{W}}

\newcommand{\StateSet}{\mathcal{S}}

\newcommand{\Goal}{\mathcal{G}}

\newcommand{\ActionSet}{\mathcal{A}}
\newcommand{\Dynamics}{\mathcal{P}}
\DeclareMathOperator*{\argmax}{argmax}
\DeclareMathOperator*{\argmin}{argmin}

\DeclareMathOperator*{\expectation}{\mathbb{E}}

\author{
    Wonje Choi, Jooyoung Kim, Honguk Woo\thanks{Honguk Woo is the corresponding author.} \\
    Department of Computer Science and Engineering, Sungkyunkwan University \\
    \texttt{\{wjchoi1995, onsamiro, hwoo\}@skku.edu}
}

\begin{document}

\maketitle

\begin{abstract}
We address the challenge of adopting language models (LMs) for embodied tasks in dynamic environments, where online access to large-scale inference engines or symbolic planners is constrained due to latency, connectivity, and resource limitations. 
To this end, we present $\ourmodel$, a novel embodied reasoning framework that compiles knowledge via neurosymbolic proceduralization, thereby equipping LM-based agents with structured, adaptive, and timely reasoning capabilities.
% 
% Drawing inspiration from a cognitive theory, our approach transforms declarative knowledge into procedural knowledge, thereby reducing reasoning load and enabling more efficient decision-making.
% 
%Our approach allows agents to internalize declarative knowledge from symbolic tools as parameterized procedural representations, seamlessly integrated into the LM’s inference process.
% In $\ourmodel$, declarative knowledge from symbolic tools is transformed into parameterized procedural representations, enabling seamless integration into the LM’s inference process. 
In $\ourmodel$, task-specific plans are first explicitly generated by a symbolic tool leveraging its declarative knowledge. These plans are then transformed into composable procedural representations  that encode the plans' implicit production rules, enabling the resulting composed procedures to be seamlessly integrated into the LM's inference process.
%
%During training, an agent learns to encode production rules into a vector-quantized procedural memory, where the resulting vectors are structured to be compositionally combinable for task-specific plan generation.
%
%At test time, without access to symbolic tools, the agent continues adaptive reasoning by generating plans in a procedural memory-augmented manner, while contrastively reconstructing their internal representations based on environmental feedback.
%
This neurosymbolic proceduralization abstracts and generalizes multi-step symbolic structured path-finding and reasoning into single-step LM inference, akin to human knowledge compilation. 
It supports efficient test-time inference without relying on external symbolic guidance, making it well suited for deployment in latency-sensitive and resource-constrained physical systems.
% 
%As a result, $\ourmodel$ enables LM-based agents to autonomously perform embodied task reasoning in dynamic environments.
% 
% 0508 성능이 강조되는 것과 함께, efficiency 가 강조되어야 함. 제안 구조의 특성이  단순히 성능을 올리는 것이 아니라, 적은 LM 도 complex reasoning 을 multi-turn inference 없이도 할 수 있게 하는 것. 
% 0509 wj: 넵. 수정했습니다.
% Across PDDLGym, ALFWorld, and VirtualHome, $\ourmodel$ achieves a 31.7\% higher task success rate than a large-scale baseline, while efficiently operating on a compact 0.5B-parameter LM. 
%
% It also exhibits 62.1\% greater robustness to noisy inputs compared to a symbolic planner, and outperforms existing memory-augmented LM methods by 13.0\%, demonstrating both efficiency and effectiveness in diverse embodied reasoning settings.
% 
% 0515 wj 간략화한 버전
We evaluate $\ourmodel$ on the embodied benchmarks PDDLGym, VirtualHome, and ALFWorld, demonstrating its efficient reasoning capabilities over large-scale reasoning models and a symbolic planner, while using more compact LMs.
% 원본
\begin{comment}
Across PDDLGym, ALFWorld, and VirtualHome benchmarks, $\ourmodel$ attains a 46.7\% higher task success rate than a large-scale reasoning model while using a 70 times smaller LM, and achieves 62.1\% greater robustness to dynamic conditions than a symbolic planner, evidencing strong structured and adaptive reasoning capabilities.
It also reduces inference latency by over 90.0\% compared to a multi-step reasoning baseline, while achieving a 36.0\% improvement in task success rate, thereby demonstrating timely and efficient reasoning capabilities. 
\end{comment}
\end{abstract}

\section{Introduction}\label{sec:introduction}
% Appl
Recent works such as Inner Monologue~\cite{llmagent:inner}, SayCan~\cite{llmagent:saycan}, and LLM‑Planner~\cite{llmagent:llmplanner} have demonstrated the potential of large-scale language models (LMs) to control embodied agents on complex tasks in dynamic environments.
Yet, the inherent limitations of autoregressive inference (e.g., shallow planning, inefficient context reuse, and lack of structure) have led researchers to explore more structured reasoning approaches.
%Nevertheless, the inherent limitations of autoregressive inference have led researchers to explore more structured reasoning approaches.
%   
% 0508 각 분야에 3-4 정도씩 citation 을 넣을 것. 
% 0509 wj: 넵. 하나씩 더 추가했습니다.
Three primary directions have emerged: (\romannumeral 1) agentic frameworks that support autonomous planning and multi‑step reasoning~\cite{llmagent:react, llmagent:reflexion, llmagent:agentcot, wu2025agentic}, (\romannumeral 2) neurosymbolic methods that integrate LMs with external symbolic reasoning frameworks~\cite{nesyagent:generalized, nesyagent:clmasp, nesyagent:llmp, nesyagent:llmrp}, and (\romannumeral 3) augmented LM approaches that incorporate memory modules~\cite{llmmem:longmem, llmmem:memlong, llmmem:rmt, llmmem:lmm}.
%These strategies aim to support more systematic decision-making, going beyond purely probabilistic inference.
% 
% 
Despite these advances, existing approaches still face significant limitations, particularly in resource-constrained dynamic environments. 
Agentic frameworks require iterative inference of large-scale models, leading to substantial computational overhead~\cite{llmagent:deder}.
Neurosymbolic methods, while using predefined rules to systematically search for accurate reasoning paths, often suffer from increased task-solving time and diminished effectiveness in dynamic settings unless the solver and its rule base are continuously updated to reflect environmental changes~\cite{nesyagent:nesyc}.
Memory-augmented LM approaches typically expand the context window to encode more information, yet they rarely retain the procedural structure needed for complex embodied tasks.

\begin{comment}
% Prob
Despite these advances, all three approaches continue to struggle with embodied task—particularly in ever-changing environments that demand rapid, automatic, and low-error decisions.
% agentic workflow 
(\romannumeral 1) Agentic frameworks~\cite{llmagent:react, llmagent:reflexion, llmagent:agentcot} often assume access to large-scale models with sufficient reasoning capabilities and rely on iterative inference, resulting in substantial computational overhead~\cite{llmagent:deder}.
% neurosymbolic
(\romannumeral 2) Neurosymbolic methods~\cite{nesyagent:generalized, nesyagent:clmasp, nesyagent:llmp} use predefined rules to systematically search for accurate reasoning paths, offloading much of the reasoning load from the neural model to symbolic tools.
However, as problem complexity increases, solving time grows considerably, and these methods struggle to remain effective in dynamic settings unless the solver and its rule base are continuously updated to reflect environmental changes~\cite{nesyagent:nesyc}.
% augmented lm
(\romannumeral 3) Memory-augmented LM approaches~\cite{llmmem:longmem, llmmem:rmt, llmmem:lmm} focus on expanding the context window to integrate large amounts of information efficiently, yet they rarely retain procedural information essential for complex embodied task reasoning.
\end{comment}

These limitations emphasize the need for a reasoning framework tailored to LM-based embodied agents, capable of structured and adaptive decision-making under time and resource constraints and without online symbolic assistance.
%
% Sol motivated by ACT
To this end, we draw inspiration from the Adaptive Control of Thought (ACT) theory~\cite{ACT}, which models skill acquisition as a process of knowledge compilation—the conversion of declarative knowledge into procedural form via repeated practice. 
Declarative knowledge is held in declarative memory as chunks that encode explicit facts such as propositions or problem states. 
By contrast, procedural knowledge consists of condition–action rules, stored in procedural memory and triggered automatically as cognitive procedures.
%Procedural knowledge resides in procedural memory as productions, namely condition–action rules that fire automatically when their conditions are satisfied. Ordered sequences of productions constitute cognitive procedures.
% 
A central mechanism in ACT, known as \textit{proceduralization}, hinges on the interplay of three memory systems.
% encodes factual task knowledge directly into implicit production rules through the interplay of three memory systems.
Declarative memory supplies the relevant chunk into working memory, which temporarily holds the current problem state during interaction with the environment. Through repeated use, the factual patterns are gradually compiled into production rules stored in procedural memory.
Once compiled, production rules fire whenever their conditions match the contents of working memory, enabling actions without further reference to declarative memory. By bypassing declarative retrieval, proceduralization reduces cognitive load and supports faster, more automatic, and less error-prone execution~\cite{oliveira2004conscious, anderson1982acquisition, tehranchi2021predicting, ritter2013declarative}.
%
% Compiling declarative knowledge into procedural form reduces reliance on declarative memory, lowers cognitive load, and enables more efficient reasoning.
% 
% \textit{proceduralization} removes the need to reference declarative memory during performance and thereby enabling faster, more automatic, and less error-prone execution~\cite{oliveira2004conscious, anderson1982acquisition, tehranchi2021predicting, ritter2013declarative}.

% in-depth
% In ACT, proceduralization arises from :
% \textit{declarative memory}, which stores explicit factual knowledge such as premises and rationales for goal-directed reasoning;
% \textit{working memory}, which ;
% \textit{procedural memory}, which encodes implicit procedures that are 
% 

% Sol Framework
Building on the ACT theory, we present $\ourmodel$, a novel embodied reasoning framework based on neurosymbolic proceduralization.
%efficient knowledge compilation. 
%
It performs knowledge compilation by abstracting and generalizing multi-step symbolic path-finding and reasoning into a single-step LM inference, thereby enabling LM-based agents to perform embodied reasoning efficiently, without relying on large-scale inference or online access to external symbolic tools.
% 
% 
% 
% \begin{comment}
% 0508 수정 필요. 문장 좀 이상함. 
% 0509 wj: 넵. 수정했습니다.
% 0508 procedural memory 가 hybrid memory ? 인가.. 구분해서 사용해야 함. framework 구현체와 ACT 용어가 혼돈이 있음. 
% 0509 wj: hybrid memory = working memory + procedural memory 구조입니다. 이는 ACT 와 동일합니다; working memory 는 declarative memory 와 procedural memory 가 동시에 접근 가능한 중간 interface 역할입니다. proceduralization을 통해 declarative memory 와의 연결이 필요 없어지면 working memory 와 procedural memory 로 만 동작가능하게 되기 때문에 agent 는 hybrid memory = working memory + procedural memory 구조를 하고 있습니다. 
% 0509 wj: 따라서 다음 문장에서는 이전 문장에 소개한 hybrid memory architecture 를 한 단계 내려서 working memory 와 procedural memory 로 구성 및 동작하는 것을 설명하고자 했습니다.
In $\ourmodel$, task-specific plans are first explicitly generated by a symbolic tool leveraging its declarative knowledge.
These plans are then transformed into composable procedural representations that encode the plans’ implicit production rules, enabling the resulting composed procedures to be seamlessly integrated into the LM’s inference process.
This proceduralization proceeds in two phases: \romannumeral 1) compositional NeSy procedure learning and \romannumeral 2) NeSy procedure contrastive planning.
%This proceduralization proceeds in two distinct phases: \romannumeral 1) compositional NeSy procedure learning and \romannumeral 2) NeSy procedure contrastive planning.
% 
During training phase \romannumeral 1), an agent learns to encode production rules into a vector-quantized procedural memory, in which the resulting vectors are structured to be composable for task-specific plan generation.
% compositionally combinable for task-specific plan generation.
% In $\ourmodel$, during training, environmental context (e.g., current state and goal) captured by the working memory is discretized via the logic codebook, restructured into task-relevant procedures, and retained in the vector quantization~\cite{vqvae} (VQ)-based procedural memory.
% \color{blue}
% Through compositional NeSy procedure learning, the logic codebook segments semantically rich contextual traces into compositions of atomic logic codes.
% \color{black}
% 0508 이런 nesy procedure learning 의 기술을 한줄 추가해서 주요 기술이 들어나게 할 것. 어떻게 composition 방법을 배우는지.. 
% 0508 그런데 위 내용이 training phase 인가요? 아니면 continual learning 처럼 inference 하면 하는 건가요? 두개가 분명이 차이가 있을턴데. 설명이 continual learning 인지 분리된 training phase 인지 헷깔리게 되어 있음. 
% 0509 wj: 기본 learning 방식은 continual learning 이 아닙니다. nesy procedure learning 으로 바꿔서 위에 내용 수정해 보았습니다.
% 
% 
% At test time, $\ourmodel$ employs procedural memory-augmented planning, specifically adopting 
% a contrastive decoding strategy~\cite{cd}, to adapt to a stream of unseen tasks over time.
% At test phase \romannumeral 2), without access to symbolic tools, the agent continues adaptive reasoning by generating plans in a procedural memory-augmented manner, while contrastively reconstructing their internal representations based on environmental feedback.
At test phase \romannumeral 2), without access to symbolic tools, the agent continues adaptive reasoning by generating plans augmented with procedural memory, while contrastively reconstructing their internal representations based on environmental feedback.

% 0508 아래 문장도 기술이 들어나게 완성해야 함. 
% \color{blue}
% The learned procedures, labeled as valid or invalid through environmental feedback, are leveraged to guide planning toward previously verified routines, thereby facilitating  adaptation.
% \color{black}
% 
%%%%%%%%%%%%%%%%%%%%%%%%%%%%%%%%%5
% \end{comment}

% Eval
We evaluate $\ourmodel$ on  PDDLGym~\cite{benchmark:pddlgym},  VirtualHome~\cite{benchmark:vh}, and ALFWorld~\cite{benchmark:alfw}, where inputs include observation, goal, and domain knowledge that are specified symbolically.
For structured reasoning, $\ourmodel$ achieves a 46.7\% higher task success rate than \textbf{DeepSeek-R1-Distill}~\cite{guo2025deepseek}, a distilled 70B-scale reasoning model, while operating with a 70 times smaller LM (as shown in Table~\ref{tab:ana:proc}).
For adaptive reasoning, it attains a 62.1\% higher success rate on unseen tasks with dynamic conditions than the \textbf{symbolic planner}~\cite{helmert2006fast} (in Table~\ref{fig:ana:er}).
For timely reasoning, it reduces inference latency by more than 90.0\% compared to \textbf{BoT}~\cite{proc:bot}, a large-scale inference baseline, while achieving a 36.0\% improvement in task success rate (in Table~\ref{tab:ana:proc}).
These results demonstrate that $\ourmodel$ endows LM-based agents with strong structured, adaptive, and timely reasoning capabilities.

Our contributions are summarized as: 
(1) We present $\ourmodel$, the neurosymbolic proceduralization-based reasoning framework inspired by ACT, which compiles multi-step symbolic reasoning into single-step LM inference, eliminating the need for online symbolic planners in embodied tasks.
(2) We develop compositional NeSy procedure learning, which encodes production rules into a vector-quantized procedural memory whose vectors can be compositionally combined to generate task-specific plans.
(3) We implement NeSy procedure contrastive planning, which adaptively generates plans by contrastively reconstructing task-specific procedures from stored procedures labeled with environmental feedback.
(4) We show the effectiveness and efficiency of $\ourmodel$ through extensive evaluations including 3 embodied benchmarks and 9 experimental scenarios, demonstrating its capabilities for structured, adaptive, and timely reasoning.

\section{Related Work}
% 
% \noindent\textbf{Related work}
% 원문
% In LM-based reasoning, agentic frameworks~\cite{llmagent:react,llmagent:reflexion,llmagent:agentcot,llmagent:lats,llmagent:mcts} support autonomous planning and multi-step reasoning by interleaving Chain-of-Thought (CoT)~\cite{cot:cot} with external feedback, but but rely on large inference engines that incur high computational cost~\cite{llmagent:deder}.
% 
% Neurosymbolic methods~\cite{nesyagent:generalized, nesyagent:clmasp, nesyagent:llmp, nesyagent:llmrp, nesyagent:ai2thor} integrate LMs with symbolic tools, offloading reasoning to rule-based engines but relying on handcrafted rules that scale poorly~\cite{nesyagent:nesyc}.
% 
% Memory-augmented LMs~\cite{llmmem:lmm, llmmem:longmem, llmmem:memlong, llmmem:trm} enhance context length by storing rich semantic information.
% 
% In contrast, $\ourmodel$ compiles symbolic knowledge into procedural form within an LM, removing reliance on large-scale inference engines or online access to external symbolic tools.
% 
% Additional related work is discussed in the Appendix.
% 
\noindent\textbf{Agentic Frameworks for Embodied Tasks.}
% common
A growing body of research has explored how LMs can be utilized to plan actions in physical or simulated environments~\cite{llmagent:inner, llmagent:saycan, llmagent:llmplanner, llmagent:wang2023describe, llmagent:zsp, llmagent:tapa, llmagent:progprompt}.
Recent studies emphasize agentic frameworks that enable autonomous planning and multi-step reasoning, rather than relying on single-step predictions.
Within this paradigm, methods such as ReAct~\cite{llmagent:react}, Reflexion~\cite{llmagent:reflexion}, and others~\cite{llmagent:agentcot, llmagent:lats, llmagent:mcts, wu2025agentic, jin2025search, zawalski2024robotic} integrate Chain-of-Thought (CoT)~\cite{cot:cot, cot:tot, cot:got, cot:got2} reasoning with environmental feedback.
Although these frameworks demonstrate strong performance, they typically rely on iterative inference of large-scale models, leading to substantial computational overhead.
In contrast, our approach employs an LM equipped with a specialized memory architecture, enabling robust reasoning without dependence on large-scale models or multi-step inference.

\noindent\textbf{Neurosymbolic Methods for Embodied Task Reasoning.}
% common
Neurosymbolic approaches integrate symbolic reasoning modules—such as rule-based or logic programming tools~\cite{lp:solver, asp:solver, sat:solver}—with neural networks to achieve interpretable and verifiable reasoning. With the advent of LMs, these hybrid systems have advanced logical reasoning in natural language tasks~\cite{nesy:linc, nesy:logiclm, nesy:reasoner, nesy:coupling, nesy:leveraging}.
This line of research has also extended into embodied tasks~\cite{nesyagent:generalized, nesyagent:clmasp, nesyagent:llmp, nesyagent:llmrp, nesyagent:ai2thor}, where delegating task reasoning to symbolic tools yielded more reliable and optimized action plans.
However, these methods depend on handcrafted domain knowledge (e.g., action rules) and require continuous updates to remain effective in dynamic environments~\cite{nesyagent:nesyc}.
Moreover, solving time increases sharply with task complexity, limiting real-time decision-making in complex settings.
In contrast, our approach embeds procedural knowledge within the LM’s memory architecture, enabling efficient reasoning and end-to-end adaptation through environmental feedback, without online symbolic assistance.

\noindent\textbf{Memory-augmented LMs for Long-term Generation.}
% common
Recent studies~\cite{llmmem:rmt, llmmem:lmm, llmmem:xl, llmmem:trime, llmmem:memreasoner, llmmem:armt} enhance LMs with external memory structures to better retain long-term context, often through recurrent memory updates within transformer architectures.
Other approaches store intermediate attention states—such as key-value pairs from relevant documents or histories—in external memory modules for retrieval during inference~\cite{llmmem:longmem, llmmem:memlong, llmmem:trm}.
While these methods expand the model’s context window to capture richer semantic information, only a few studies~\cite{memagent:optimus2, memagent:omnidrive, memagent:think} explore memory architectures specialized for embodied tasks.
In contrast, our approach introduces a procedural memory architecture that encodes task-level procedural knowledge, enabling efficient reasoning and adaptive behavior in dynamic embodied environments.

\section{Problem formulation}\label{sec:problem}
We consider embodied reasoning in dynamic settings, where an agent engages with a stream of tasks and must adapt to changing states and goals over time. Each task is defined as a tuple $\tau\!=\!(\StateSet,\!\ActionSet,\!\Dynamics,\!g)$, where $s\!\in\!\StateSet$ is the state, $a\!\in\!\ActionSet$ is the action, $\Dynamics: \StateSet \times \ActionSet \rightarrow \StateSet$ is the transition function describing dynamics, $g\!\in\!\Goal$ denotes the goal. 
Due to partial observability~\cite{sutton2018reinforcement}, the agent receives an observation $o_t$ at each timestep $t$.
%The transition function $\Dynamics: \StateSet \times \ActionSet \rightarrow \StateSet$ describes the dynamics, and a goal $g\!\in\!\Goal$ is sampled from the goal space.% to specify the task objective.
%
Unlike conventional multitask settings~\cite{llmagent:inner,llmagent:llmplanner}, the agent must solve a sequence of tasks $\mathcal{T}\!=\!\{\tau_1,\!\tau_2,\!\dots,\!\tau_N\}$, over time, where both $g$ and $\Dynamics$ may vary across tasks~\cite{abel2023definition, lee2024incremental}.
Our objective is to develop an LM-based agent (LM policy) that solves tasks autonomously and continuously at test time with no online access to any symbolic tools, while internalizing procedural knowledge from symbolic guidance during training. Note that Eq.~\eqref{eq:objective} defines the ideal objective, which is approximated in practice by supervising the LM on planner-computed action sequences with symbolic inputs.
%that internalizes procedural knowledge from symbolic guidance during training, and solves tasks autonomously without symbolic tools at test time.
% 
\begin{equation}\label{eq:objective}
    \pi_{\mathrm{LM}}^* = \argmax_{\pi_{\mathrm{LM}}} 
    {\textstyle\sum\limits_{i=1}^{N}}
    \expectation_{\tau_i} \left[ 
    {\textstyle\sum\limits_{t=0}^{T}}
    \mathrm{SR}(s_t, \pi_{\mathrm{LM}}(o_t, g)) - \mathrm{D}_{\mathrm{KL}}(\pi_{\mathrm{LM}}(\cdot \mid o_t, g) \parallel \pi_\mathrm{tool}(\cdot \mid o_t, g)) \right]
\end{equation}
Here, $\mathrm{SR} : \StateSet \times \ActionSet \rightarrow \{0,1\}$ indicates whether actions taken in current states $s_t$ lead to task success, and $\mathrm{D}_{\mathrm{KL}}$ measures the divergence between the LM-based policy 
$\pi_{\mathrm{LM}}$ and the symbolic policy $\pi_{\mathrm{tool}}$ derived from external tools in \cite{asp:solver, helmert2006fast, hoffmann2001ff}.
Accordingly, $\pi_{\mathrm{LM}}^*$ aims at maximizing task success while aligning its learned procedural behavior with the tool’s declarative guidance.

\begin{figure*}[t]
\begin{center}
\centerline{\includegraphics[width=0.99\linewidth]{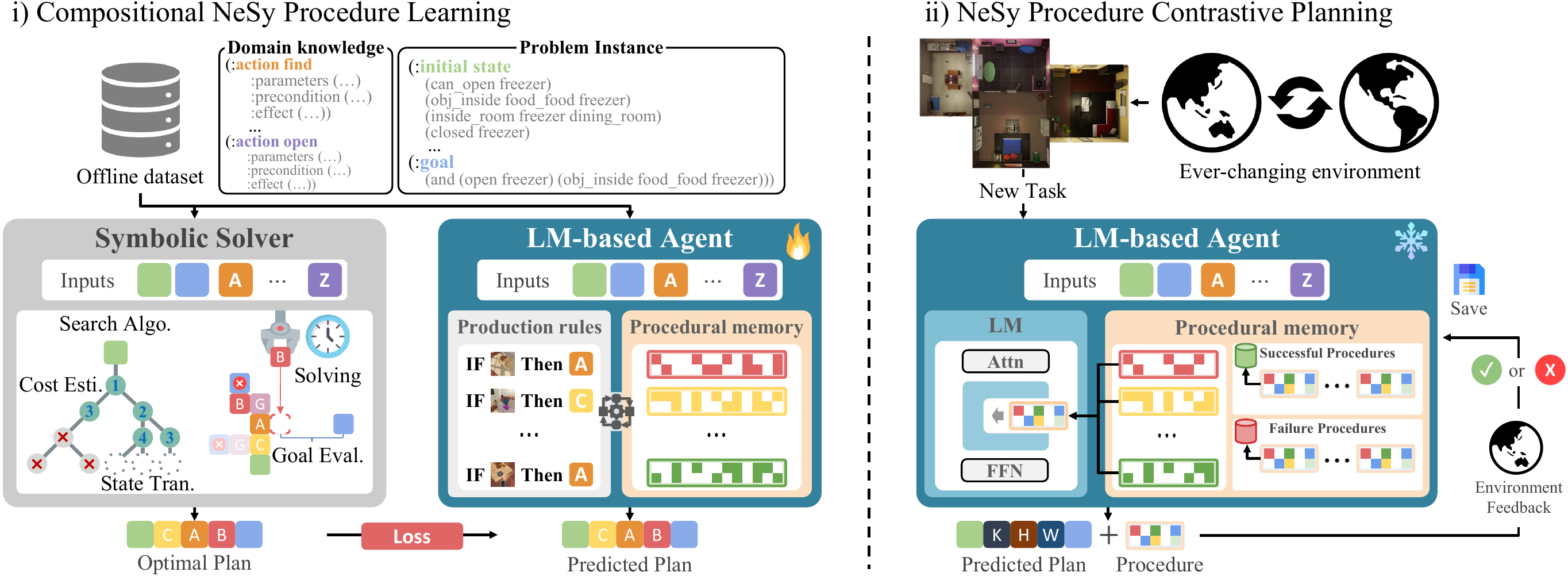}}
% \vspace{-3pt}
\caption{The framework architecture of $\ourmodel$}
\label{fig:1}
\end{center}
\vspace{-20pt}
\end{figure*}

\section{\ourmodel: Neurosymbolic Proceduralization}\label{sec:approach}
% 0507 wj: intro-prob-method 문제 언급과 왜 이게 최선인지로 설득시켜야
%Building on the objective in Eq.~\eqref{eq:objective}, we introduce $\ourmodel$, a memory-augmented neurosymbolic proceduralization framework for knowledge compilation.
% 
%Drawing on ACT~\cite{ACT}, $\ourmodel$ adapts the interplay among declarative, working, and procedural memory to convert declarative knowledge from symbolic tools into parameterized procedures embeddable in LMs, thereby equipping agents with structured and adaptive reasoning for complex embodied tasks in dynamic environments.

To equip agents with structured and adaptive reasoning for diverse embodied tasks, $\ourmodel$ learns to encode production rules and compose procedures, compressed representations derived from the declarative knowledge of symbolic tools. We refer to this end-to-end learning and utilization process as \textit{neurosymbolic proceduralization}. 
As illustrated in Figure~\ref{fig:1}, neurosymbolic proceduralization operates in two phases: \romannumeral 1) a training phase, \textit{compositional NeSy procedure learning}, where procedural knowledge is structured within procedural memory using plans generated by a symbolic tool, and \romannumeral 2) a test phase, \textit{NeSy procedure contrastive planning}, where the agent autonomously adapts to new tasks by contrastively reconstructing procedural representations, without relying on symbolic tools.

% offline data 를 사용해서 학습하는 거고, 학습데이터는 어떻게 구성되며, 각 memory 는 어떻게 구현되어 있는지
During phase~\romannumeral 1), the agent trains on offline data comprising symbolically defined problem instances (observations and goals) and associated domain knowledge (action rules).
%
% 0512 무엇이 procedural memory 에 저장되는지 언급되어야 단계 2와 연결이 가능함. 아래 수정 필요. 
% 0513 wj: 넵 수정했습니다.
The declarative knowledge used by symbolic tools for problem-solving, such as search algorithms, state transitions, cost estimation, and goal evaluation, is internalized as production rules. 
The agent composes these rules into task-solving procedures in procedural memory, which it then exploits to generate plans.
In phase~\romannumeral 2), using the procedural memory established during phase~\romannumeral 1) training, the agent performs structured reasoning without access to external symbolic tools (i.e., declarative memory). 
It further engages in adaptive reasoning by contrastively reconstructing procedures from prior ones labeled as successes or failures via environmental feedback. The agent continually reinforces plans aligned with valid procedures while suppressing those associated with invalid ones.
% 
% we model declarative memory as an external symbolic tool and equip the LM-based agent with a hybrid memory architecture that integrates working and procedural memory.
% Working memory retains the current observation and goal, while a VQ~\cite{vqvae}-driven logic codebook abstracts them into parameterized procedures that populate procedural memory.
% Working memory captures the current observation and goal, which are restructured—via a logic codebook that applies VQ~\cite{vqvae}—into parameterized procedures within procedural memory.
% 
% During \romannumeral 2), without symbolic tools, we apply a memory-based contrastive decoding~\cite{cd} strategy and leverage environmental feedback to guide predictions toward previously validated procedures.
% 
Accordingly, $\ourmodel$ enables LM-based agents to reason robustly across tasks and adapt efficiently to ever-changing environments.

\subsection{Compositional NeSy Procedure Learning}\label{subsec:memory}
\begin{figure*}[t]
\begin{center}
\centerline{\includegraphics[width=0.99\linewidth]{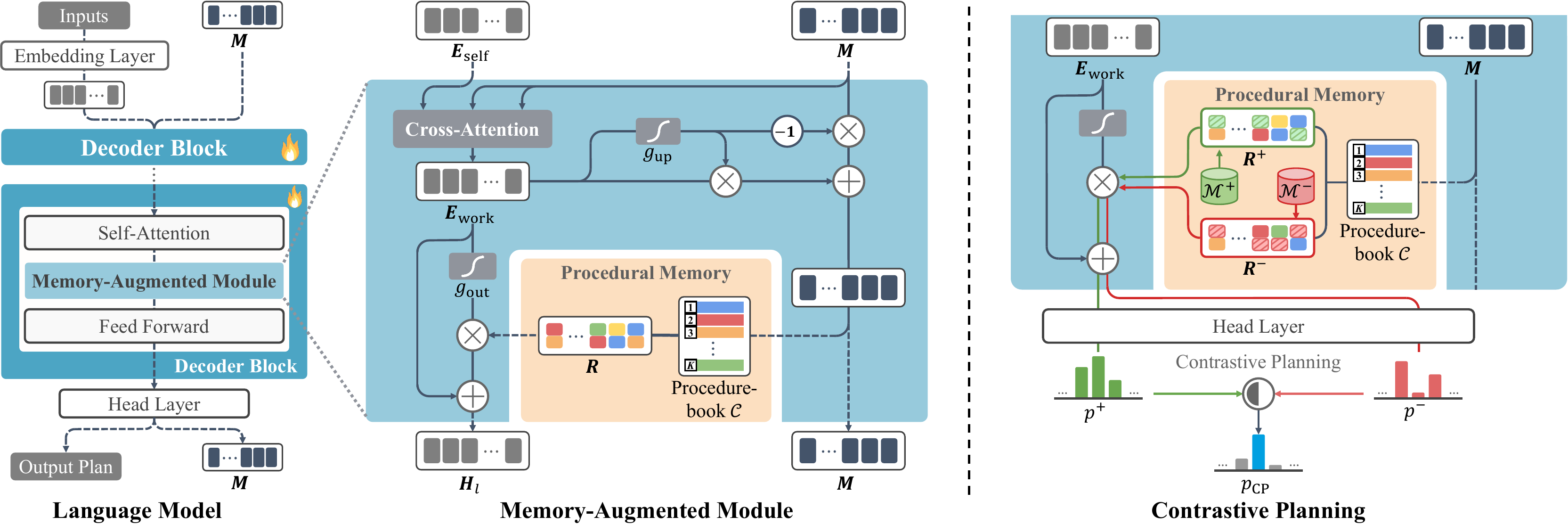}}
% \vspace{-4pt}
\caption{Vector-quantized procedural memory and contrastive planning}
\label{fig:2}
\end{center}
\vspace{-20pt}
\end{figure*}

As shown in Figure~\ref{fig:2}, our learning method incorporates a procedural memory that extends existing approaches~\cite{llmmem:lmm, llmmem:rmt} across $L$ layers to support structured reasoning.
Working memory $\WMem$ extends the context window by accumulating symbolic inputs from the environment.
The procedural memory then performs vector quantization (VQ)~\cite{vqvae}, encoding production rules into discrete procedure-units stored in a procedure-book $\mathcal{C}$, which are composed to generate plans.
% 
% For each layer $l\!\in\!\{1,\dots,L\}$, the hidden state $\HS_{l-1}$ from the previous layer is first augmented with $\WMem$ and then refined using $\PMem$ for the current task $\tau$.
% 
\begin{equation}\label{eq:decoder}
    \HS_l, \WMem = \mathrm{DecoderBlock}_l \bigl( \HS_{l-1}, \WMem \bigr), \quad \WMem \triangleq [\bm{e}_1, \bm{e}_2, \dots, \bm{e}_S], \ \bm{e}_i \in \mathbb{R}^D
\end{equation}
At each layer $l\!\in\!\{1,\dots,L\}$, the decoder block $\mathrm{DecoderBlock}_l$ takes the previous hidden state $\HS_{l-1}$ and $\WMem$ as input. 
Each slot $\bm{e}_i$ encodes environmental context in dimension $D$,
% , which typically matches the model’s hidden size
with $S$ defining memory capacity.
Runtime procedure $\PMem$ integrated into $\mathrm{DecoderBlock}_l$ contributes to refining $\HS_l$.
% 
% Within AMQ, $\PMem$ is derived from symbolic restructuring of $\WMem$ using VQ.
% 
% This abstract representation is 

\noindent\textbf{Memory-augmented module.}
$\WMem$ encodes the current environmental state, using a memory-augmented cross-attention adapted from \cite{llmmem:lmm}.
To enable information exchange between the self-attention output $\Emb_\mathrm{self}\!\in\!\mathbb{R}^{T \times D}$ from $\HS_{l-1}$ and $\WMem$, we apply a cross-attention. %, where $\Emb_\mathrm{self}$ serves as the query and $\WMem$ as both the key and value.
\begin{equation}\label{eq:cross_attn} 
    {\textstyle\Emb_\mathrm{work} = \mathrm{softmax}
    \left( \frac{QK^\top}{\sqrt{D}} \right)V, \quad 
    Q = \Emb_\mathrm{self} \Wei_Q,
    K = \WMem \Wei_K, 
    V = \WMem \Wei_V}
\end{equation}
Here, $\Wei_Q, \Wei_K, \Wei_V\!\in\!\mathbb{R}^{D\!\times\!D}$ are learnable projection matrices.
$\WMem$ is then updated via a gating mechanism that merges the original memory with the cross-attended representation $\Emb_\mathrm{work}$.
\begin{equation}\label{eq:gating}
    \WMem \leftarrow g_\mathrm{up} \odot \alpha(\Emb_\mathrm{work}) + (1 - g_\mathrm{up}) \odot \WMem, \quad 
    g_\mathrm{up} = \sigma\left( \alpha(\Emb_\mathrm{work}) \Wei_\mathrm{up} \right)
\end{equation}
Here, $\alpha$ is alignment operator ensuring dimensional consistency, $\sigma$ denotes the sigmoid activation function, $\odot$ represents slot-wise multiplication, and $\Wei_\mathrm{up}\!\in\!\mathbb{R}^{D\!\times\!D}$ is a learnable projection matrix.

In \textit{procedural memory}, $\PMem$ is obtained by applying VQ to $\WMem$ using a procedure-book $\mathcal{C}\!=\!\{c_1,\!c_2,\!\ldots,\!c_K\}$, where each procedure-unit $c_j \in \mathbb{R}^d$ is a $d$-dimensional vector.
Each slot $\bm{e}_i \in \WMem$ is partitioned into contiguous $d$-dimensional chunks to align with the procedure-units.
\begin{equation}\label{eq:chunk}
    \bm{e}_i = \bigl[ e_{i}^{(1)}; e_{i}^{(2)}; \dots ; e_{i}^{(q)} \bigr], \quad \ q = \lfloor D/d \rfloor, \ e_{i}^{(r)} \in \mathbb{R}^d
\end{equation}
Each $e_i^{(r)}$ is replaced with its nearest procedure-units $c_{k_r}\!\in\!\mathcal{C}$, selected by minimizing the Euclidean distance.
\begin{equation}\label{eq:vq}
    \bm{c}_i = \bigl[ c_{k_1}; c_{k_2}; \ldots; c_{k_q} \bigr], \quad c_{k_r} = \argmin_{c_j \in \mathcal{C}} \| e_{i}^{(r)} - c_j \|_2
\end{equation}
Concatenating the selected procedure-units forms a composite procedure $\bm{c}_i$. Aggregating all such procedures across memory slots yields runtime procedure: $\PMem\!=\![\bm{c}_1,\!\bm{c}_2,\!\ldots,\!\bm{c}_S]\!\in\!\mathbb{R}^{S\!\times\!D}$.
To integrate $\PMem$ into the model reasoning process, we combine $\PMem$ with $\Emb_\mathrm{work}$, enhancing $\HS_l$.
\begin{equation}\label{eq:integration}
    \HS_l = \mathrm{FFN}\left( \Emb_\mathrm{work} + g_\mathrm{out} \odot \alpha(\PMem) \right), \quad 
    g_\mathrm{out} = \sigma\left( \Emb_\mathrm{work} \Wei_\mathrm{out} \right), \Wei_\mathrm{out} \in \mathbb{R}^{D \times D}
\end{equation}
Here, $\mathrm{FFN}$ is a feed-forward submodule, $\Wei_{\mathrm{out}}$ is a learnable matrix, and $g_\mathrm{out}$ is a gating matrix.

\noindent\textbf{Learning objective.}
To train the procedure-book $\mathcal{C}$ end-to-end for compositional procedures, we combine the task objective (e.g., LM fine-tuning) with a VQ loss applied at each layer $l$.
\begin{equation}\label{eq:vq_loss}
    \mathcal{L}_\mathrm{VQ}^{(l)} = 
    \| \mathrm{sg}(\WMem^{(l)}) - \PMem^{(l)} \|_F^2 
    + \beta \| \WMem^{(l)} - \mathrm{sg}(\PMem^{(l)}) \|_F^2
\end{equation}
Here, $\mathrm{sg}$ is the stop-gradient operator, $\|\!\cdot\!\|_F$ is the Frobenius norm~\cite{skean2023frossl}, and $\beta$ is a weighting coefficient that controls $\WMem$ to align with $\PMem$. The overall training objective is defined as
\begin{equation}\label{eq:total_loss}
    \mathcal{L} = -\expectation_{\mathcal{T}} \left[\log \pi_{\theta}(a \mid o,g,\WMem)\right] + \lambda {\textstyle\sum\limits_{l=1}^L} \mathcal{L}_\mathrm{VQ}^{(l)}
\end{equation}
where $\theta$ denotes the learnable parameters of the LM, and $\lambda$ balances the task-specific objective and the VQ term. 
$\mathcal{C}$ is updated using an exponential moving average (EMA)~\cite{ema}, providing compositional procedural representations to $\PMem$ for use in structured reasoning.
% allowing $\PMem$ to be captured composable procedural representations for structured reasoning.

\subsection{NeSy Procedure Contrastive Planning}\label{subsec:decoding}
%
% 0512 이 서브섹션 설명 순서에 맞추어 아래 문장을 약간 수정하 필요가 있음.
% 0513 wj: 넵 수정했습니다.
% 0514 wj: 첫번째 코멘트와 유사.
To support adaptive reasoning at test time without symbolic tools, we introduce a procedure-based contrastive planning strategy that reconstructs composed procedures from feedback-labeled stored procedures and contrastively generates plans aligned with dynamic environments.
% To support adaptive reasoning at test time without symbolic tools, we implement a contrastive planning mechanism that incorporates environmental feedback and reconstructs procedure representations. %It enables the agent to reinforce procedural memory by contrasting successful and failed executions.

\noindent\textbf{Procedure reconstruction.}
We maintain two procedure banks: $\PMB^{+}$ for successful procedures and $\PMB^{-}$ for failures.  
For each $\bm{c}_i\!\in\!\PMem$, before integration in Eq.~\eqref{eq:integration}, we reconstruct two versions: a positive procedure $\bm{c}_i^{+}$ by matching against $\PMB^{+}$, and a negative procedure $\bm{c}_i^{-}$ against $\PMB^{-}$.
\begin{equation}\label{eq:procedure-validation}
    \bm{c}^{+}_i\!\leftarrow\!
    \begin{cases}
        \bm{c}^{+}\! & \!\text{if } \exists  \bm{c}^{+}\!\in\!\PMB^{+}\!\land\!\mathrm{sim}(\bm{c}_i, \bm{c}^{+})\!\ge\!\upsilon \\
        \bm{c}_i\! & \!\text{otherwise}
    \end{cases}
    ,
    \bm{c}^{-}_i\!\leftarrow\!
    \begin{cases}
        \bm{c}^{-}\! & \!\text{if } \exists  \bm{c}^{-}\!\in\!\PMB^{-}\!\land\!\mathrm{sim}(\bm{c}_i, \bm{c}^{-})\!\ge\!\upsilon \\
        \bm{c}_i\! & \!\text{otherwise}
    \end{cases}
\end{equation}
Here, $\mathrm{sim}$ denotes a similarity function (e.g., cosine similarity), and $\upsilon$ is a reconstruction threshold.
We then reconstruct two runtime procedures: $\PMem^{+}\!=\![\bm{c}^{+}_1,\!\bm{c}^{+}_2,\!\ldots,\!\bm{c}^{+}_S]$ and $\PMem^{-}\!=\![\bm{c}^{-}_1,\!\bm{c}^{-}_2,\!\ldots,\!\bm{c}^{-}_S]$.
These are used to generate two versions of the hidden state $\HS_l$ within a single batch: one conditioned on $\PMem^{+}$ and the other on $\PMem^{-}$.
Both are then used in contrastive decoding~\cite{cd}.
%
% In addition to matching against $\PMB^{+}$, we compare $\bm{c}_i$ to failure patterns in $\PMB^{-}$; otherwise, $\bm{c}_i$ remains unchanged.
% If sufficiently similar failure procedures are detected, we apply contrastive decoding~\cite{cd} by computing two versions of $\HS_l$ within a single batch: one using $\PMB^{+}$ and the other using $\PMB^{-}$.

\noindent\textbf{Contrastive planning.}
Following~\cite{acd}, we guide generation toward successful procedures by computing a contrastive score for each token $x_i$ within an adaptive plausibility set $\mathcal{V}_\mathrm{head}(x_{<i})$,
\begin{equation}
    \mathcal{V}_\mathrm{head}(x_{<i}) = \{ x_i\!\in\!\mathcal{V}\!\mid\!p^{+}(x_i \mid x_{<i}, \WMem; \PMB^{+}) \geq \vartheta \max_{x'} p^{+}(x'\!\mid\!x_{<i}, \WMem; \PMB^{+}) \}
\end{equation}
where $\vartheta\!=\!0.1$ controls the truncation threshold.
For $x_i\!\in\!\mathcal{V}_\mathrm{head}(x_{<i})$, we obtain contrastive score by 
\begin{equation}\label{eq:contrastive_score}
    % S(x_i) = \log \frac{p^{+}(x_i \mid x_{<i}, \WMem; \PMB^{+})}{p^{-}(x_i \mid x_{<i}, \WMem; \PMB^{-})}
    S(x_i) = \log p^{+}(x_i \mid x_{<i}, \WMem; \PMB^{+}) - \log p^{-}(x_i \mid x_{<i}, \WMem; \PMB^{-})
\end{equation}
where $p^{+}$ and $p^{-}$ denote token distributions conditioned on successful and failed procedures, respectively.
The final next-token distribution is then reshaped as follows.
\begin{equation}
    p_\mathrm{CP}(x_i \mid x_{<i}) =
    \begin{cases}
        \mathrm{softmax}\left(S(x_i)\right) \cdot \sum_{x' \in \mathcal{V}_\mathrm{head}} p^{+}(x' \mid x_{<i}) & \text{if } x_i \in \mathcal{V}_\mathrm{head}(x_{<i}) \\
        p^{+}(x_i \mid x_{<i}) & \text{otherwise}
    \end{cases}
\end{equation}
This decoding process suppresses failure patterns and promotes plans aligned with the environment, enabling adaptive reasoning without symbolic guidance. Algorithm of $\ourmodel$ is in Appendix.

% This decoding suppresses failure patterns and promotes planning aligned with prior successes.

% The neurosymbolic proceduralization algorithm of our framework is detailed in Appendix.

\section{Evaluation}\label{sec:evaluation}
\subsection{Experiment Setting}
\noindent\textbf{Environments.} We evaluate $\ourmodel$ across diverse embodied benchmarks, including multiple domains from PDDLGym~\cite{benchmark:pddlgym} (e.g., Minecraft, Rearrangement, GlibRearrangement), VirtualHome~\cite{li2024embodied}, and ALFWorld~\cite{benchmark:alfw}.
To assess embodied reasoning performance in dynamic environments, as described in Section~\ref{sec:problem}, we configure each benchmark to support task sequences that require multi-step planning and continual adaptation.
% PDDLGym
In PDDLGym, where symbolic planners are built-in~\cite{helmert2006fast, hoffmann2001ff}, observations are provided in symbolic form. We construct 9 distinct task sequences by randomly composing tasks, ensuring consistent evaluation settings across all baselines.
% ALFWorld, VirtualHome
For VirtualHome and ALFWorld, we utilize symbolic observation interfaces provided by their respective open-source implementation. 
We evaluate under continual task settings with behavior-incremental and environment-incremental configurations, using 4 distinct task sequences for each, following \cite{clalfred}.
During the evaluation, agents receive only binary feedback at the task level (success or failure), and no gradient updates are allowed.

\noindent\textbf{Datasets.}
For training, we use a small set of problem instances paired with plans generated by symbolic planners~\cite{helmert2006fast}.
% PDDLGym
In PDDLGym, the train sets include 29 instances for Minecraft, 20 for Rearrangement, and 40 for GlibRearrangement.
% The corresponding test sets contain 389, 400, and 80 instances, respectively, all disjoint from the train data.
The test sets contain 389, 400, and 80 instances respectively, all disjoint from the train data.
% ALFWorld, VirtualHome
For VirtualHome and ALFWorld, the train sets consist of 77 and 549 instances, respectively.  
Each test set is split into seen and unseen sets. 
% The seen subset contains 1,509 tasks and 112 tasks respectively, and shares the same goal as the train set, but varies in object placement and inter-object relations.
The seen set contains 112 and 1,509 instances respectively, and shares the same goal as the train set, but varies in object placement and inter-object relations.
% The unseen subset contains 1,369 tasks and 52 tasks respectively, and introduces entirely new tasks,
The unseen set contains 52 and 1,369 instances respectively, and introduces entirely new tasks,
% 0513 either 가 아니라 both 일것 같은데, 확인필요
% 0513 wj: 넵 맞습니다.
not present in both the train and the seen sets.
% not observed in both the train and seen subsets.
%not observed in either the train or seen sets.
% 
At test time, agents are given only the current observation and goal, with no access to symbolic tools.
% At test time, agents are provided only with the current observation and goal, without access to symbolic tools.

\noindent\textbf{Baselines.} For comparison, we organize the baselines into four categories:
% naive llm based approaches
(\romannumeral 1) \text{Single-step planning}, including ZSP~\cite{llmagent:zsp}, RAP~\cite{llmagent:rap}, and LLM-Planner~\cite{llmagent:llmplanner}, generates action plans in an inference step.
% agentic workflow
(\romannumeral 2) \text{Agentic workflow}, such as CoT~\cite{cot:cot}, ToT~\cite{cot:tot}, GoT~\cite{cot:got}, ReAct~\cite{llmagent:react}, and Reflexion~\cite{llmagent:reflexion}, performs multi-step reasoning through multiple LM calls.
% augmented lm
(\romannumeral 3) \text{Memory-augmented LM}, including LongMem~\cite{llmmem:longmem} and LM2~\cite{llmmem:lmm}, incorporates long-term memory into the LM attention mechanism. Optimus-2~\cite{memagent:optimus2} and DT-Mem~\cite{memagent:think} extend this approach for embodied agents. We also include BUTLER~\cite{butler}, a parameter-efficient fine-tuning method.
% proceduralization
(\romannumeral 4) \text{Proceduralization} such as 
BoT~\cite{proc:bot} stores abstract reasoning templates in a meta-buffer and dynamically instantiates them to guide procedural LM reasoning. Furthermore, 
Large Reasoning Model (LRM)~\cite{proc:lrm} integrates CoT-style reasoning through reinforcement learning, with compact variants distilled from larger models.
PlaSma~\cite{proc:plasma} distills procedural knowledge from larger LMs into compact models.
By default, we use LLaMA-3.2-1B~\cite{llama3.2} for PDDLGym, and Qwen2.5-0.5B~\cite{qwen2.5} for VirtualHome and ALFWorld.

% \begin{itemize}[leftmargin=1em, topsep=0pt] % noitemsep
% \item \text{Single-step planning}, including ZSP~\cite{llmagent:zsp}, RAP~\cite{llmagent:rap}, and LLM-Planner~\cite{llmagent:llmplanner}, generates action plans in an inference step.
% \item \text{Agentic workflow}, such as CoT~\cite{cot:cot}, ToT~\cite{cot:tot}, GoT~\cite{cot:got}, ReAct~\cite{llmagent:react}, and Reflexion~\cite{llmagent:reflexion}, performs multi-step reasoning through multiple LM calls.
% \item \text{Memory-augmented LM}, including LongMem~\cite{llmmem:longmem} and LM2~\cite{llmmem:lmm}, incorporates long-term memory into the LM attention mechanism. Optimus-2~\cite{memagent:optimus2} and DT-Mem~\cite{memagent:think} extend this approach for embodied agents. We also include BUTLER~\cite{butler}, a parameter-efficient fine-tuning method.
% \item \text{Proceduralization} such as 
% BoT~\cite{proc:bot} stores abstract reasoning templates in a meta-buffer and dynamically instantiates them to guide procedural LM reasoning. Furthermore, 
% Large Reasoning Model (LRM)~\cite{proc:lrm} integrates CoT-style reasoning through reinforcement learning, with compact variants distilled from larger models.
% PlaSma~\cite{proc:plasma} distills procedural knowledge from larger LMs into compact models.
% \end{itemize}
% % 
% By default, we use LLaMA-3.2-1B~\cite{llama3.2} for PDDLGym, and Qwen2.5-0.5B~\cite{qwen2.5} for VirtualHome and ALFWorld.

\noindent\textbf{Metrics.} We use four standard metrics, following~\cite{llmagent:zsp, ALFRED20, anderson2018evaluation}.
Cumulative Task Success Rate (CSR) measures the percentage of tasks where all sub-goals are achieved.  
Cumulative Goal-Conditioned Success Rate (CGC) reports the fraction of individual sub-goals achieved across all tasks.
Executability (Exe) assesses if each selected action is feasible.
Success rate weighted by Path Length (SPL) reflects both task success and path efficiency.

Further details of the experimental settings are provided in the Appendix.

\begin{table*}[th]
\caption{
Performance on open-loop continual embodied tasks in \text{PDDLGym}. Metrics are averaged over 9 random seeds, with standard deviations to indicate consistency across runs. \textsc{Params} denotes the total number of model parameters, with the ratio of trainable parameters in parentheses.
}
\label{tab:main:open}
\vspace{-5pt}
\begin{center}
\begin{small}
\begin{sc}
\begin{adjustbox}{width=0.99\linewidth}
\begin{tabular}{l c cccc cccc}
    \toprule
    \multirow{2}{*}{Method}
    & \multirow{2}{*}{Params}
    & \multicolumn{4}{c}{Train} & \multicolumn{4}{c}{Test} \\ %& \multicolumn{3}{c}{Glibrearrangement}  \\ 
    \cmidrule(rl){3-6} \cmidrule(rl){7-10} %\cmidrule(rl){9-11}
    & & CSR ($\uparrow$) & CGC ($\uparrow$) & Exe ($\uparrow$) & SPL ($\uparrow$) & CSR ($\uparrow$) & CGC ($\uparrow$) & Exe ($\uparrow$) & SPL ($\uparrow$) \\
    \midrule
    \multicolumn{7}{l}{\textbf{Domain}: Minecraft} & \\
    \midrule
    \rowcolor[HTML]{FFFFFF} ZSP & 1.7B (0.0\%)
    & 76.4$\pm$5.5 & 81.7$\pm$4.8 & 100.0$\pm$0.0 & 0.8$\pm$0.1
    & 16.4$\pm$1.5 & 35.9$\pm$2.0 & 100.0$\pm$0.1 & 0.1$\pm$0.0 \\
    
    \rowcolor[HTML]{FFFFFF} RAP & 1.7B (0.0\%)
    & 76.4$\pm$5.5 & 81.7$\pm$4.8 & 100.0$\pm$0.0 & 0.8$\pm$0.1
    & 16.8$\pm$0.9 & 18.0$\pm$1.9 & 100.0$\pm$0.1 & 0.1$\pm$0.0 \\
    
    % \midrule
    \rowcolor[HTML]{E9ECEF} CoT & 1.7B (0.0\%)
    & 83.3$\pm$6.7 & 83.3$\pm$9.3 & 100.0$\pm$0.0 & 0.8$\pm$0.1
    & 17.0$\pm$0.5 & 25.7$\pm$1.8 & 100.0$\pm$0.0 & 0.1$\pm$0.0 \\

    \rowcolor[HTML]{E9ECEF} ToT & 1.7B (0.0\%)
    & 85.1$\pm$4.2 & 85.7$\pm$3.7 & 100.0$\pm$0.0 & 0.9$\pm$0.0
    & 18.7$\pm$1.0 & 32.2$\pm$2.6 & 100.0$\pm$0.0 & 0.1$\pm$0.0 \\

    \rowcolor[HTML]{E9ECEF} GoT & 1.7B (0.0\%)
    & 89.1$\pm$5.1 & 94.6$\pm$7.7 & 100.0$\pm$0.0 & 0.9$\pm$0.1
    & 18.9$\pm$0.5 & 25.9$\pm$1.2 & 100.0$\pm$0.0 & 0.1$\pm$0.0 \\
    
    % \midrule
    \rowcolor[HTML]{FFFFFF} BUTLER & 1.2B (0.6\%)
    & 100.0$\pm$0.0 & 100.0$\pm$0.0 & 100.0$\pm$0.0 & 1.0$\pm$0.0
    & 51.4$\pm$1.9 & 56.7$\pm$2.6 & 99.7$\pm$0.3 & 0.4$\pm$0.0 \\

    \rowcolor[HTML]{FFFFFF} LongMem & 1.6B (24.6\%)
    & 100.0$\pm$0.0 & 100.0$\pm$0.0 & 100.0$\pm$0.0 & 1.0$\pm$0.0
    & 53.3$\pm$3.7 & 56.3$\pm$4.1 & 99.8$\pm$0.1 & 0.5$\pm$0.0 \\

    \rowcolor[HTML]{FFFFFF} LM2 & 1.3B (6.3\%)
    & 100.0$\pm$0.0 & 100.0$\pm$0.0 & 100.0$\pm$0.0 & 1.0$\pm$0.0
    & 47.9$\pm$8.4 & 56.5$\pm$4.6 & 99.9$\pm$0.1 & 0.4$\pm$0.0 \\

    \rowcolor[HTML]{E9ECEF} $\ourmodel$ & 1.3B (6.3\%)
    & 100.0$\pm$0.0
    & 100.0$\pm$0.0
    & 100.0$\pm$0.0
    & 1.0$\pm$0.0
    & \textbf{65.2}$\pm$1.4
    & \textbf{68.9}$\pm$2.4
    & 100.0$\pm$0.1
    & \textbf{0.6}$\pm$ 0.0\\
    \midrule
    \multicolumn{7}{l}{\textbf{Domain}: Rearrangement} & \\
    \midrule
    \rowcolor[HTML]{FFFFFF} ZSP & 1.7B (0.0\%)
    & 94.2$\pm$3.8 & 97.6$\pm$1.2 & 100.0$\pm$0.0 & 0.9$\pm$0.0
    & 15.3$\pm$2.1 & 22.4$\pm$3.4 & 100.0$\pm$0.0 & 0.1$\pm$0.0 \\
    
    \rowcolor[HTML]{FFFFFF} RAP & 1.7B (0.0\%)
    & 94.2$\pm$3.8 & 97.6$\pm$1.2 & 100.0$\pm$0.0 & 0.9$\pm$0.0
    & 18.3$\pm$1.7 & 29.0$\pm$2.6 & 100.0$\pm$0.0 & 0.1$\pm$0.0 \\
    
    % \midrule
    \rowcolor[HTML]{E9ECEF} CoT & 1.7B (0.0\%)
    & 95.0$\pm$4.5 & 98.5$\pm$2.5 & 100.0$\pm$0.0 & 1.0$\pm$0.0
    & 19.2$\pm$0.9 & 29.9$\pm$1.5 & 100.0$\pm$0.0 & 0.1$\pm$0.0 \\

    \rowcolor[HTML]{E9ECEF} ToT & 1.7B (0.0\%)
    & 95.0$\pm$6.3 & 99.5$\pm$1.2 & 100.0$\pm$0.0 & 1.0$\pm$0.1
    & 20.2$\pm$1.2 & 33.9$\pm$1.4 & 100.0$\pm$0.0 & 0.2$\pm$0.0 \\

    \rowcolor[HTML]{E9ECEF} GoT & 1.7B (0.0\%)
    & 95.8$\pm$4.9 & 100.0$\pm$0.0 & 100.0$\pm$0.0 & 1.0$\pm$0.0
    & 23.0$\pm$1.5 & 37.1$\pm$1.4 & 100.0$\pm$0.0 & 0.2$\pm$0.0 \\
    
    % \midrule
    \rowcolor[HTML]{FFFFFF} BUTLER & 1.2B (0.6\%)
    & 100.0$\pm$0.0 & 100.0$\pm$0.0 & 100.0$\pm$0.0 & 1.0$\pm$0.0
    & 56.5$\pm$2.0 & 69.8$\pm$2.9 & 100.0$\pm$0.0 & 0.5$\pm$0.0 \\

    \rowcolor[HTML]{FFFFFF} LongMem & 1.6B (24.6\%)
    & 100.0$\pm$0.0 & 100.0$\pm$0.0 & 100.0$\pm$0.0 & 1.0$\pm$0.0
    & 58.2$\pm$1.9 & 69.9$\pm$1.4 & 100.0$\pm$0.0 & 0.5$\pm$0.0 \\

    \rowcolor[HTML]{FFFFFF} LM2 & 1.3B (6.3\%)
    & 100.0$\pm$0.0 & 100.0$\pm$0.0 & 100.0$\pm$0.0 & 1.0$\pm$0.0
    & 57.4$\pm$4.0 & 69.6$\pm$8.9 & 100.0$\pm$0.0 & 0.5$\pm$0.1 \\

    \rowcolor[HTML]{E9ECEF} $\ourmodel$ & 1.3B (6.3\%)
    & 100.0$\pm$0.0
    & 100.0$\pm$0.0
    & 100.0$\pm$0.0
    & 1.0$\pm$0.0
    & \textbf{73.5}$\pm$3.0
    & \textbf{80.8}$\pm$1.0
    & 100.0$\pm$0.0
    & \textbf{0.7}$\pm$ 0.0\\
    \bottomrule
\end{tabular}
\end{adjustbox}
\end{sc}
\end{small}
\end{center}
\vspace{-5pt}
\end{table*}

\subsection{Main Result}
\noindent\textbf{Open-loop continual embodied tasks.}
% 0513 실험 목적이 명확하지 않음; 어떤 특성을 검증하기 위한 어떤 실험 세팅이라는 표현이 필요함.
% 0513 wj: 넵 수정했습니다.
To evaluate the generalization performance of $\ourmodel$ on open-loop continual task planning, we conduct experiments in Table~\ref{tab:main:open} using multiple domains from PDDLGym.
In this setting, the agent generates a complete action sequence without intermediate observations and receives only binary task feedback (success or failure) before proceeding to the next task.
% 0513 누구보다 더 좋은 것이라는 표현이 명확해야 함. 
% 0513 wj: 넵 수정했습니다.
$\ourmodel$ outperforms the strongest baseline, LongMem, in the Minecraft and Rearrangement domains, achieving improvements of 13.6\% in \text{CSR}, 11.7\% in \text{CGC}, and 0.15 in \text{SPL} on the test set, thereby demonstrating superior structured and adaptive reasoning capabilities. Similar performance gains are observed in the GlibRearrangement domain as well. More experimental results are provided in Appendix.
% 세부 분석
The single-step planning baselines such as \text{ZSP} and \text{RAP}, which rely on in-context retrieval-augmented generation~\cite{icrag}, show limited reasoning capacity on unseen tasks. The agentic workflow baselines such as \text{CoT}, \text{ToT}, and \text{GoT}, which use reasoning-guidance prompts crafted from the train set~\cite{llmagent:deder}, perform slightly better, but remain far from achieving reliable task success. The memory-augmented LMs such as \text{LongMem} and \text{LM2} surpass the fine-tuning baseline \text{BUTLER}, but their average CSR on the test set is still 15.2\% lower than that of $\ourmodel$.
% This result highlights the effectiveness of our specialized memory structure for embodied task reasoning.

% \begin{comment}
\begin{table*}[b]
\vspace{-5pt}
\caption{
Performance on closed-loop continual embodied tasks in VirtualHome and ALFWorld.
}
\label{tab:main:closed}
\vspace{-5pt}
\begin{center}
\begin{small}
\begin{sc}
\begin{adjustbox}{width=0.99\linewidth}
\begin{tabular}{l ccc ccc ccc}
    \toprule
    \multirow{2}{*}{Method}
    & \multicolumn{3}{c}{Train} & \multicolumn{3}{c}{Seen} & \multicolumn{3}{c}{Unseen} \\
     \cmidrule(rl){2-4} \cmidrule(rl){5-7} \cmidrule(rl){8-10}
     & CSR ($\uparrow$) & CGC ($\uparrow$) & SPL ($\uparrow$)  & CSR ($\uparrow$)  & CGC ($\uparrow$) & SPL ($\uparrow$)  & CSR ($\uparrow$) & CGC ($\uparrow$) & SPL ($\uparrow$) \\
    \midrule
    \multicolumn{7}{l}{\textbf{Benchmark}: VirtualHome} & \\
    \midrule
    \rowcolor[HTML]{FFFFFF} LLM-planner
    & 61.0$\pm$3.5 & 66.4$\pm$2.9 & 0.4$\pm$0.0
    & 45.5$\pm$1.9 & 47.4$\pm$2.0 & 0.3$\pm$0.0
    & 28.8$\pm$2.2 & 30.0$\pm$2.1 & 0.2$\pm$0.0  \\
    \rowcolor[HTML]{E9ECEF} ReAct
    & 63.6$\pm$1.1 & 70.0$\pm$0.6 & 0.4$\pm$0.0
    & 54.7$\pm$1.3 & 56.4$\pm$0.8 & 0.4$\pm$0.0
    & 32.7$\pm$3.5 & 34.3$\pm$3.6 & 0.3$\pm$0.0  \\
    \rowcolor[HTML]{E9ECEF} Reflexion
    & 60.8$\pm$5.9 & 68.8$\pm$5.5 & 0.4$\pm$0.0
    & 57.2$\pm$3.8 & 61.6$\pm$5.4 & 0.4$\pm$0.0
    & 33.7$\pm$2.5 & 35.3$\pm$2.1 & 0.3$\pm$0.0  \\
    \rowcolor[HTML]{FFFFFF} LongMem
    & 80.5$\pm$2.9 & 86.2$\pm$2.5 & 0.8$\pm$0.0
    & 63.3$\pm$5.6 & 68.3$\pm$6.3 & 0.6$\pm$0.1
    & 45.7$\pm$7.3 & 52.2$\pm$8.9 & 0.4$\pm$0.1  \\
    \rowcolor[HTML]{FFFFFF} LM2
    & 80.2$\pm$4.2 & 85.2$\pm$2.9 & 0.8$\pm$0.0
    & 53.6$\pm$5.9 & 57.6$\pm$5.2 & 0.5$\pm$0.1
    & 38.8$\pm$4.4 & 43.3$\pm$4.4 & 0.3$\pm$0.0  \\
    \rowcolor[HTML]{FFFFFF} DT-Mem
    & 77.3$\pm$3.8 & 80.8$\pm$4.8 & 0.7$\pm$0.0
    & 69.3$\pm$5.7 & 71.9$\pm$5.9 & 0.7$\pm$0.1
    & 48.7$\pm$5.9 & 52.6$\pm$6.0 & 0.5$\pm$0.1  \\
    \rowcolor[HTML]{FFFFFF} Optimus-2
    & 79.3$\pm$5.4 & 83.9$\pm$4.0 & 0.8$\pm$0.1
    & 70.4$\pm$4.4 & 74.0$\pm$3.4 & 0.7$\pm$0.1
    & 44.0$\pm$6.1 & 50.9$\pm$7.3 & 0.4$\pm$0.1  \\
    \rowcolor[HTML]{E9ECEF} $\ourmodel$
    & \textbf{89.8}$\pm$1.9 & \textbf{92.3}$\pm$1.1 & \textbf{0.9}$\pm$0.0
    & \textbf{78.9}$\pm$4.5 & \textbf{81.7}$\pm$2.5 & \textbf{0.8}$\pm$0.0
    & \textbf{61.1}$\pm$2.2 & \textbf{69.3}$\pm$2.6 & \textbf{0.6}$\pm$0.0  \\
    \midrule
    \multicolumn{7}{l}{\textbf{Benchmark}: ALFWorld} & \\
    \midrule
    \rowcolor[HTML]{FFFFFF} LLM-planner
    & 52.4$\pm$2.6 & 68.1$\pm$2.0 & 0.5$\pm$0.0
    & 14.0$\pm$0.8 & 21.7$\pm$0.7 & 0.1$\pm$0.0
    & 3.3$\pm$0.4 & 11.7$\pm$0.5 & 0.0$\pm$0.0  \\
    \rowcolor[HTML]{E9ECEF} ReAct
    & 46.3$\pm$1.5 & 65.3$\pm$1.1 & 0.4$\pm$0.0
    & 12.8$\pm$0.5 & 21.4$\pm$0.6 & 0.1$\pm$0.0
    & 2.9$\pm$0.2 & 11.6$\pm$0.2 & 0.0$\pm$0.0  \\
    \rowcolor[HTML]{E9ECEF} Reflexion
    & 44.3$\pm$0.7 & 64.1$\pm$0.7 & 0.4$\pm$0.0
    & 13.0$\pm$0.5 & 21.6$\pm$0.8 & 0.1$\pm$0.0
    & 2.9$\pm$0.4 & 11.8$\pm$0.4 & 0.0$\pm$0.0  \\
    \rowcolor[HTML]{FFFFFF} LongMem
    & 50.1$\pm$3.1 & 58.6$\pm$2.4 & 0.5$\pm$0.0
    & 48.6$\pm$0.6 & 57.3$\pm$0.7 & 0.5$\pm$0.0
    & 45.2$\pm$0.8 & 54.5$\pm$0.8 & 0.5$\pm$0.0  \\
    \rowcolor[HTML]{FFFFFF} LM2
    & 63.8$\pm$1.5 & 71.5$\pm$1.4 & 0.6$\pm$0.0
    & 42.6$\pm$2.4 & 46.9$\pm$2.0 & 0.4$\pm$0.0
    & 38.7$\pm$2.1 & 45.2$\pm$2.5 & 0.4$\pm$0.0  \\
    \rowcolor[HTML]{FFFFFF} DT-Mem
    & 57.7$\pm$2.2 & 61.7$\pm$2.7 & 0.6$\pm$0.0
    & 41.3$\pm$2.9 & 48.0$\pm$3.3 & 0.4$\pm$0.0
    & 38.0$\pm$3.8 & 44.1$\pm$4.2 & 0.4$\pm$0.0  \\
    \rowcolor[HTML]{FFFFFF} Optimus-2
    & 59.5$\pm$1.5 & 67.4$\pm$1.3 & 0.6$\pm$0.0
    & 52.8$\pm$0.9 & 61.6$\pm$0.7 & 0.5$\pm$0.0
    & 49.1$\pm$0.7 & 58.7$\pm$0.6 & 0.5$\pm$0.0  \\
    \rowcolor[HTML]{E9ECEF} $\ourmodel$
    & \textbf{69.6}$\pm$2.7 & \textbf{76.2}$\pm$2.1 & \textbf{0.7}$\pm$0.0
    & \textbf{61.1}$\pm$1.1 & \textbf{68.6}$\pm$1.3 & \textbf{0.6}$\pm$0.0
    & \textbf{59.7}$\pm$1.4 & \textbf{67.9}$\pm$1.3 & \textbf{0.6}$\pm$0.0  \\
    \bottomrule
    \end{tabular}
\end{adjustbox}
\end{sc}
\end{small}
\end{center}
% \vspace{-10pt}
\end{table*}
% \end{comment}

\noindent\textbf{Closed-loop continual embodied tasks.}
% 0513 이것도 이 실험의 목적이 좀 더 명확했으면 함. 
% 0513 테이블이 각 환경별로 되어 있는데, 성능 비교는 통합되어 있는 건가요?... 테이블과 성능 비교 설명이 일관되게 처리되어야 함. 
% 0513 wj: 넵 수정했습니다.
% 0514 wj: seen unseen 을 사용한 이유를 고려해서 목적 구체화화
To further evaluate the generalization performance of $\ourmodel$ alongside its adaptability in dynamic settings, we conduct experiments under a closed-loop continual task planning setup in VirtualHome and ALFWorld. The test set is divided into seen and unseen sets, enabling a detailed assessment of the agent’s structured and adaptive reasoning capabilities.
Unlike open-loop settings, the agent selects actions sequentially in response to intermediate observations.
% 0513 이것도 명확하게 누구 대비 좋은 성적인지를 명시해야 함. 
% 0513 train, seen, unseen 간 성능 패턴에 대한 분석 있으면 좋겠음. 원래는 성능 차이가 unseen 더 나으면 좋을 것 같은데요.?
% 0513 wj: 넵 수정했습니다.
% 
In Table~\ref{tab:main:closed}, $\ourmodel$ outperforms the strongest baseline in VirtualHome, \text{DT-Mem}, with average improvements of 12.5\% in \text{CSR} and 11.5\% in \text{CGC} on the train set, 9.6\% and 9.8\% on the seen set, and 12.4\% and 16.7\% on the unseen set, respectively.
In ALFWorld, $\ourmodel$ outperforms the strongest baseline, \text{Optimus-2}, achieving average gains of 9.8\% in \text{CSR} and 8.8\% in \text{CGC} on the train set, 8.3\% and 7.0\% on the seen set, and 10.6\% and 9.2\% on the unseen set, respectively.
Notably, the unseen sets show greater performance gains than the seen sets. Combined with an average improvement of 0.12 in \text{SPL}, these results indicate that $\ourmodel$ performs effective symbolic reasoning.
In Appendix, additional results for each incremental configuration are provided, along with complete baseline comparisons.
% 세부 분석
Specifically, both \text{LLM-Planner} and agentic workflows such as \text{ReAct} and \text{Reflexion} exhibit limited capability for symbolic reasoning across benchmarks.
While \text{Reflexion} leverages past experiences via verbal feedback, it appears to lack the robust reasoning capabilities required in dynamic and complex tasks.
Memory-augmented approaches for embodied agents, such as \text{DT-Mem} and \text{Optimus-2}, outperform other baselines.
Yet, $\ourmodel$ achieves higher performance, surpassing both \text{DT-Mem} and \text{Optimus-2} by an average of 11.2\% in \text{CSR} and 11.6\% in \text{CGC}, showing the effectiveness of neurosymbolic proceduralization.

\subsection{Analysis and Ablation}

\begin{table*}[ht]
    % \vspace{-5pt}
    \caption{Analysis on proceduralization. \textsc{Latency} denotes the agent's planning time in seconds. \textsc{Tokens} denote the total number of input and output tokens used.}
    \vspace{-5pt}
    \begin{center}
    \begin{small}
    \begin{sc}
    \begin{adjustbox}{width=0.99\linewidth}
    \begin{tabular}{ll ccc ccc}
    \toprule
    \multirow{2}{*}{Method}
    & \multirow{2}{*}{LM}
    & \multicolumn{3}{c}{Task performance} & \multicolumn{3}{c}{Reasoning load}  \\ 
    \cmidrule(rl){3-5} \cmidrule(rl){6-8}
    & & CSR ($\uparrow$) & CGC ($\uparrow$) & SPL ($\uparrow$) & Latency ($\downarrow$) & In tokens ($\downarrow$) & Out tokens ($\downarrow$)  \\
    \midrule
    \multirow{3}{*}{BoT} & \cellcolor[HTML]{E9ECEF}LLaMa3.1-8B       
    & 53.0$\pm$0.5 & 63.5$\pm$0.4 & 0.3$\pm$0.0 & 59.5$\pm$1.9 & 8007.9$\pm$103.9 & 1315.4$\pm$28.1 \\
    & LLaMa3.1-70B      
    & 81.9$\pm$0.4 & 85.1$\pm$0.3 & 0.6$\pm$0.0 & 75.1$\pm$3.8 & 7651.0$\pm$127.7 & 794.1$\pm$33.4 \\
    & GPT4.1
    & 92.1$\pm$0.3 & 93.6$\pm$0.2 & 0.7$\pm$0.0 & 22.2$\pm$2.8 & 7986.1$\pm$144.2 & 1202.2$\pm$197.2 \\
    \midrule
    \multirow{3}{*}{LRM} & \cellcolor[HTML]{E9ECEF}DeepSeek-R1-8B
    & 11.5$\pm$0.3 & 15.6$\pm$0.3 & 0.1$\pm$0.0 & 111.0$\pm$3.3 & 3198.5$\pm$15.8 & 2187.6$\pm$69.0 \\
    & DeepSeek-R1-70B
    & 26.5$\pm$0.4 & 27.5$\pm$0.4 & 0.2$\pm$0.0 & 209.4$\pm$9.2 & 3198.5$\pm$15.8 & 1679.3$\pm$87.5 \\
    & o3-mini
    & 78.9$\pm$0.4 & 80.8$\pm$0.4 & 0.5$\pm$0.0 & 18.6$\pm$1.7 & 3214.6$\pm$15.7 & 2113.9$\pm$63.2 \\
    \midrule
    \multirow{3}{*}{PlaSma} & LLaMa3.2-1B
    & 67.4$\pm$0.5 & 71.9$\pm$0.4 & 0.7$\pm$0.0    & 2.7$\pm$0.5 & 3221.8$\pm$45.3 & 32.7$\pm$4.6 \\
    & LLaMa3.2-3B
    & 70.7$\pm$0.4 & 75.7$\pm$0.3 & 0.7$\pm$0.0 & 7.2$\pm$0.7 & 3247.7$\pm$17.2 & 29.5$\pm$5.5 \\
    & \cellcolor[HTML]{E9ECEF}LLaMa3.1-8B
    & 80.5$\pm$0.5 & 89.2$\pm$2.3 & 0.8$\pm$0.0 & 18.4$\pm$5.8 & 3371.0$\pm$13.5 & 122.4$\pm$41.6 \\
    \midrule
    \multirow{3}{*}{$\ourmodel$} & LLaMa3.2-1B
    & 73.2$\pm$0.4 & 76.0$\pm$0.4 & 0.7$\pm$0.0 & 1.2$\pm$0.3 & 3168.5$\pm$0.0 & 30.1$\pm$5.3 \\
    & LLaMa3.2-3B
    & 83.6$\pm$2.0 & 88.8$\pm$2.0 & 0.8$\pm$0.0 & 3.5$\pm$0.3 & 3169.5$\pm$0.0 & 43.6$\pm$5.3 \\
    & \cellcolor[HTML]{E9ECEF}LLaMa3.1-8B
    & 89.0$\pm$2.0 & 93.5$\pm$1.8 & 0.9$\pm$0.0 & 5.2$\pm$0.7 & 3155.5$\pm$0.0 & 41.9$\pm$6.0 \\
    \bottomrule
    \end{tabular}
    \end{adjustbox}
    \end{sc}
    \end{small}
    \end{center}
    \label{tab:ana:proc}
    \vspace{-5pt}
\end{table*}
\noindent\textbf{Analysis on proceduralization.}
% 0513 어떤 관점에서 proceduralization 을 평가하는지 명확하게 명시할 것. 
% 0513 wj: 넵 수정했습니다.
% 0514 wj: existing proceduralization methods  좀 더 앞으로
Table~\ref{tab:ana:proc} presents a comparative analysis of our neurosymbolic proceduralization method to existing proceduralization methods, evaluated in terms of task performance and reasoning efficiency, with a particular focus on enabling timely reasoning through single-step inference.
% The comparison includes existing proceduralization methods as originally applied to different LMs in their respective papers.
To ensure a fair comparison, we additionally include a unified setting in which all methods are evaluated under identical inference conditions using the same LLaMA 3.1-8B backbone, highlighted in gray background in the table.
%  0513 숫자를 넣어서, 비교군을 넣어서. 어떤점을 분석한 것인지를 명확하게.
% 0513 wj: 넵 수정했습니다.
$\ourmodel$ achieves the lowest average plan generation latency, the highest task success rate, and the minimal input and output token usage.
% Under these conditions, $\ourmodel$ achieves the lowest average plan generation latency of 5.0 seconds, the highest task success rate of 89.0\%, and the fewest input and output tokens—averaging 3155.5 and 41.9, respectively.
\text{BoT} and \text{LRM} exhibit latencies that are 54.3 and 105.8 seconds longer than $\ourmodel$, respectively, along with 6,125.9 and 2,188.7 more total tokens consumed, due to their reliance on multi-step reasoning.
% 0513 좀 더명확하게 비교 필요; 숫자 넣어서.
% 0513 wj: 넵 수정했습니다.
\text{PlaSma}, which distills procedural knowledge from larger to smaller LMs, achieves competitive results with efficient inference, reaching 80.5\% in CSR using an 8B LM. Yet, $\ourmodel$ outperforms it with a higher CSR of 83.6\% while operating with only a 3B LM.

\begin{figure*}[h]
% \vspace{-5pt}
\begin{center}
\centerline{\includegraphics[width=0.99\linewidth]{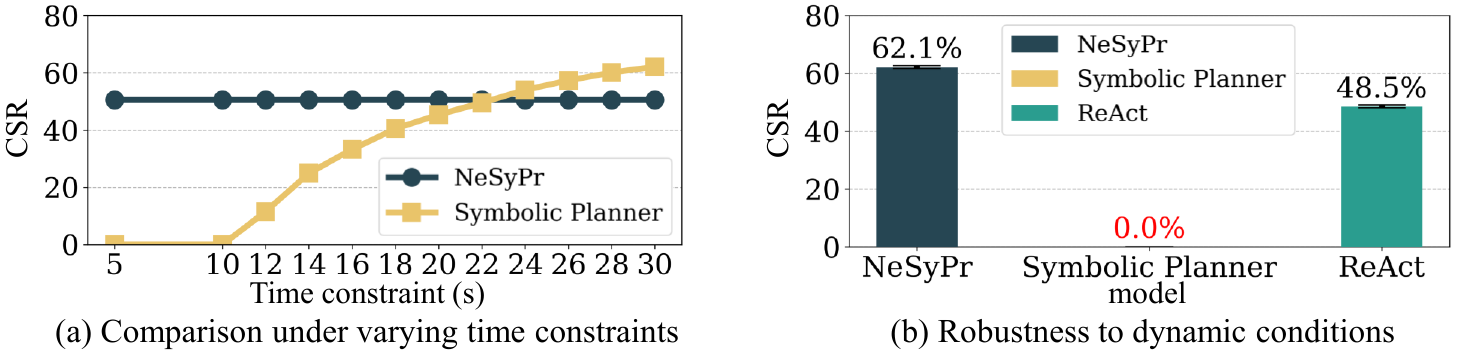}}
% \vspace{-5pt}
\caption{Comparison with symbolic planner}
\label{fig:ana:er}
\end{center}
\vspace{-10pt}
\end{figure*}

\noindent\textbf{Comparison with online symbolic planner.} 
%  0513 이것은 online symbol tool 과의 비교로 설명하는게 어떤가요? 제목이 와닿지 않음.  그리고 위에 있는 latency 와 response time 도 차이가 있는 듯 한데.. 이건 방법론과는 별개로 측정 방법이 달라야 할 듯. 
% 0513 wj: 넵 constratin 관련 설명으로 바꿨습니다.
Figure~\ref{fig:ana:er} analyzes the impact of neurosymbolic proceduralization on automated decision-making in agents.
Figure~\ref{fig:ana:er}(a) compares the task success rate of $\ourmodel$ and a symbolic planner under a strict time constraint, allowing 1\% violation.
For tasks where the symbolic planner takes over 10 seconds to find a solution, $\ourmodel$ completes planning within a 5-second constraint while achieving 50.6\% in CSR. In comparison, the symbolic planner takes up to 22 seconds to reach similar performance.
Figure~\ref{fig:ana:er}(b) evaluates robustness on unseen tasks with dynamic conditions.
While the symbolic planner fails when input information is incomplete, $\ourmodel$ maintains stable performance and even outperforms ReAct using GPT-4o~\cite{gpt4o}.

\begin{table*}[ht]
% \vspace{-5pt}
\caption{
Analysis on continual embodied task adaptation scenario. During continual task inference, the entire test set is periodically evaluated across 15 intermediate continual evaluation phases.
}
\label{tab:ana:ca}
\vspace{-5pt}
\begin{center}
\begin{small}
\begin{sc}
\begin{adjustbox}{width=0.99\linewidth}
\begin{tabular}{l c ccccc ccccc ccccc}
    \toprule
    \multirow{2}{*}{Method}
    & \multirow{2}{*}{Metric}
    & \multicolumn{15}{c}{Continual evaluation phase}\\ 
    \cmidrule(rl){3-17}
    & & 1 & 2 & 3 & 4 & 5 & 6 & 7 & 8 & 9 & 10 & 11 & 12 & 13 & 14 & 15 \\
    \midrule
    \multirow{3}{*}{LongMem} & 
    SR ($\uparrow$)     & 57.1  & 56.0  & 58.6  & 54.6  & 47.5  & 51.1  & 52.4  & 51.5  & 52.2  & 52.4  & 53.0  & 53.6  & 55.7  & 54.4  & 53.8 \\
    \cmidrule(l){2-17}
    & \cellcolor[HTML]{E9ECEF}FWT ($\uparrow$) & \cellcolor[HTML]{E9ECEF}0.0 & \cellcolor[HTML]{E9ECEF}0.0 & \cellcolor[HTML]{E9ECEF}0.0 & \cellcolor[HTML]{E9ECEF}0.0 & \cellcolor[HTML]{E9ECEF}0.0 & \cellcolor[HTML]{E9ECEF}0.0 & \cellcolor[HTML]{E9ECEF}0.0 & \cellcolor[HTML]{E9ECEF}0.0 & \cellcolor[HTML]{E9ECEF}0.0 & \cellcolor[HTML]{E9ECEF}0.0 & \cellcolor[HTML]{E9ECEF}0.0 & \cellcolor[HTML]{E9ECEF}0.0 & \cellcolor[HTML]{E9ECEF}0.0 & \cellcolor[HTML]{E9ECEF}0.0 & \cellcolor[HTML]{E9ECEF}0.0 \\
    & \cellcolor[HTML]{E9ECEF}BWT ($\uparrow$) & \cellcolor[HTML]{E9ECEF}0.0 & \cellcolor[HTML]{E9ECEF}0.0 & \cellcolor[HTML]{E9ECEF}0.0 & \cellcolor[HTML]{E9ECEF}0.0 & \cellcolor[HTML]{E9ECEF}0.0 & \cellcolor[HTML]{E9ECEF}0.0 & \cellcolor[HTML]{E9ECEF}0.0 & \cellcolor[HTML]{E9ECEF}0.0 & \cellcolor[HTML]{E9ECEF}0.0 & \cellcolor[HTML]{E9ECEF}0.0 & \cellcolor[HTML]{E9ECEF}0.0 & \cellcolor[HTML]{E9ECEF}0.0 & \cellcolor[HTML]{E9ECEF}0.0 & \cellcolor[HTML]{E9ECEF}0.0 & \cellcolor[HTML]{E9ECEF}0.0 \\
    \midrule
    
    \multirow{5}{*}{LM2} & 
    SR ($\uparrow$)     & 64.3  & 52.0  & 55.2  & 57.6  & 52.5  & 53.3  & 50.8  & 50.0  & 49.3  & 48.8  & 49.4  & 50.0  & 52.3  & 52.2  & 51.6 \\
    \cmidrule(l){2-17}
    & \cellcolor[HTML]{E9ECEF}FWT ($\uparrow$)  & \cellcolor[HTML]{E9ECEF}0.0   & \cellcolor[HTML]{E9ECEF}0.0   & \cellcolor[HTML]{E9ECEF}0.0   & \cellcolor[HTML]{E9ECEF}0.0   & \cellcolor[HTML]{E9ECEF}-2.5  & \cellcolor[HTML]{E9ECEF}2.2   & \cellcolor[HTML]{E9ECEF}1.6   & \cellcolor[HTML]{E9ECEF}1.5   & \cellcolor[HTML]{E9ECEF}1.4   & \cellcolor[HTML]{E9ECEF}1.2   & \cellcolor[HTML]{E9ECEF}1.2   & \cellcolor[HTML]{E9ECEF}1.2   & \cellcolor[HTML]{E9ECEF}1.1   & \cellcolor[HTML]{E9ECEF}1.1   & \cellcolor[HTML]{E9ECEF}2.2 \\
    & \cellcolor[HTML]{E9ECEF}BWT ($\uparrow$)  & \cellcolor[HTML]{E9ECEF}0.0   & \cellcolor[HTML]{E9ECEF}0.0   & \cellcolor[HTML]{E9ECEF}0.0   & \cellcolor[HTML]{E9ECEF}0.0   & \cellcolor[HTML]{E9ECEF}0.0   & \cellcolor[HTML]{E9ECEF}-4.4  & \cellcolor[HTML]{E9ECEF}-3.2  & \cellcolor[HTML]{E9ECEF}-3.0  & \cellcolor[HTML]{E9ECEF}-2.9  & \cellcolor[HTML]{E9ECEF}-2.4  & \cellcolor[HTML]{E9ECEF}-2.4  & \cellcolor[HTML]{E9ECEF}-1.2  & \cellcolor[HTML]{E9ECEF}-2.3  & \cellcolor[HTML]{E9ECEF}-1.1  & \cellcolor[HTML]{E9ECEF}-2.2 \\
    \cmidrule(l){2-17}
    & FR ($\downarrow$) & 0.0   & 0.0   & 0.0   & 0.0   & 9.1   & 8.0   & 6.1   & 5.9   & 5.7   & 4.9   & 4.8   & 4.7   & 4.3   & 4.2   & 4.0 \\
    & RR ($\uparrow$)   & 0.0   & 0.0   & 0.0   & 0.0   & 0.0   & 0.0   & 0.0   & 0.0   & 0.0   & 0.0   & 0.0   & 2.4   & 0.0   & 2.4   & 0.0 \\
    % & MF                & 1.05M & 1.05M & 1.05M & 1.05M & 1.05M & 1.05M & 1.05M & 1.05M & 1.05M & 1.05M & 1.05M & 1.05M & 1.05M & 1.05M & 1.05M \\
    \midrule

    \multirow{5}{*}{$\ourmodel$} & 
    SR ($\uparrow$)     & 64.3  & 64.0  & 62.1  & 63.6  & 65.0  & 66.7  & 61.9  & 60.6  & 60.9  & 61.0  & 61.5  & 61.9  & 63.6  & 62.2  & 61.3 \\
    \cmidrule(l){2-17}
    & \cellcolor[HTML]{E9ECEF}FWT ($\uparrow$)  & \cellcolor[HTML]{E9ECEF}7.1   & \cellcolor[HTML]{E9ECEF}4.0   & \cellcolor[HTML]{E9ECEF}3.4   & \cellcolor[HTML]{E9ECEF}3.0   & \cellcolor[HTML]{E9ECEF}3.3   & \cellcolor[HTML]{E9ECEF}2.2   & \cellcolor[HTML]{E9ECEF}1.6   & \cellcolor[HTML]{E9ECEF}1.5   & \cellcolor[HTML]{E9ECEF}1.4   & \cellcolor[HTML]{E9ECEF}1.2   & \cellcolor[HTML]{E9ECEF}1.2   & \cellcolor[HTML]{E9ECEF}1.2   & \cellcolor[HTML]{E9ECEF}1.1   & \cellcolor[HTML]{E9ECEF}2.2   & \cellcolor[HTML]{E9ECEF}3.2 \\
    & \cellcolor[HTML]{E9ECEF}BWT ($\uparrow$)  & \cellcolor[HTML]{E9ECEF}0.0   & \cellcolor[HTML]{E9ECEF}4.0   & \cellcolor[HTML]{E9ECEF}3.4   & \cellcolor[HTML]{E9ECEF}3.0   & \cellcolor[HTML]{E9ECEF}1.7   & \cellcolor[HTML]{E9ECEF}2.2   & \cellcolor[HTML]{E9ECEF}4.8   & \cellcolor[HTML]{E9ECEF}4.5   & \cellcolor[HTML]{E9ECEF}4.3   & \cellcolor[HTML]{E9ECEF}4.9   & \cellcolor[HTML]{E9ECEF}4.8   & \cellcolor[HTML]{E9ECEF}4.8   & \cellcolor[HTML]{E9ECEF}4.5   & \cellcolor[HTML]{E9ECEF}4.4   & \cellcolor[HTML]{E9ECEF}4.3 \\
    \cmidrule(l){2-17}
    & FR ($\downarrow$) & 0.0   & 0.0   & 0.0   & 0.0   & 0.0   & 0.0   & 0.0   & 0.0   & 0.0   & 0.0   & 0.0   & 0.0   & 0.0   & 0.0   & 0.0 \\
    & RR ($\uparrow$)   & 0.0   & 12.5  & 10.0  & 9.1   & 7.7   & 7.1   & 13.0  & 12.0  & 11.5  & 12.9  & 12.9  & 12.9  & 12.9  & 12.5  & 12.1 \\
    % & MF        & 1025    & 1030    & 1066    & 1066    & 1066    & 1117    & 1121    & 1126    & 1194    & 1243 \\
    \bottomrule
\end{tabular}
\end{adjustbox}
\end{sc}
\end{small}
\end{center}
\vspace{-5pt}
\end{table*}

\noindent\textbf{Analysis on continual task adaptation.} 
Table~\ref{tab:ana:ca} presents continual adaptation results, highlighting $\ourmodel$'s adaptive reasoning using metrics from continual learning~\cite{liu2023libero}.
% metrics
\text{Forward Transfer (FWT)} measures how newly acquired procedures improve performance on future tasks by comparing average per-task \text{SR} with overall \text{CSR}.
\text{Backward Transfer (BWT)} compares current \text{CSR} with that obtained when re-evaluating earlier tasks using retained procedures.
\text{Forgetting Rate (FR)} is the proportion of previously successful tasks that fail upon re-evaluation, while \text{Recovery Rate (RR)} is the proportion of previously failed tasks that later succeed.
% memory perspective
\text{LongMem}, which stores key-value states for retrieval, shows no improvement in \text{FWT} and no degradation in \text{BWT}.
In contrast, \text{LM2}, which implicitly maintains a working memory to extend context, shows moderate improvement in \text{FWT} but fails to preserve \text{BWT}.
By leveraging both valid and invalid procedures, $\ourmodel$ achieves superior performance in both \text{FWT} and \text{BWT}.
Notably, its \text{FR} converges to zero, and it attains about 12.0\% \text{RR}, demonstrating effective adaptation without symbolic tools.
% \end{comment}

\begin{figure*}[ht]
% \vspace{-5pt}
\begin{center}
\centerline{\includegraphics[width=0.99\linewidth]{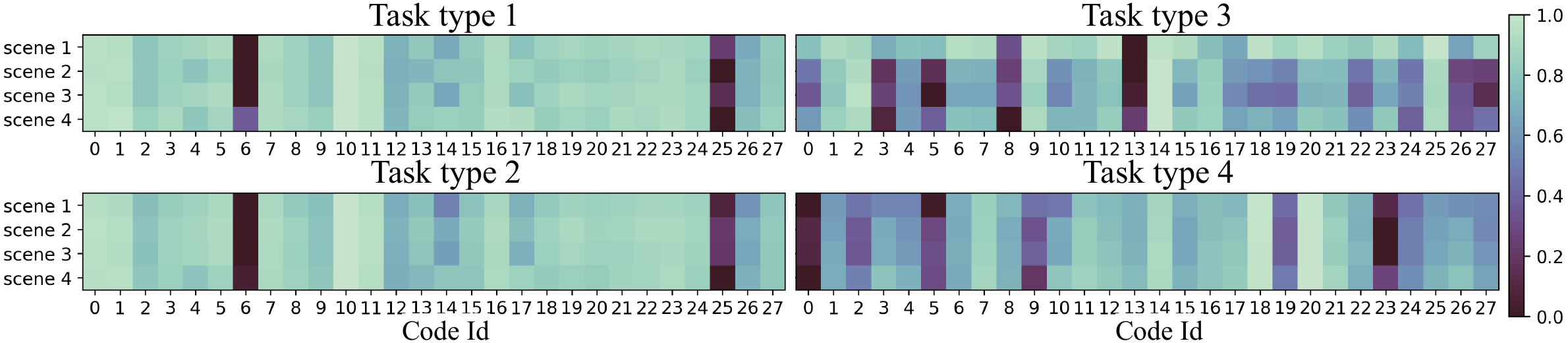}}
% \vspace{-5pt}
\caption{Analysis on procedural memory interpretability}
\label{tab:ana:mem_inter}
\end{center}
\vspace{-10pt}
\end{figure*}

\noindent\textbf{Analysis on procedural memory interpretation.} 
Figure~\ref{tab:ana:mem_inter} shows a heatmap of procedure-unit $c$ (in Eq.~\eqref{eq:vq}) usage across 4 task types and scenes.
Task types 1 and 2 share similar solution and exhibit consistent $c$ usage patterns across different scenes.
In contrast, task types 3 and 4 involve object-specific actions (e.g., picking up or turning on items), which are more sensitive to scene variations and lead to more divergent patterns.

\begin{figure*}[ht]
    % \vspace{-5pt}
    \begin{minipage}[t]{.425\linewidth}
    \captionof{table}{Application with different LMs}
    \vspace{-5pt}
    \begin{center}
    \begin{small}
    \begin{sc}
    \begin{adjustbox}{width=0.98\linewidth}
    \begin{tabular}{l l cc}
    \toprule
    Method & LM
    & CSR ($\uparrow$) & CGC ($\uparrow$) \\
    \midrule
    BUTLER & Qwen2-0.5B
    & 41.3$\pm$1.0 & 54.7$\pm$3.3 \\
    \rowcolor[HTML]{DEE2E6} $\ourmodel$ & Qwen2-0.5B
    & 51.7$\pm$1.7 & 60.7$\pm$2.9 \\

    BUTLER & Llama3.2-1B
    & 51.4$\pm$1.9 & 66.7$\pm$2.6 \\
    \rowcolor[HTML]{DEE2E6} $\ourmodel$ & Llama3.2-1B
    & 65.2$\pm$1.4 & 68.9$\pm$2.4 \\

    BUTLER & Qwen2.5-1.5B
    & 50.3$\pm$1.5 & 58.6$\pm$3.4 \\
    \rowcolor[HTML]{DEE2E6} $\ourmodel$ & Qwen2.5-1.5B
    & 56.7$\pm$1.3 & 59.0$\pm$1.5 \\

    BUTLER & Gemma2-2B
    & 57.5$\pm$1.3 & 59.9$\pm$0.5 \\
    \rowcolor[HTML]{DEE2E6} $\ourmodel$ & Gemma2-2B
    & 66.4$\pm$4.1 & 73.8$\pm$0.5 \\

    BUTLER & Llama3.2-3B
    & 64.2$\pm$1.2 & 73.8$\pm$2.5 \\
    \rowcolor[HTML]{DEE2E6} $\ourmodel$ & Llama3.2-3B
    & 74.9$\pm$2.4 & 77.1$\pm$1.8 \\
    \bottomrule
    \end{tabular}
    \end{adjustbox}
    \end{sc}
    \end{small}
    \end{center}
    \label{tab:ana:lms}
    \end{minipage}
    \hfill
    \begin{minipage}[t]{.53\linewidth}
    \captionof{table}{Ablation study of $\ourmodel$}
    \vspace{-5pt}
    \begin{center}
    \begin{small}
    \begin{sc}
    \begin{adjustbox}{width=0.98\linewidth}
    \begin{tabular}{l cc}
    \toprule
    Method & CSR ($\uparrow$) & CGC ($\uparrow$) \\
    \midrule
    $\ourmodel$ (Full)
    & 65.2$\pm$1.4 & 68.9$\pm$2.4 \\

    \rowcolor[HTML]{DEE2E6} $\ourmodel$ w/o EMA update of $\mathcal{C}$
    & 63.0$\pm$3.1 & 64.8$\pm$3.7 \\
    
    \rowcolor[HTML]{DEE2E6} $\ourmodel$ w/o $\mathcal{C}$
    & 47.9$\pm$8.4 & 56.5$\pm$4.6 \\

    $\ourmodel$ w/ $\PMB^{+}$ only
    & 63.2$\pm$1.3 & 65.1$\pm$2.4 \\

    $\ourmodel$ w/ $\PMB^{-}$ only
    & 63.3$\pm$1.4 & 66.4$\pm$2.5 \\

    \rowcolor[HTML]{DEE2E6} $\ourmodel$ w/o CP
    & 59.9$\pm$1.4 & 62.9$\pm$2.4 \\

    \rowcolor[HTML]{DEE2E6} $\ourmodel$ w/o CP, follow $\PMB^{+}$
    & 60.9$\pm$1.1 & 64.4$\pm$2.2 \\

    \rowcolor[HTML]{DEE2E6} $\ourmodel$ w/o CP, follow $\PMB^{-}$
    & 59.7$\pm$1.4& 62.2$\pm$2.3 \\
    \bottomrule
    \end{tabular}
    \end{adjustbox}
    \end{sc}
    \end{small}
    \end{center}
    \label{tab:ana:abl}
    \vspace{-10pt}
    \end{minipage}
\vspace{-10pt}
\end{figure*}

\noindent\textbf{Application with different LMs.} 
Table~\ref{tab:ana:lms} reports the performance of $\ourmodel$ using three LM families across five model sizes~\cite{yang2024qwen2technicalreport, qwen2025qwen25technicalreport, llama3.2, team2024gemma}, evaluated on the \text{Minecraft} domain.
Across all LMs, $\ourmodel$ achieves an average \text{CSR} improvement of 10.0\% over BUTLER.  
Performance generally improves with model size, although the degree of improvement varies across different LM families.

\noindent\textbf{Ablation study.} 
%  0513 전반적으로 각 component 을 언급하면서, 서능 저하와 연결하면 좋겠음. 
% 0513 wj: 넵 수정했습니다.
Table~\ref{tab:ana:abl} presents an ablation study that evaluates the contribution of each component in $\ourmodel$, using \text{Minecraft}.
Using the procedure-book $\mathcal{C}$ without EMA updates (i.e., $\ourmodel$ \textsc{w/o EMA update of} $\mathcal{C}$) results in a performance drop of 2.2\% in \text{CSR} and 4.1\% in \text{CGC}.  
When $\mathcal{C}$ is removed entirely (i.e., $\ourmodel$ \textsc{w/o} $\mathcal{C}$), the performance drops by 17.3\% in \text{CSR} and 12.4\% in \text{CGC}, demonstrating the importance of a learned $\mathcal{C}$. % 0514 wj: 좀 더 구체적인 importance 
% 0513 심볼에 설명이 되게 이름 추가할 것. 
% 0513 wj: 넵 수정했습니다.
Models using only successful procedures $\PMB^{+}$ (i.e., $\ourmodel$ \textsc{w/} $\PMB^{+}$ \textsc{only}) or only failure procedures $\PMB^{-}$ (i.e., $\ourmodel$ \textsc{w/} $\PMB^{-}$ \textsc{only})-where the original composed procedure is used as the counterpart in contrastive planning, perform 2.0\% and 1.9\% worse in \text{CSR}, respectively, compared to the full version.
Disabling contrastive planning (i.e., $\ourmodel$ \textsc{w/o CP}) leads to a further performance drop of 5.3\% in \text{CSR}.
Without CP, simply following either $\PMB^{+}$ or $\PMB^{-}$ (i.e., $\ourmodel$ \textsc{w/o CP, follow} $\PMB^{+}$ and $\ourmodel$ \textsc{w/o CP, follow} $\PMB^{-}$) leads to less reliable plan generation, with \text{CSR} reductions of 4.3\% and 5.5\%, respectively.
% The full model performs better because it contrasts both signals to guide decision-making.

\begin{figure*}[ht]
% \vspace{-5pt}
\begin{center}
\centerline{\includegraphics[width=0.99\linewidth]{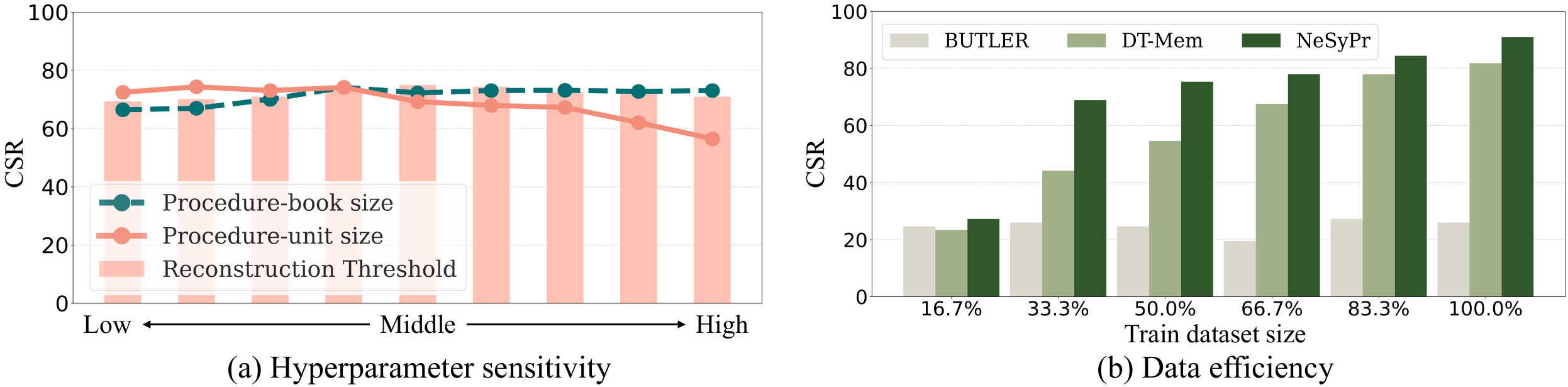}}
% \vspace{-5pt}
\caption{Ablation study on procedural memory hyperparameters and learning data efficiency}
\label{fig:ana:learning}
\end{center}
\vspace{-10pt}
\end{figure*}

% \color{red}
\noindent\textbf{Hyperparameter sensitivity and data efficiency.} 
% 0513 그림하고 용어가 틀림; 그림 글씨가 잘 보임; 이 실험은 그림과 내용이 아예 다른 것 같은데요?
% 0513 wj: 넵 용어 본문과 통일하고 글씨 약간 키웠습니다.
% Figure~\ref{tab:ana:learning} analyzes the hyperparameter sensitivity and interpretability of the procedure-book $\mathcal{C}$.
% % (a)
Figure~\ref{fig:ana:learning}(a) shows that a small size of $\mathcal{C}$ limits task coverage, while a sufficiently large one enables consistently high performance.
In contrast, increasing the size of each procedure-unit $c$ reduces diversity in the composed procedures, leading to performance degradation.
We also observe that setting a low reconstruction threshold $\upsilon$ causes over-generalization by accepting low-similarity procedures, while a high $\upsilon$ leads to under-generalization by rejecting valid ones.
% setting reconstruction threshold $\upsilon$ too low results in over-generalization, as even low-similarity procedures are reconstructed.
% Conversely, a high $\upsilon$ leads to under-generalization by rejecting most validated procedures.
% 
Figure~\ref{fig:ana:learning}(b) shows the data efficiency of each method by reporting task success rates across varying training dataset sizes. $\ourmodel$ consistently achieves higher CSR with less data, indicating superior learning efficiency compared to BUTLER and DT-Mem.

% 
% When the number of codes available for training and inference is insufficient, the agent fails to adequately cover the diverse tasks in the benchmark. However, once the logic codebook reaches a sufficient size, $\ourmodel$ consistently achieves high performance.
% Conversely, as the code size increases, the diversity in restructuring working memory slots is reduced, leading to a decline in performance.
% This is similar to Inductive Logic Programming (ILP)~\cite{ilp:survey}, where determining the noise tolerance during rule generalization is crucial for deciding how broad the subsumption~\cite{ilp} should be.
% 
% (b)
% Figure~\ref{tab:ana:mem}(b) shows a heatmap of $c$ usage across 4 task types and scenes.
% Task types 1 and 2 share similar solution and exhibit consistent $c$ usage patterns across different scenes.
% In contrast, task types 3 and 4 involve object-specific actions (e.g., picking up or turning on items), which are more sensitive to scene variations and lead to more divergent patterns.
% \color{black}

\section{Conclusion}\label{sec:conclusion}
We presented the $\ourmodel$ framework that employs neurosymbolic proceduralization inspired by ACT theory. It performs knowledge compilation by abstracting and generalizing multi-step symbolic path-finding and reasoning into single-step inference within an LM. This enables LM-based agents to conduct embodied reasoning efficiently, without relying on large-scale inference engines or online access to symbolic tools.
Experimental results on PDDLGym, ALFWorld, and VirtualHome show that $\ourmodel$ enables structured, adaptive, and timely reasoning in dynamic embodied environments.

\noindent\textbf{Limitation and future direction.}
As shown in Table~\ref{tab:ana:lms}, $\ourmodel$'s performance partially depends on the pretrained knowledge of the LM.
To mitigate this, we plan to explore a joint learning strategy that combines knowledge distillation from larger LMs with neurosymbolic proceduralization. 
This approach has the potential to enhance generalization to more complex, real-world scenarios.

\newpage
\section*{Acknowledgement}
This work was supported by the Institute of Information \& communications Technology Planning \& Evaluation (IITP) grant funded by the Korea government (MSIT),
% 개성형성
(RS-2022-II220043 (2022-0-00043), Adaptive Personality for Intelligent Agents,
% 자기주도
RS-2022-II221045 (2022-0-01045), Self-directed multi-modal Intelligence for solving unknown, open domain problems,
% AGI
RS-2025-02218768, Accelerated Insight Reasoning via Continual Learning,
% 스타펠로우십
RS-2025-25442569, AI Star Fellowship Support Program (Sungkyunkwan Univ.), and
% AI 대학원
RS-2019-II190421, Artificial Intelligence Graduate School Program (Sungkyunkwan University)),
%  ITRC
IITP-ITRC (Information Technology Research Center) grant funded by the Korea government (MIST) (IITP-2025-RS-2024-00437633, 10\%),
%  ICT 명품 인재
IITP-ICT Creative Consilience Program grant funded by the Korea government (MSIT) (IITP-2025-RS-2020-II201821, 10\%), 
% 삼성
and by Samsung Electronics.

\printbibliography

\newpage

\tableofcontents
\newpage
\appendix

\newpage

\begin{figure*}[t]
\begin{center}
\centerline{\includegraphics[width=0.99\linewidth]{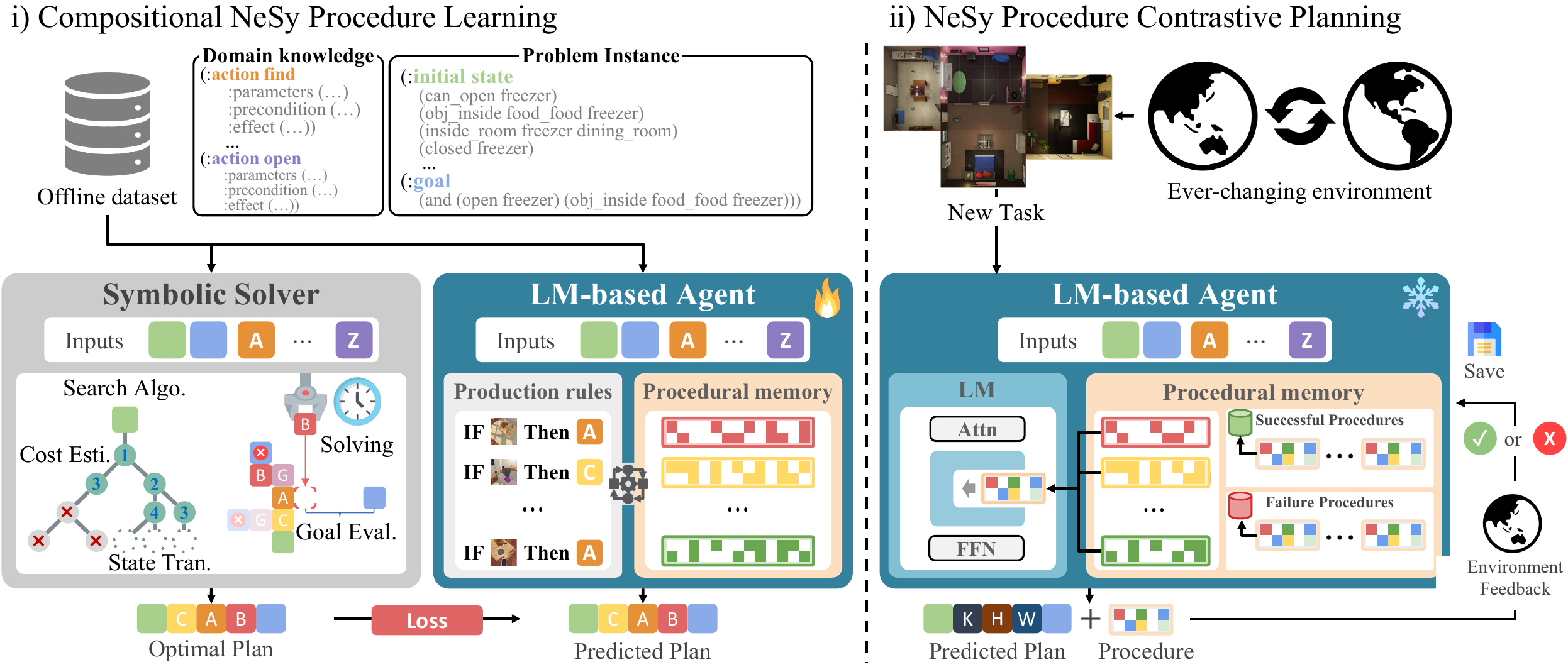}}
% \vspace{-3pt}
\caption{The framework architecture of $\ourmodel$}
\label{app:fig:1}
\end{center}
\vspace{-20pt}
\end{figure*}

\section{\ourmodel: Neurosymbolic Proceduralization}\label{app:sec:approach}
% 0507 wj: intro-prob-method 문제 언급과 왜 이게 최선인지로 설득시켜야
%Building on the objective in Eq.~\eqref{eq:objective}, we introduce $\ourmodel$, a memory-augmented neurosymbolic proceduralization framework for knowledge compilation.
% 
%Drawing on ACT~\cite{ACT}, $\ourmodel$ adapts the interplay among declarative, working, and procedural memory to convert declarative knowledge from symbolic tools into parameterized procedures embeddable in LMs, thereby equipping agents with structured and adaptive reasoning for complex embodied tasks in dynamic environments.

To equip agents with structured and adaptive reasoning for diverse embodied tasks, $\ourmodel$ learns to encode production rules and compose procedures, compressed representations derived from the declarative knowledge of symbolic tools. We refer to this end-to-end learning and utilization process as \textit{neurosymbolic proceduralization}. 
As illustrated in Figure~\ref{app:fig:1}, neurosymbolic proceduralization operates in two phases: \romannumeral 1) a training phase, \textit{compositional NeSy procedure learning}, where procedural knowledge is structured within procedural memory using plans generated by a symbolic tool, and \romannumeral 2) a test phase, \textit{NeSy procedure contrastive planning}, where the agent autonomously adapts to new tasks by contrastively reconstructing procedural representations without relying on symbolic tools.

\subsection{Algorithm}
During phase~\romannumeral 1), the agent trains on offline data comprising symbolically defined problem instances (observations and goals) and associated domain knowledge (action rules).  
The declarative knowledge used by symbolic tools for problem-solving—such as search algorithms, state transitions, cost estimation, and goal evaluation—is internalized as production rules.  
The agent composes these rules into task-solving procedures in procedural memory, which it then exploits to generate plans.  
The algorithm for phase~\romannumeral 1) is provided in Algorithm~\ref{app:alg:learning}.

\begin{algorithm}[h]
\caption{Compositional NeSy Procedure Learning}
\label{app:alg:learning}
\textbf{Initialize} working memory $\WMem \gets [\bm{e}_1, \bm{e}_2, \dots, \bm{e}_S]$ \\
\textbf{Initialize} procedure-book $\mathcal{C} \gets \{c_1, c_2, \ldots, c_K \}$ \\
Memory-augmented LM policy $\pi_\mathrm{LM}$ \\
Offline dataset $D$ \\
Symbolic planner $\pi_{\mathrm{tool}}$
\begin{algorithmic}[1]
\ForAll{$(o, g, \mathrm{DK}) \in D$}
    \State $a \gets \pi_{\mathrm{tool}}(o, g, \mathrm{DK})$ \Comment{\textcolor{black}{Calculate action plan $a$ based on domain knowledge $\mathrm{DK}$}}
    \Statex \textcolor{black}{\text{\ \ \ \ \ \ }\textit{/* LM forward pass */}}
    \State $\HS_0 \gets \text{EmbeddingLayer}(o, g, \mathrm{DK})$
    \For{$l = 1$ \textbf{ to } $L$} 
        \Statex \textcolor{black}{\text{\ \ \ \ \ \ \ \ \ \ \ \ \ }\textit{/* $\mathrm{DecoderBlock}_l$} */}
        \State $\Emb_{\mathrm{self}} \gets \text{SelfAttention}(\HS_{l-1})$
        \State $\Emb_{\mathrm{work}} \gets \text{CrossAttention}(\Emb_{\mathrm{self}}, \WMem)$ \Comment{\textcolor{black}{cf. Eq.~(3)}}
        \State Update $\WMem$ by gated merge of $\Emb_{\mathrm{work}}$ and $\WMem$ \Comment{\textcolor{black}{cf. Eq.~(4)}}
        \State $\PMem \gets \text{VQ}(\WMem, \mathcal{C})$ \Comment{\textcolor{black}{cf. Eq.~(5), (6)}}
        \State $\HS_l \gets \text{FFN}(\Emb_{\mathrm{work}} + \text{Gate}(\PMem))$ \Comment{\textcolor{black}{cf. Eq.~(7)}}
    \EndFor
    \Statex \textcolor{black}{\text{\ \ \ \ \ \ }\textit{/* Loss computation and model update */}}
    \State $\mathcal{L} = -\sum\log \pi_\mathrm{LM}(a \mid o,g,\WMem) + \sum\mathcal{L}_\mathrm{VQ}$ \Comment{\textcolor{black}{cf. Eq.~(8), (9)}}
    \State Update $\pi_\mathrm{LM}$ parameter $\theta$ via loss $\mathcal{L}$
    \State Update $\mathcal{C}$ via EMA
\EndFor
\end{algorithmic}
\end{algorithm}

In phase~\romannumeral 2), using the procedural memory established during phase~\romannumeral 1), the agent performs structured reasoning without access to external symbolic tools (i.e., declarative memory).  
It further engages in adaptive reasoning by contrastively reconstructing procedures from prior ones labeled as successes or failures via environmental feedback.  
The agent continually reinforces plans aligned with valid procedures while suppressing those associated with invalid ones.  
The algorithm for phase~\romannumeral 2) is provided in Algorithm~\ref{app:alg:inference}.

Accordingly, $\ourmodel$ enables LM-based agents to reason robustly across tasks and adapt efficiently to ever-changing environments.

\begin{algorithm}[h]
\caption{NeSy Procedure Contrastive Planning}
\label{app:alg:inference}
\textbf{Initialize} working memory $\WMem \gets [\bm{e}_1, \bm{e}_2, \dots, \bm{e}_S]$ \\
\textbf{Initialize} procedure banks $\PMB^{+} \gets \{ \ \}, \PMB^{-} \gets \{ \ \}$ \\
pretrained procedure-book $\mathcal{C}$ \\
pretrained LM policy $\pi_\mathrm{LM}$ \\
Environment $env$
\begin{algorithmic}[1]
\ForAll{$\tau \in \mathcal{T}$}
    \State $t \gets 0$
    \State $(o_t, g, \mathrm{DK}) \gets env.\mathrm{reset}(\tau)$
    \State $\mathrm{done} \gets \mathrm{False}$
    \While{\textbf{not} $\mathrm{done}$}
        \Statex \textcolor{black}{\text{\ \ \ \ \ \ \ \ \ \ \ \ \ }\textit{/* LM forward pass */}}
        \State $\HS_0 \gets \mathrm{EmbeddingLayer}(o_t, g, \mathrm{DK})$
        \For{$l = 1$ \textbf{to} $L$}
            \State $(\HS_l, \WMem) \gets \mathrm{DecoderBlock}_l(\HS_{l-1}, \WMem)$ \Comment{\textcolor{black}{with reconstruction of $\PMem$, cf.~Eq.~(10)}}
        \EndFor
        \State $i \gets |\HS_0| + 1$ \Comment{\textcolor{black}{context length $+1$}}
        \State $a_t \gets [ \ ]$
        \While{\textbf{not} $\text{EOS}(x_{<i})$}
            \Statex \textcolor{black}{\text{\ \ \ \ \ \ \ \ \ \ \ \ \ \ \ \ \ \ \ }\textit{/* Contrastive decoding */}}
            \State Obtain $\mathcal{V}_{\mathrm{head}}$ (top-$\vartheta$ head set w.r.t.\ $p^{+}$) \Comment{\textcolor{black}{cf.~Eq.~(11)}}
            \ForAll{$x \in \mathcal{V}_{\mathrm{head}}$}
                % \State $S(x) \gets \log p^{+}(x \mid x_{<i}; R^{+}) - \log p^{-}(x \mid x_{<i}; R^{-})$ \Comment{\textcolor{black}{cf.~Eq.~(12)}}
                \State Compute contrastive score $S(x)$ using $p^{+},p^{-}$ \Comment{\textcolor{black}{cf.~Eq.~(12)}}
            \EndFor
            \State Sample $x_i \sim p_{\mathrm{CP}}(\cdot \mid x_{<i})$ \Comment{\textcolor{black}{cf.~Eq.~(13)}}
            \State $a_t \gets [a_t, x_i]$ \Comment{\textcolor{black}{append token $x_i$}}
            \State $i \gets i + 1$
        \EndWhile
        \State $(o_{t+1}, \mathrm{done}) \gets env.\mathrm{step}(a_t)$
        \State $t \gets t + 1$
    \EndWhile
    \If{task success}
        \State $\PMB^{+} \gets \PMB^{+}\cup\PMem^{+}$
    \Else
        \State $\PMB^{-} \gets \PMB^{-}\cup\PMem^{-}$
    \EndIf
\EndFor
\end{algorithmic}
\end{algorithm}

\subsection{Implementation and Hyperparamter Setting}
We instantiate $\ourmodel$ on a set of backbone LMs and equip it with a memory-augmented module that operates during both training and inference.
During the \textit{Compositional NeSy Procedure Learning} phase, symbolic reference plans are produced by the symbolic planner~\cite{helmert2006fast} and encoded into a procedure-book $\mathcal{C}$ within procedural memory.
The procedure-book $\mathcal{C}$ is composed of $K$ procedure-units of dimension $d$. We set $(K,d)=(256,32)$ for \text{PDDLGym} and $(224,28)$ for \text{VirtualHome} and \text{ALFWorld}.
$\ourmodel$ is trained with AdamW (lr $=2\times10^{-4}$, batch $=1$) for 50 epochs on \text{PDDLGym} and 20 epochs on 
VirtualHome and ALFWorld.

At inference time, during NeSy Procedure Contrastive Planning, the agent retrieves procedures from the positive ($\PMB^+$) and negative ($\PMB^-$) procedure banks based on cosine similarity, using a reconstruction threshold of $\upsilon=0.95$.
Contrastive decoding is applied with a truncation threshold $\vartheta = 0.1$, promoting selection of valid actions while suppressing failure patterns. 
Generation is performed using deterministic decoding with top-$p = 1.0$ and temperature 0.0.

All experiments run on Python 3.10, PyTorch 2.3.0, and Transformers 4.47.1 with an Intel i9-10980XE CPU and a single NVIDIA RTX A6000 GPU.
A concise summary of the hyperparameter settings is provided in Table~\ref{tab:impl}.

\begin{table}[htbp]
\centering
\caption{Hyperparamer settings for $\ourmodel$}
\label{tab:impl}
\begin{tabular}{lccc}
\toprule
\textbf{Setting} & \textbf{PDDLGym} & \textbf{VirtualHome} & \textbf{ALFWorld} \\ 
\midrule
Procedure-book size $K$ & 256 & 224 & 224 \\
Procedure-unit dim.\ $d$ & 32 & 28 & 28 \\
Training epochs & 50 & 20 & 20 \\
Optimizer / lr & \multicolumn{3}{c}{AdamW / $2\times10^{-4}$} \\
Batch size & \multicolumn{3}{c}{1} \\
EMA decay ($\mathcal{C}$) & \multicolumn{3}{c}{0.99} \\
Reconstruction threshold $\upsilon$ & \multicolumn{3}{c}{0.95} \\
Contrastive trunc.\ $\vartheta$ & \multicolumn{3}{c}{0.10} \\
Decoding top-$p$ / temp. & \multicolumn{3}{c}{1.0 / 0.0} \\
\bottomrule
\end{tabular}
\end{table}

%%%%%%%%%%%%%%%%%%%%%%%%%%%%%%%%%%%%%%%%%%%%%%%%%%%%%%%%%%%%%%%%%%%%%%%%%%%%%%%
%%%%%%%%%%%%%%%%%%%%%%%%%%%%%%%%%%%%%%%%%%%%%%%%%%%%%%%%%%%%%%%%%%%%%%%%%%%%%%%
\newpage
\section{Evaluation}\label{app:sec:evaluation}
\subsection{Experiment Setting}\label{app:sec:evaluation:exp_setting}
We evaluate $\ourmodel$ on standard embodied benchmarks: three PDDLGym~\cite{benchmark:pddlgym} domains (e.g., Minecraft, Rearrangement, GlibRearrangement), VirtualHome~\cite{li2024embodied}, and ALFWorld~\cite{benchmark:alfw}.
Each benchmark is converted to a continual task setting in which the agent solves a sequence of tasks that require multi-step planning and continual adaptation. 
Tasks are provided in symbolic form, consisting of the current observation, the goal, and domain knowledge that specifies the action set and transition dynamics. Input examples are shown below.

\definecolor{darkgray}{rgb}{0.2,0.2,0.2}
\definecolor{lightgray}{rgb}{0.95,0.95,0.95}
\begin{tcolorbox}[
    colback=lightgray,
    colframe=darkgray,
    coltitle=white,
    title=Symbolic Input: problem instance,
    fonttitle=\bfseries,
    arc=1mm,
    breakable
]
\color{black}
(:observation \\ 
\hspace*{1.5em}obj\_next\_to(kitchen\_counter:object,dish\_soap:object) \\ 
\hspace*{1.5em}obj\_next\_to(plate:object,kitchen\_counter:object) \\ 
\hspace*{1.5em}obj\_next\_to(microwave:object,kitchen\_counter:object) \\ 
\hspace*{1.5em}$\vdots$ \\
\hspace*{1.5em}plugged\_out(dishwasher:object) \\ 
\hspace*{1.5em}off(dishwasher:object) \\
\hspace*{1.5em}clean(dishwasher:object) \\\
) \\
~\\
(:goal \\ 
\hspace*{1.5em}(and \\ 
\hspace*{3em}(closed dishwasher) \\ 
\hspace*{3em}(on dishwasher) \\ 
\hspace*{3em}(obj\_ontop dish\_soap dishwasher) \\ 
\hspace*{3em}(obj\_ontop plate dishwasher) \\ 
\hspace*{1.5em}) \\
)
\end{tcolorbox}

\definecolor{darkgray}{rgb}{0.2,0.2,0.2}
\definecolor{lightgray}{rgb}{0.95,0.95,0.95}
\begin{tcolorbox}[
    colback=lightgray,
    colframe=darkgray,
    coltitle=white,
    title=Symbolic Input: domain knowledge,
    fonttitle=\bfseries,
    arc=1mm,
    breakable
]
\color{black}
(:action walk\_towards \\ 
\hspace*{1.5em}:parameters (?char - character ?obj - object) \\
\hspace*{1.5em}:precondition (and \\ 
\hspace*{7.1em}(not (sitting ?char)) \\ 
\hspace*{7.1em}(not (lying ?char)) \\ 
\hspace*{1.5em}) \\ 
\hspace*{1.5em}:effect (and \\ 
\hspace*{4.3em}(next\_to ?char ?obj) \\ 
\hspace*{4.3em}(forall (?far\_obj - object) \\ 
\hspace*{5.8em}(when (not (obj\_next\_to ?far\_obj ?obj)) (not (next\_to ?char ?far\_obj))) \\ 
\hspace*{4.3em}) \\ 
\hspace*{4.3em}(forall (?close\_obj - object) \\ 
\hspace*{5.8em}(when (obj\_next\_to ?close\_obj ?obj) (next\_to ?char ?close\_obj)) \\ 
\hspace*{4.3em}) \\ 
\hspace*{4.3em}) \\ 
) \\ 
$\vdots$ \\
(:action plug\_in \\ 
\hspace*{1.5em}:parameters (?char - character ?obj - object) \\ 
\hspace*{1.5em}:precondition (or \\ 
\hspace*{8.6em}(and \\ 
\hspace*{8.6em}(next\_to ?char ?obj) \\ 
\hspace*{8.6em}(has\_plug ?obj) \\ 
\hspace*{8.6em}(plugged\_out ?obj) \\ 
\hspace*{8.6em}) \\ 
\hspace*{8.6em}(and \\ 
\hspace*{8.6em}(next\_to ?char ?obj) \\ 
\hspace*{8.6em}(has\_switch ?obj) \\ 
\hspace*{8.6em}(plugged\_out ?obj) \\ 
\hspace*{8.6em}) \\ 
\hspace*{1.5em}) \\ 
\hspace*{1.5em}:effect (and \\ 
\hspace*{4.3em}(plugged\_in ?obj) \\ 
\hspace*{4.3em}(not (plugged\_out ?obj)) \\ 
\hspace*{1.5em}) \\ 
) \\ 
~ \\ 
(:predicates \\ 
\hspace*{1.5em}(closed ?obj - object) \\ 
\hspace*{1.5em}(open ?obj - object) \\ 
\hspace*{1.5em}(on ?obj - object) \\ 
\hspace*{1.5em}$\vdots$ \\
\hspace*{1.5em}(surfaces ?obj - object) \\ 
\hspace*{1.5em}(sittable ?obj - object) \\ 
\hspace*{1.5em}(lieable ?obj - object) \\ 
)
\end{tcolorbox}

\newpage
\subsubsection{PDDLGym}
% PDDLGym 설명
PDDLGym~\cite{benchmark:pddlgym} is an embodied task benchmark that converts classical-planning problems expressed in PDDL~\cite{pddl} into OpenAI Gym environments~\cite{openai}, providing a unified interface across more than twenty relational domains, such as Blocksworld, Sokoban, Minecraft-Crafting, and various object-rearrangement tasks.
In each environment, observations are represented as sets of first-order predicates, and actions are defined by parameterized operator schemas. Solving tasks in such settings requires object-centric reasoning, long-horizon planning, and symbol grounding—core capabilities for embodied agents.
Because every problem instance is specified declaratively, it can be exported directly to symbolic planners or grounded in low-level simulators, making it easy to compare planning and learning methods on exactly the same set of tasks. 
As a result, PDDLGym has become a standard benchmark for neurosymbolic planning, relational reinforcement learning, and LM–based embodied decision making, enabling side-by-side comparisons of symbolic planners, policy learners, and hybrid approaches within exactly the same suite of environments.

\begin{figure}[htbp]
    \centering
    \subfigure[Example of ``(:goal (and (inventory new-0) (agentat loc-0-3) ))'']
    {
        \centering
        \includegraphics[width=0.32\linewidth]{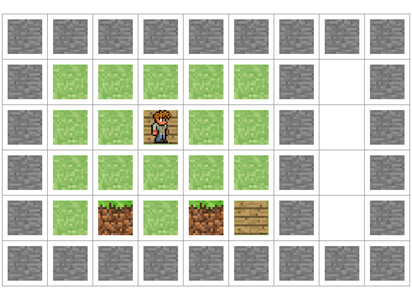}
        \includegraphics[width=0.32\linewidth]{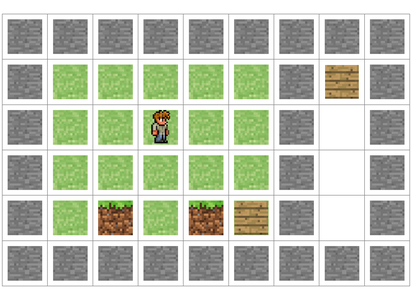}
        \includegraphics[width=0.32\linewidth]{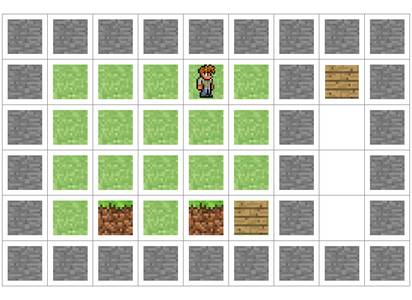}
        
    }
    % \hspace{10pt}
    \centering
    \subfigure[Example of ``(:goal (and (equipped grass-2 agent) (inventory log-3) ))'']
    {
        \centering
        \includegraphics[width=0.24\linewidth]{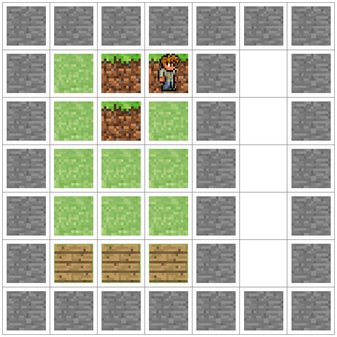}
        \includegraphics[width=0.24\linewidth]{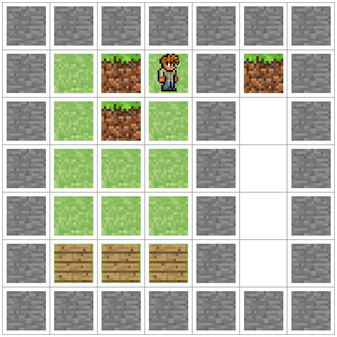}
        \includegraphics[width=0.24\linewidth]{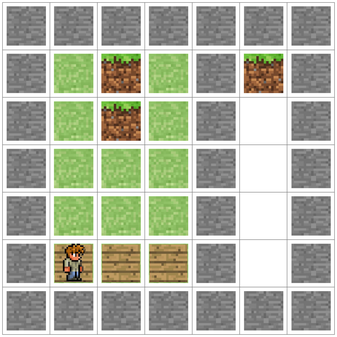}
        \includegraphics[width=0.24\linewidth]{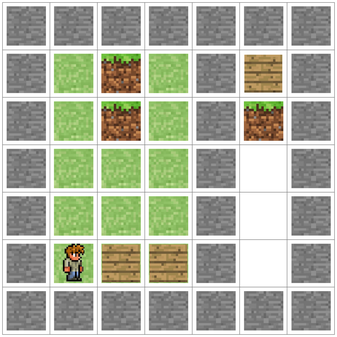}
    }
    \caption{Illustrative examples of tasks in the Minecraft domain}
    \label{app:fig:ben:minecraft}
\end{figure}

% 우리가 사용한 세팅들
We conduct our experiments on the Minecraft, Rearrangement, and GlibRearrangement domains from PDDLGym.
% 각 도메인 차이점 언급
The Minecraft is a domain focused on object manipulation and crafting, where the agent must gather raw materials and perform a sequence of actions to synthesize target items.
The Rearrangement involves moving specific objects to designated locations, requiring spatial reasoning and sequential planning in structured environments.
The GlibRearrangement extends Rearrangement by introducing additional relational constraints, such as requiring the agent to hold certain objects, thereby increasing task complexity.
While the overall goal remains to reposition specific objects, the agent must now account for more intricate action preconditions and inter-object relations during planning.
These domains in PDDLGym include 430, 420, and 120 problem instances, respectively, most of which can be verified using symbolic planners~\cite{helmert2006fast, hoffmann2001ff}, making them suitable for our evaluation. We construct nine distinct task sequences in PDDLGym for continual tasks by randomly composing tasks under a shared evaluation setting.
% 
% 예시들
Examples of task and action types for the Minecraft domain are summarized in Table~\ref{app:tab:ben:minecraft}, with corresponding visualizations shown in Figure~\ref{app:fig:ben:minecraft}.

\begin{table}[htbp]
\caption{Examples of tasks and actions in the Minecraft domain}
\label{app:tab:ben:minecraft}
\begin{center}
\begin{adjustbox}{width=0.99\linewidth}
\begin{tabular}{l ll}
    \toprule
    & \textbf{Type} & \textbf{Example} \\
    \midrule
    \multirow{6}{*}{Tasks}  
    & Move \& Equip & (:goal (and  (equipped new-2 agent)  (equipped grass-0 agent) )) \\
    & Collect \& Move & (:goal (and  (equipped grass-2 agent)  (agentat loc-3-1) )) \\
    & Craft \& Equip & (:goal (and  (equipped new-2 agent)  (isplanks new-2) )) \\
    & Move \& Inventory & (:goal (and  (inventory new-0)  (agentat loc-0-3) )) \\
    & Equip \& Inventory & (:goal (and  (equipped new-2 agent)  (inventory grass-2) )) \\
    & Craft \& Inventory & (:goal (and  (inventory grass-0)  (isplanks new-0) )) \\
    \midrule
    \multirow{5}{*}{Actions}
    & (\textbf{recall} ?var0 - moveable ?var1 - agent) & recall (grass-0:moveable, agent:agent) \\
    & (\textbf{move} ?var0 - static ?var1 - static) & move (loc-2-0:static, loc-2-1:static) \\
    & (\textbf{craftPlank} ?var0 - moveable ?var1 - agent ?var2 - moveable) & craftplank (new-0:moveable, agent:agent, log-2:moveable) \\
    & (\textbf{equip} ?var0 - moveable ?var1 - agent) & equip (log-2:moveable, agent:agent) \\
    & (\textbf{pick} ?var0 - moveable ?var1 - static) & pick (grass-0:moveable, loc-0-1:static) \\
    \bottomrule
\end{tabular}
\end{adjustbox}
\end{center}
\end{table}

%Craft a new item, put it in your inventory, and move to location (0,3).

\noindent\textbf{Datasets.}
We split the full set of problem instances for each domain into separate train and test datasets.
The train sets consist of 29 instances for the Minecraft, 20 for the Rearrangement, and 40 for the GlibRearrangement.
The corresponding test sets contain 389, 400, and 80 instances, respectively, with no overlap with the training data.
For training, we use a relatively small number of problem instances, each paired with a reference plan generated by a symbolic planner.
For evaluation, we filter the test sets to include only solvable instances, which are verified in advance, and do not provide reference plans at test time.

\newpage
\subsubsection{VirtualHome}
% virtualhome 기본 설명
VirtualHome~\cite{benchmark:vh} is an embodied simulation platform designed to replicate real-life household activities. In this environment, agents must interact with their surroundings using high-level actions to accomplish diverse and complex tasks, such as "Make coffee" or "Wash dishes with the dishwasher", making VirtualHome a well-suited benchmark for evaluating the adaptive and structured reasoning capabilities of LM-based agents.

\begin{figure}[ht]
    \centering
    \subfigure[Example of scene 0]{
        \centering
        \includegraphics[width=0.46\linewidth]{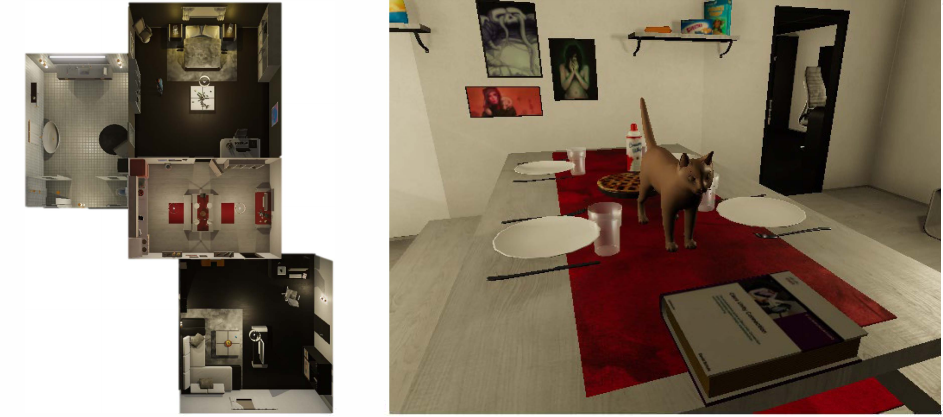}
    }
    \hfill
    \subfigure[Example of scene 1]{
        \centering
        \includegraphics[width=0.46\linewidth]{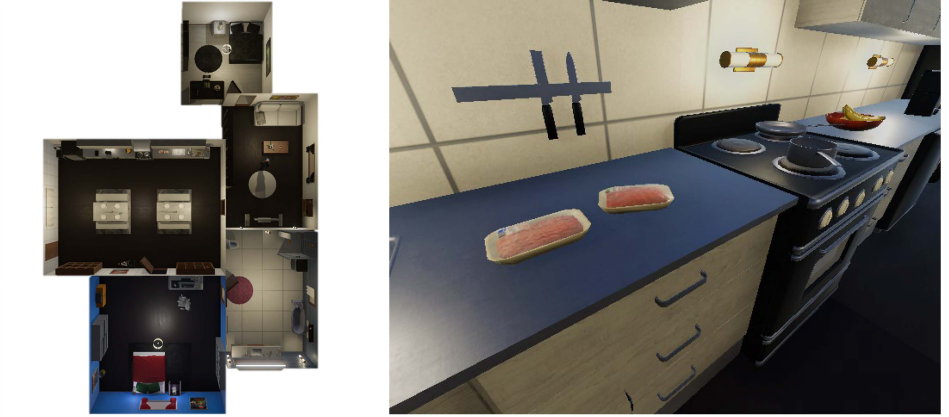}
    }
    
    \centering
    \subfigure[Example of scene 2]{
        \centering
        \includegraphics[width=0.46\linewidth]{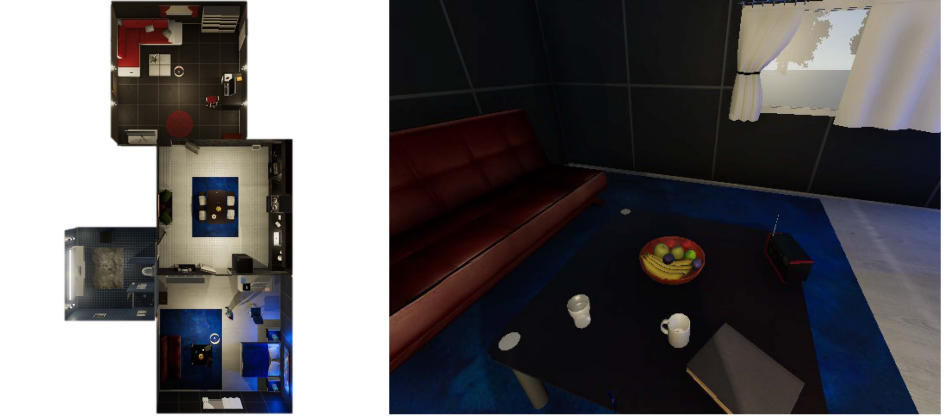}
    }
    \hfill
    \subfigure[Example of scene 3]{
        \centering
        \includegraphics[width=0.46\linewidth]{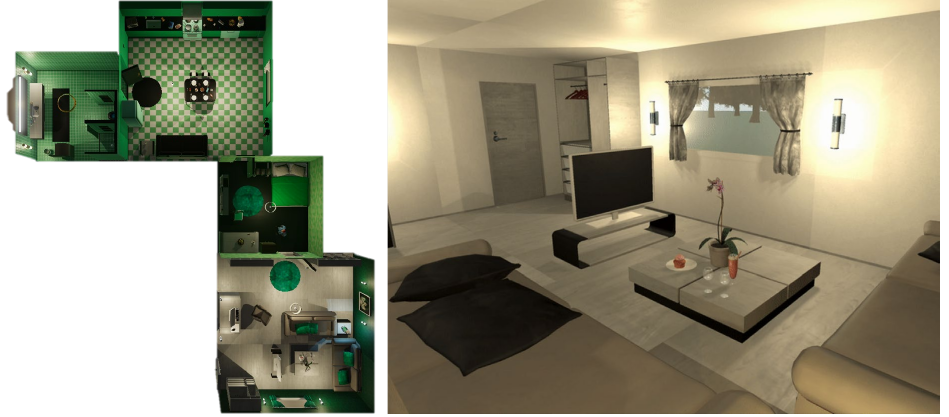}
    }
    \caption{Illustrative examples of scenes in VirtualHome}
    \label{app:fig:ben:vh}
\end{figure}

% 우리가 사용한 세팅
For our experiments, we build upon the open-source Embodied Agent Interface\cite{li2024embodied}, which provides a domain file and problem files for 338 distinct tasks in VirtualHome. We implement the environment by simulating the files using TextWorld\cite{cote2019textworld}, resulting in VirtualHome environment where the observations and actions are represented in PDDL format. To evaluate agents in continual task settings, we follow \cite{clalfred}, constructing two configurations: behavior-incremental and environment-incremental. In each configuration, tasks are grouped into four semantically similar sets based on their respective incremental properties. Agents are then evaluated across four distinctly ordered sequences of these task sets to assess their robustness to task order variations.
These sequences are constructed for both seen and unseen settings. In the seen setting, tasks share the same types as those in the training set, differing only in object placement. In contrast, the unseen setting consists of entirely novel task types, requiring the agent to adaptively reason and generalize based on previously acquired knowledge.
% 
% table 및 그림 좌표
Details of the tasks and actions are presented in Table~\ref{app:tab:ben:vh}, and visualizations of various indoor scenes are depicted in Figure~\ref{app:fig:ben:vh}.

\begin{table}[ht]
\caption{Examples of tasks and actions in VirtualHome}
\label{app:tab:ben:vh}
\begin{center}
\begin{small}
\begin{adjustbox}{width=0.99\linewidth}
\begin{tabular}{l ll}
\toprule
& \textbf{Type} & \textbf{Example} \\
\midrule
\multirow{6}{*}{Tasks} 
                        & Plug \& Switch & (:goal (and (closed cd\_player) (plugged\_in cd\_player) (on cd\_player) )) \\
                        & Hold & (:goal (and (holds\_lh character tooth\_paste) (holds\_rh character toothbrush) )) \\
                        & Sit/Lie & (:goal (and (lying character) (ontop character bed) )) \\ 
                        & Be next to & (:goal (and (next\_to character shower) )) \\
                        & Object on & (:goal (and (obj\_ontop plate table) )) \\ 
                        & Object in & (:goal (and (open freezer) (plugged\_in freezer) (obj\_inside food\_food freezer) )) \\
                        
\midrule
\multirow{14}{*}{Actions} 
                        & (\textbf{walk\_towards} ?char - character ?obj - object) & walk\_towards (character:character, computer:object) \\ 
                        & (\textbf{walk\_into} ?char - character ?room - object) & walk\_into (character:character, bathroom:object) \\
                        & (\textbf{sit} ?char - character ?obj - object) & sit (character:character, couch:object) \\
                        & (\textbf{standup} ?char - character) & standup (character:character) \\
                        & (\textbf{grab} ?char - character ?obj - object) & grab (character:character, plate:object) \\
                        & (\textbf{open} ?char - character ?obj - object) & open (character:character, dishwasher:object) \\
                        & (\textbf{close} ?char - character ?obj - object) & close (character:character, dishwasher:object) \\
                        & (\textbf{put\_on} ?char - character ?obj1 - object ?obj2 - object) & put\_on (character:character, ground\_coffee:object, coffee\_maker:object) \\
                        & (\textbf{put\_inside} ?char - character ?obj1 - object ?obj2 - object) & put\_inside (character:character, plate:object, kitchen\_cabinet:object) \\ 
                        & (\textbf{switch\_on} ?char - character ?obj - object) & switch\_on (character:character, computer:object) \\
                        & (\textbf{turn\_to} ?char - character ?obj - object) & turn\_to (character:character, toilet:object) \\
                        & (\textbf{lie} ?char - character ?obj - object) & lie (character:character, bed:object) \\
                        & (\textbf{plug\_in} ?char - character ?obj - object) &  plug\_in (character:character, dishwasher:object) \\
\bottomrule

\end{tabular}
\end{adjustbox}
\end{small}
\end{center}
\end{table}

\noindent\textbf{Datasets.}
We split the full set of problem instances into separate train and test sets, with the test set further divided into seen and unseen subsets.
The train set contains 77 instances, each paired with an optimal plan generated by a symbolic planner~\cite{helmert2006fast}.
The seen set includes 112 instances that share the same goals as those in the train set but differ in object placement and inter-object relations.
The unseen subset consists of 52 entirely novel tasks that do not appear in either the train or seen sets.
At test time, agents are given only the current observation and goal, with no access to symbolic tools or reference plans.

\newpage
\subsubsection{ALFWorld}
% alfworld 기본 설명
ALFWorld\cite{benchmark:alfw} is TextWorld-based embodied benchmark built upon the tasks and interaction capabilities of ALFRED benchmark\cite{shridhar2020alfred}. This household environment is designed to train and evaluate agents on action planning under long and compositional task instructions, making it a challenging benchmark for embodied agents.

% 우리가 사용한 세팅
In ALFWorld, a total of 3,554 task instances are available, each provided with its respective problem file. To enable evaluation on ALFWorld under symbolic settings, we modify the environment to provide observations and accept actions in PDDL format, similar to our setup in VirtualHome. All other aspects of the evaluation framework remain consistent with those used in VirtualHome  experiments, including the presence of behavior- and environment-incremental configurations, as well as the division of test set into seen and unseen sets.
% 
% table 및 그림 좌표
Details of the tasks and actions are presented in Table~\ref{app:tab:ben:alfw}.

\begin{table}[ht]
\caption{Examples of tasks and actions in ALFWorld}
\label{app:tab:ben:alfw}
\begin{center}
\begin{small}
\begin{adjustbox}{width=0.99\linewidth}
\begin{tabular}{l ll}
\toprule
& \textbf{Type} & \textbf{Example} \\
\midrule
\multirow{14}{*}{Tasks} 
                        & \multirow[]{2}{*}{Heat} 
                          & (:goal (and (exists (?r - receptacle) (exists (?o - object) (and (heatable ?o) (objecttype ?o eggtype)\\ 
                          & & (receptacletype ?r garbagecantype) (ishot ?o) (inreceptacle ?o ?r) ))))) \\
                        & \multirow[]{2}{*}{Cool} 
                          & \rule{0pt}{12pt}(:goal (and (exists (?r - receptacle) (exists (?o - object) (and (coolable ?o) (objecttype ?o lettucetype) \\ 
                          & & (receptacletype ?r countertoptype) (iscool ?o) (inreceptacle ?o ?r) ))))) \\
                        & \multirow[]{2}{*}{Clean} 
                          & \rule{0pt}{12pt}(:goal (and (exists (?r - receptacle) (exists (?o - object) (and (cleanable ?o) (objecttype ?o spatulatype)  \\ 
                          & & (receptacletype ?r diningtabletype) (isclean ?o) (inreceptacle ?o ?r) ))))) \\
                        & \multirow[]{2}{*}{Pick \& Place} 
                          & \rule{0pt}{12pt}(:goal (and (exists (?r - receptacle) (exists (?o - object) (and (inreceptacle ?o ?r) \\ 
                          & & (objecttype ?o soapbottletype) (receptacletype ?r countertoptype) ))))) \\
                        & \multirow[]{3}{*}{Pick2 \& Place} 
                          & \rule{0pt}{12pt}(:goal (and (exists (?r - receptacle) (exists (?o1 - object) (and (objecttype ?o1 vasetype) \\ 
                          & & (receptacletype ?r desktype) (inreceptacle ?o1 ?r) (exists (?o2 - object) (and (not (= ?o1 ?o2)) \\
                          & & (objecttype ?o2 vasetype) (receptacletype ?r desktype) (inreceptacle ?o2 ?r) ))))))) \\
                        & \multirow[]{3}{*}{Examine} 
                          & \rule{0pt}{12pt}(:goal (and (exists (?ot - object r - receptacle ?a - agent ?l - location) (and (objecttype ?ot desklamptype) \\
                          & & (toggleable ?ot) (istoggled ?ot) (receptacleatlocation ?r ?l) (atlocation ?a ?l) (inreceptacle ?ot ?r))) \\
                          & & (exists (?o - object ?a - agent) (and (objecttype ?o booktype) (holds ?a ?o) )))) \\

\midrule
\multirow{8}{*}{Actions} 
                        & (\textbf{gotolocation} ?a - agent ?lstart - location ?lend - location ?r - receptacle) & gotolocation (agent1:agent, loc\_5:location, loc\_12:location, countertop\_1:receptacle) \\ 
                        & (\textbf{pickupobject} ?a - agent ?l - location ?o - object ?r - receptacle) & pickupobject (agent1:agent, loc\_21:location, tomato\_1:object, countertop\_1:receptacle) \\
                        % & (\textbf{putobject} ?a - agent ?l - location ?o - object ?r - receptacle ?ot - otype ?rt - rtype) & putobject (agent1:agent, loc\_8:location, tomato\_1:object, sidetable\_1:receptacle, tomatotype:ot, sidetabletype:rt) \\
                        & (\textbf{openobject} ?a - agent ?l - location ?r - receptacle) & openobject (agent1:agent, loc\_2:location, cabinet\_1:receptacle) \\
                        & (\textbf{heatobject} ?a - agent ?l - location ?r - receptacle ?o - object) & heatobject (agent1:agent, loc\_1:location, microwave\_1:receptacle, tomato\_1:object) \\
                        & (\textbf{coolobject} ?a - agent ?l - location ?r - receptacle ?o - object) & coolobject (agent1:agent, loc\_4:location, fridge\_1:receptacle, tomato\_1:object) \\
                        & (\textbf{cleanobject} ?a - agent ?l - location ?r - receptacle ?o - object) & cleanobject (agent1:agent, loc\_7:location, sink\_1:receptacle, cup\_4:object) \\
                        & (\textbf{toggleobject} ?a - agent ?l - location ?o - object ?r - receptacle) & toggleobject (agent1:agent, loc\_4:location, desklamp\_1:object, desk\_1:receptacle) \\
\bottomrule

\end{tabular}
\end{adjustbox}
\end{small}
\end{center}
\end{table}

\begin{figure}[htbp]
    \centering
    \subfigure[Example of bedroom type scene]{
        \centering
        \includegraphics[width=0.46\linewidth]{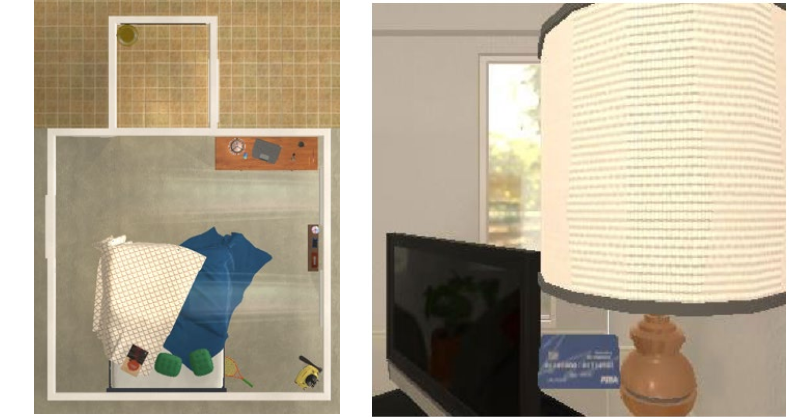}
    }
    \hfill
    \subfigure[Example of bathroom type scene]{
        \centering
        \includegraphics[width=0.46\linewidth]{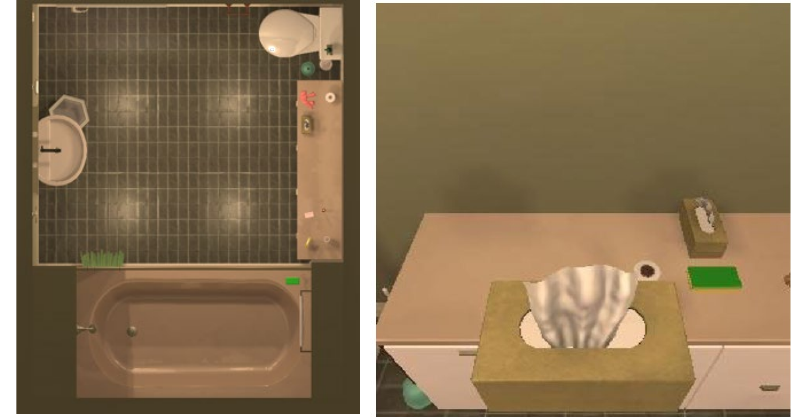}
    }
    
    \subfigure[Example of livingroom type scene]{
        \centering
        \includegraphics[width=0.46\linewidth]{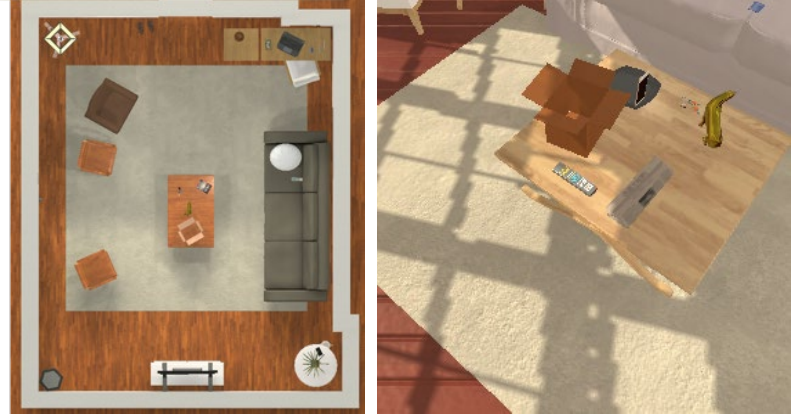}
    }
    \hfill
    \subfigure[Example of kitchen type scene]{
        \centering
        \includegraphics[width=0.46\linewidth]{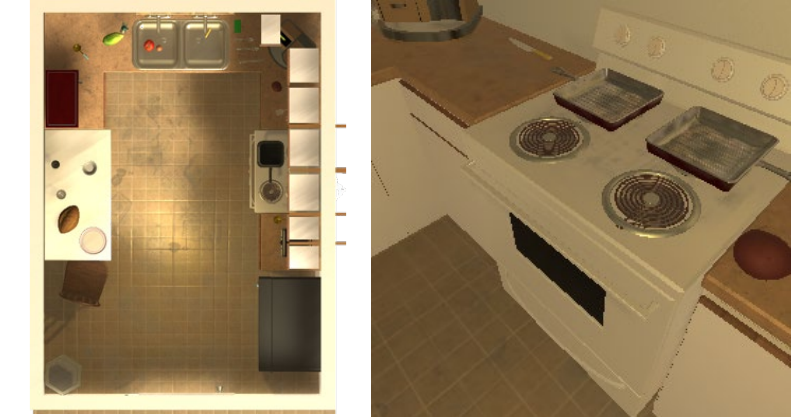}
    }
    \caption{Illustrative examples of scenes ALFWorld}
    \label{app:fig:ben:alfworld}
\end{figure}

\noindent\textbf{Datasets.}
Similar to VirtualHome, we partition the full set of problem instances into separate train and test sets, with the test set further divided into seen and unseen sets.
The train set comprises 549 instances, each paired with an optimal plan generated by a symbolic planner~\cite{helmert2006fast}.
The seen set includes 1,509 instances that share goals with the train set but vary in object placement and inter-object relations.
The unseen set contains 1,369 entirely novel tasks not present in either the train or seen sets.
At test time, agents receive only the current observation and goal, without access to symbolic tools or reference plans.

\newpage
\subsubsection{Baselines}\label{baselines}
For comparison, we organize the baselines into four categories:

\begin{itemize}[leftmargin=*, itemsep=0pt, topsep=0pt] % noitemsep
    \item Single-step planning approaches generate the entire action sequence in a single, unified inference step, without intermediate reasoning or iterative refinement.
    \begin{itemize}[leftmargin=*, itemsep=0pt, topsep=0pt] % noitemsep
        \item ZSP~\cite{llmagent:zsp} leverages pretrained LMs to translate high-level natural language instructions into executable action sequences for embodied agents in a zero-shot manner.
        \item RAP~\cite{llmagent:rap} enables LM-based agents to improve decision-making by retrieving and leveraging past experiences stored in contextual memory.
        \item LLM-Planner~\cite{llmagent:llmplanner} leverages the few-shot in-context learning capabilities of LMs to aid action planning. By retrieving demonstrations relevant to the current observation throughout the episode, it enables more grounded and context-aware planning.
    \end{itemize}
    \item Agentic workflows perform multi-step reasoning through multiple LM calls, allowing for self-refinement and iterative decision-making.
    \begin{itemize}[leftmargin=*, itemsep=0pt, topsep=0pt] % noitemsep
        \item CoT~\cite{cot:cot} enables LMs to solve complex reasoning tasks by generating intermediate reasoning steps, leading to improved performance on arithmetic, commonsense, and symbolic reasoning benchmarks.
        \item ToT~\cite{cot:tot} empowers LMs to perform deliberate problem-solving by exploring and evaluating multiple reasoning paths, significantly improving performance on complex tasks.
        \item GoT~\cite{cot:got} enhances LMs reasoning by structuring intermediate thoughts as a graph, allowing for advanced operations like feedback loops and aggregation, leading to improved performance on complex tasks.
        \item ReAct~\cite{llmagent:react} interleaves reasoning steps between the interactions with the environment, enabling more adaptive decision-making based on rationales generated from interpreting the current observation.
        \item Reflexion~\cite{llmagent:reflexion} incorporates verbal feedback from failed experiences to guide subsequent decision-making, enabling the model to learn from experiences without parameter updates.
    \end{itemize}
    \item Memory-augmented LM
    \begin{itemize}[leftmargin=*, itemsep=0pt, topsep=0pt] % noitemsep
        \item LongMem~\cite{llmmem:longmem} enhances LMs by incorporating a decoupled memory module, enabling the models to store and retrieve extended context information, thereby improving performance on tasks requiring long-term dependencies.
        \item LMM~\cite{llmmem:lmm} augments the Transformer~\cite{attn:causal} architecture with an auxiliary memory module to improve multi-step reasoning and long-term context modeling.
        \item Optimus-2~\cite{memagent:optimus2} introduces a memory-augmented agent for embodied task planning in open-world environment, enabling long-horizon action generation by conditioning on goal, observation, and action history.
        \item DT-Mem~\cite{memagent:think} enhances generalization and adaptability of reinforcement learning agents by storing and reusing knowledge across tasks through an internal working memory module.
        \item BUTLER~\cite{butler} fine-tunes a pretrained LMs using a small number of expert demonstrations and limited environment interactions to enable effective action planning in text-based environments. Although it does not incorporate memory mechanisms, it serves as a learning-based baseline for evaluating memory-augmented LM-based embodied agents.
    \end{itemize}
    \item Proceduralization
    \begin{itemize}[leftmargin=*, itemsep=0pt, topsep=0pt] % noitemsep
        \item BoT~\cite{proc:bot} stores abstract thought templates in a meta-buffer and dynamically retrieves and instantiates them to guide procedural LM reasoning.
        \item LRM~\cite{proc:lrm} internalizes CoT reasoning through reinforcement learning, and also offers compact student variants distilled from larger teacher models.
        \item PlaSma~\cite{proc:plasma} equips smaller LMs with procedural knowledge and planning capabilities by distilling knowledge from larger LMs and employing a verifier-guided decoding algorithm, enabling them to achieve performance comparable to their larger counterparts.
    \end{itemize}
\end{itemize}

\noindent\textbf{ZSP.}
This work investigates whether a pretrained LMs can convert high-level natural language instructions into executable action sequences without additional training. 
The entire action plan is generated in a single inference step and remains fixed, without incorporating feedback from the environment or adapting to changes in state. 

% implementation
We refer to the publicly available implementation~\footnote{https://github.com/huangwl18/language-planner}. 
We follow the default prompt templates described in the paper and codes, with minor modifications to adapt to the observation-goal input format in our benchmark environments. 
% hyperparameters
For the planning LM, we use LLaMA3.2-1B~\cite{llama3.2} with a temperature of 0.0 and a maximum output length of 128 tokens for PDDLGym.
For the translation LM, we use sentence embedding model (i.e., all-mpnet-base-v2) from Sentence-Former~\cite{reimers-2019-sentence-bert} to retrieve the top-$k$ most similar instructions along with their associated symbolic actions. 
As shown in Table~\ref{app:hyper:llm_planner}, input prompts include five in-context examples sampled from the training set. Generation is performed using deterministic decoding with top-$p = 1.0$.

\begin{table}[htbp]   
        \centering
        \caption{Hyperparamer settings for ZSP}
        \label{app:hyper:zsp}
        \begin{tabular}{l c}
        \toprule
        \textbf{Hyperparameters} & \textbf{Value} \\
        \midrule
        In-context samples    &  5 \\
        Decoding & Greedy \\ 
        Maximum new tokens & 128 \\
        \bottomrule
    \end{tabular}
\end{table}

\noindent\textbf{RAP.}
RAP addresses the challenge of enabling LM agents to utilize past experiences in current decision-making processes, a behavior innate to humans. The framework stores past experiences in a memory module and retrieves relevant information based on the similarity to the current context. This retrieved information is then used to inform the agent's planning through in-context learning.

Here, we extend ZSP by incorporating the contextual memory concept from RAP. Specifically, we enable past successful experiences to be dynamically composed as in-context examples based on the current task context. This allows the agent to adaptively retrieve and organize relevant demonstrations as the memory grows. We build on the publicly available implementation of RAP~\footnote{https://github.com/PanasonicConnect/rap}. All other components, including planning and retrieval, remain consistent with ZSP.

\noindent\textbf{LLM-Planner.} 
LLM-Planner augments ZSP with few-shot demonstrations, leveraging the in-context learning capabilities of LMs to better ground the agent in the environment. It also incorporates a replanning mechanism that enables the agent to recover from infeasible plans.

We implement referring to the official implementation~\footnote{https://github.com/OSU-NLP-Group/LLM-Planner}. At each timestep, the top-k most relevant demonstrations are dynamically retrieved from train set based on the current observation and goal. Candidates are first filtered using cosine similarity with the goal, followed by selection based on Jaccard similarity over current observation predicates.
For replanning, LLM-Planner is implemented to reset the action history when three consecutive non-executable actions are generated. The hyperparameter settings for LLM-Planner are provided in Table~\ref{app:hyper:llm_planner}. By default, we use Qwen2.5-0.5B~\cite{qwen2.5} for both VirtualHome and ALFWorld.

\begin{table}[htbp]   
        \centering
        \caption{Hyperparamer settings for LLM-Planner}
        \label{app:hyper:llm_planner}
        \begin{tabular}{l c}
        \toprule
        \textbf{Hyperparameters} & \textbf{Value} \\
        \midrule
        In-context samples    &  3 \\
        Decoding & Greedy \\ 
        Maximum new tokens & 120 \\
        \bottomrule
    \end{tabular}
\end{table}

\noindent\textbf{CoT.} CoT prompting improves the reasoning ability of LMs by presenting step-by-step exemplars that guide the model through intermediate reasoning steps, leading to better performance on complex, multi-step tasks.

Our implementation of CoT is used as a baseline for PDDLGym, built upon the RAP framework and extended with additional rationales to guide symbolic reasoning. We base our implementation on the publicly available GoT repository~\footnote{https://github.com/spcl/graph-of-thoughts} which includes CoT as part of its baseline methods.
Specifically, we augment the train set with step-by-step rationales that reflect key features of the underlying MDP structure. The rationale design follows the approach proposed in \cite{llmagent:deder}, using the problem instances and corresponding plans in the train set to provide structured, interpretable reasoning traces. All other hyperparameters follow the same configuration as used in ZSP and RAP.

\noindent\textbf{ToT.} ToT prompting extends the chain-of-thought (CoT) framework by enabling exploration of multiple reasoning paths instead of committing to a single linear sequence. At each step, the model generates and evaluates multiple candidate thoughts, forming a tree structure that allows for backtracking and selection of the most promising reasoning trajectory. This approach improves performance on tasks that benefit from deliberation, such as mathematical reasoning and complex decision making. 

Our implementation of ToT builds on the same additionally designed rationales used in CoT, and leverages the ToT implementation provided in the GoT repository. To enable tree-structured search, we adapt the branching, self-evaluation, and selection prompts to align with the characteristics of PDDLGym tasks, setting the maximum branch factor to 5. All other hyperparameters follow the same configuration as used in ZSP and RAP.

\noindent\textbf{GoT.} GoT generalizes the tree-based reasoning structure of ToT by allowing thoughts to be organized as arbitrary graphs rather than strict trees. This enables the model to support richer reasoning patterns such as merging, revisiting, and cross-referencing intermediate thoughts. Unlike CoT and ToT, which follow fixed linear or hierarchical paths, GoT allows for flexible traversal and aggregation of reasoning substructures, making it well-suited for tasks that require integration of multiple partial solutions or iterative refinement.

Our implementation of GoT is based on the official repository, with additional modifications to the aggregation and refinement prompts to better suit the structure of PDDLGym tasks. The branch factor is set to a maximum of 5. All other hyperparameters follow the same configuration as used in ZSP and RAP.

\noindent\textbf{ReAct.} ReAct interleaves reasoning steps between observation inputs and decision-making, enabling the agent to adaptively solve tasks in the environment. The generated reasoning may involve goal analysis or interpretation of the current observation, thereby aiding the production of appropriate actions.

We refer to official repository~\footnote{https://github.com/ysymyth/ReAct} for implementation. The few-shot demonstrations annotated with reasoning steps are provided as in-context examples to guide the LM's behavior. These annotations are generated by prompting the LLaMa3.1-8B model to produce rationales based on the current observation and goal. Demonstration retrieval follows the same procedure as in LLM-Planner. During inference, the model generates the reasoning step first, followed by the corresponding action.
ReAct adopts the same hyperparameter settings as LLM-Planner, as shown in Table~\ref{app:hyper:llm_planner}. By default, we use Qwen2.5-0.5B for both VirtualHome
and ALFWorld.

\noindent\textbf{Reflexion.} Reflexion is an LM-based agent, similar to ReAct, that alternates between reasoning and acting. It additionally introduces a self-reflection mechanism, generating feedback by analyzing failed trajectories and using this feedback to better guide subsequent attempts.

The original framework incorporates extracted feedback by appending it into the prompt, encouraging the LM to consider it during generation. In our setting, we incorporate feedback by adding the failed action to the "bad words" list, thereby directly preventing the model from generating that action in subsequent attempts.
Reflexion adopts the same hyperparameter settings as LLM-Planner, as shown in Table 4. By default, we use Qwen2.5-0.5B for both VirtualHome and ALFWorld.

\noindent\textbf{LongMem.} LongMem is a composite model consisting of a frozen LM backbone and a trainable memory retrieval network, Residual Sidenet. Once trained, it automatically incorporates previous information via the key-value cache of earlier chunks into the current generation, enabling LM to effectively memorize and model long-term dependencies.

To adapt LongMem for benchmarks such as PDDLGym, VirtualHome, and ALFWorld, we apply several modifications and hyperparameter adjustments based on the official implementation~\footnote{https://github.com/Victorwz/LongMem}. The memory module is configured to store key-value pairs for up to 400 problem instances, with each input sequence stored in 64-token chunks. During retrieval, the model selects the top 6 most relevant chunks based on similarity to the current query. By default, we use LLaMA3.2-1B for PDDLGym and Qwen2.5-0.5B for VirtualHome and ALFWorld.

\noindent\textbf{LMM.} LMM augments a conventional decoder-only transformer with an auxiliary memory flow, enabling it to maintain long-term dependencies throughout generation. A dedicated memory module allows for preserving important information by dynamically interacting with input embeddings, while the memory itself continuously updating through a gating mechanism.

Based on the official implementation~\footnote{https://github.com/convergence-ai/lm2}, we make slight adaptations to use LMM as an LM-based agent. The memory capacity and slot size are set to match the hidden size of the backbone LM. Additionally, both the number of attention heads and the hidden size used in the memory-augmented attention mechanism follow the same configuration as the backbone model. By default, we use LLaMA3.2-1B for PDDLGym and Qwen2.5-0.5B for VirtualHome and ALFWorld.

\noindent\textbf{Optimus-2.} Optimus-2 is an embodied agent model designed to tackle various open-world tasks, with the focus on Minecraft. To effectively model the complex relationships among observations, actions, and language, Optimus-2 proposes action-guided behavior encoder, which integrates these three elements along with behavior history memory to produce behavior tokens.

Optimus-2 is used as a closed-loop task planning baseline for VirtualHome and ALFWorld. Since the official code is not available on the GitHub repository~\footnote{https://github.com/JiuTian-VL/Optimus-2}, we implemented it based on the descriptions provided in the original paper. We use Qwen2.5-0.5B as the backbone LM. The memory module is configured with a maximum history length of 1,000, and both the attention dimensionality and number of attention heads in the memory mechanism are set to match those of the backbone LM.

\noindent\textbf{DT-Mem.} DT-Mem addresses the forgetting problem in Decision Transformers, which arises from storing task-specific skills in implicit memory. Inspired by human working memory, it introduces a distributed memory architecture that allows for explicit storage and retrieval of multiple skills, effectively mitigating forgetting.

For the VirtualHome and ALFWorld baselines, we employ Qwen2.5-0.5B as the backbone LM, providing it with a concatenation of the current observation, the goal description, and domain knowledge. 
Only the memory module is borrowed from the public DT-Mem implementation~\footnote{https://github.com/luciferkonn/DT\_Mem}, configured with 128 memory slots whose dimensionality equals the model's hidden size.

\noindent\textbf{BUTLER.} BUTLER is a LM-based agent that generates high-level textual actions based on current observations and a given goal. Built upon a pre-trained LM, it is fine-tuned with a small set of expert demonstrations to better adapt to the target environment. BUTLER serves as the baseline for other models that also involve fine-tuning on demonstrations.

We adopt a LoRA-based fine-tuning approach~\cite{hu2022lora} for training. For the PDDLGym baseline, we use LLaMA3.2-1B as the backbone LM.

\noindent\textbf{BoT.} Buffer of Thoughts (BoT) utilizes thought templates, which are guidelines for high-level reasoning, and uses them as scaffold for reasoning. This improves both the efficiency and accuracy of the LM's reasoning process.

BoT begins by distilling key information from the problem into a well-structured list. It then retrieves the most relevant thought template from the meta-buffer and instantiates it with the key information to perform reasoning. Once reasoning is complete, the instantiated reasoning is distilled back into a new thought template and added to the meta-buffer.
For implementation, we refer to official code repository~\footnote{https://github.com/YangLing0818/buffer-of-thought-llm}. We use GPT-4.1, LLaMa3.1-8B, and LLaMa3.1-70B as base LMs. For problem distillation, however, we use GPT-4.1 across all models to ensure better performance. The hyperparameter settings for BoT are provided in Table~\ref{app:hyper:bot}.

\begin{table}[htbp]   
        \centering
        \caption{Hyperparamer settings for BoT}
        \label{app:hyper:bot}
        \begin{tabular}{l c}
        \toprule
        \textbf{Hyperparameters} & \textbf{Value} \\
        \midrule
        Retrieval Threshold    &  0.6 \\
        Maximum new tokens & 3,000 \\
        \bottomrule
    \end{tabular}
\end{table}

\noindent\textbf{LRM.} LRM is a class of LMs in which the ability to generate reasoning chains is internalized within the model parameters, enabling more adaptive and diverse reasoning than the in-context learning approach such as CoT.

These models are often trained via reinforcement learning, as in DeepSeek-R1, and can also serve as teacher models for training smaller variants through distillation.
In our work, we use two distilled models—DeepSeek-R1-Distill-LLaMa-8B and 70B—as well as one proprietary model, o3-mini, for proceduralization baselines. The hyperparameter settings for the distilled models are provided in Table~\ref{app:hyper:lrm}, while default arguments are used for o3-mini.

\color{black}

\begin{table}[htbp]   
        \centering
        \caption{Hyperparamer settings for LRM}
        \label{app:hyper:lrm}
        \begin{tabular}{l c}
        \toprule
        \textbf{Hyperparameters} & \textbf{Value} \\
        \midrule
        Temperature    &  0.6 \\
        Maximum new tokens & 3,000 \\
        \bottomrule
    \end{tabular}
\end{table}

\noindent\textbf{PlaSma.} PlaSma distills procedural knowledge from a large LM into a small LM by training it to generate both standard, and counterfactual plans under given constraints. To further enhance planning quality, a step-wise verifier is utilized during decoding process, guiding the small LM to produce more coherent plans.

As a proceduralization baseline, we employ GPT-4o~\cite{gpt4o} as the teacher LLM to synthesize both positive augmented problem instances and counterfactual variants. We fine-tune a RoBERTa-Large verifier~\cite{liu2019roberta} on the synthesized samples and embed it in a verifier-guided, step-wise beam-search decoder implemented with the public PlaSma code~\footnote{https://github.com/allenai/PlaSma}. As backbone models, we evaluate LLaMa3.2-1B, LLaMa3.2-3B, and LLaMa3.1-8B.

\subsubsection{Metrics} 
We evaluate performance using four standard metrics, following~\cite{llmagent:zsp, ALFRED20, anderson2018evaluation}.
\begin{itemize}[leftmargin=*, itemsep=0pt, topsep=0pt]
    \item \textbf{Cumulative Task Success Rate (CSR)} measures the proportion of tasks in which all required sub-goals are successfully completed, indicating overall task-level performance.
    \item \textbf{Cumulative Goal-Conditioned Success Rate (CGC)} computes the fraction of individual sub-goals achieved across all tasks, capturing the agent's partial progress even when full task completion is not attained.
    \item \textbf{Executability (Exe)} evaluates whether each action selected by the agent is executable within the environment, reflecting the syntactic and semantic correctness of generated actions.
    \item \textbf{Success rate weighted by Path Length (SPL)} accounts for both task success and the efficiency of the action sequence, rewarding shorter and more optimal plans that still achieve the goal.
\end{itemize}

\newpage
\subsection{Additional Experimental Results}

\begin{table*}[t]
\caption{
Detailed performance on open-loop continual embodied tasks across each \text{PDDLGym} domain.
These results correspond to Table~1 in the main paper (Section~4.2).
}
\label{app:tab:main:open}
% \vspace{-5pt}
\begin{center}
\begin{sc}
\begin{adjustbox}{width=0.98\linewidth}
\begin{tabular}{l c cccc cccc}
    \toprule
    \multirow{2}{*}{Method}
    & \multirow{2}{*}{Params}
    & \multicolumn{4}{c}{Train} & \multicolumn{4}{c}{Test} \\ %& \multicolumn{3}{c}{Glibrearrangement}  \\ 
    \cmidrule(rl){3-6} \cmidrule(rl){7-10} %\cmidrule(rl){9-11}
    & & CSR ($\uparrow$) & CGC ($\uparrow$) & Exe ($\uparrow$) & SPL ($\uparrow$) & CSR ($\uparrow$) & CGC ($\uparrow$) & Exe ($\uparrow$) & SPL ($\uparrow$) \\
    \midrule
    \multicolumn{7}{l}{\textbf{Domain}: Minecraft} & \\
    \midrule
    \rowcolor[HTML]{FFFFFF} ZSP & 1.7B (0.0\%)
    & 76.4$\pm$5.5 & 81.7$\pm$4.8 & 100.0$\pm$0.0 & 0.8$\pm$0.1
    & 16.4$\pm$1.5 & 35.9$\pm$2.0 & 100.0$\pm$0.1 & 0.1$\pm$0.0 \\
    
    \rowcolor[HTML]{FFFFFF} RAP & 1.7B (0.0\%)
    & 76.4$\pm$5.5 & 81.7$\pm$4.8 & 100.0$\pm$0.0 & 0.8$\pm$0.1
    & 16.8$\pm$0.9 & 18.0$\pm$1.9 & 100.0$\pm$0.1 & 0.1$\pm$0.0 \\
    
    % \midrule
    \rowcolor[HTML]{E9ECEF} CoT & 1.7B (0.0\%)
    & 83.3$\pm$6.7 & 83.3$\pm$9.3 & 100.0$\pm$0.0 & 0.8$\pm$0.1
    & 17.0$\pm$0.5 & 25.7$\pm$1.8 & 100.0$\pm$0.0 & 0.1$\pm$0.0 \\

    \rowcolor[HTML]{E9ECEF} ToT & 1.7B (0.0\%)
    & 85.1$\pm$4.2 & 85.7$\pm$3.7 & 100.0$\pm$0.0 & 0.9$\pm$0.0
    & 18.7$\pm$1.0 & 32.2$\pm$2.6 & 100.0$\pm$0.0 & 0.1$\pm$0.0 \\

    \rowcolor[HTML]{E9ECEF} GoT & 1.7B (0.0\%)
    & 89.1$\pm$5.1 & 94.6$\pm$7.7 & 100.0$\pm$0.0 & 0.9$\pm$0.1
    & 18.9$\pm$0.5 & 25.9$\pm$1.2 & 100.0$\pm$0.0 & 0.1$\pm$0.0 \\
    
    % \midrule
    \rowcolor[HTML]{FFFFFF} BUTLER & 1.2B (0.6\%)
    & 100.0$\pm$0.0 & 100.0$\pm$0.0 & 100.0$\pm$0.0 & 1.0$\pm$0.0
    & 51.4$\pm$1.9 & 56.7$\pm$2.6 & 99.7$\pm$0.3 & 0.4$\pm$0.0 \\

    \rowcolor[HTML]{FFFFFF} LongMem & 1.6B (24.6\%)
    & 100.0$\pm$0.0 & 100.0$\pm$0.0 & 100.0$\pm$0.0 & 1.0$\pm$0.0
    & 53.3$\pm$3.7 & 56.3$\pm$4.1 & 99.8$\pm$0.1 & 0.5$\pm$0.0 \\

    \rowcolor[HTML]{FFFFFF} LMM & 1.3B (6.3\%)
    & 100.0$\pm$0.0 & 100.0$\pm$0.0 & 100.0$\pm$0.0 & 1.0$\pm$0.0
    & 47.9$\pm$8.4 & 56.5$\pm$4.6 & 99.9$\pm$0.1 & 0.4$\pm$0.0 \\

    \rowcolor[HTML]{E9ECEF} $\ourmodel$ & 1.3B (6.3\%)
    & 100.0$\pm$0.0
    & 100.0$\pm$0.0
    & 100.0$\pm$0.0
    & 1.0$\pm$0.0
    & \textbf{65.2}$\pm$1.4
    & \textbf{68.9}$\pm$2.4
    & 100.0$\pm$0.1
    & \textbf{0.6}$\pm$ 0.0\\
    \midrule
    \multicolumn{7}{l}{\textbf{Domain}: Rearrangement} & \\
    \midrule
    \rowcolor[HTML]{FFFFFF} ZSP & 1.7B (0.0\%)
    & 94.2$\pm$3.8 & 97.6$\pm$1.2 & 100.0$\pm$0.0 & 0.9$\pm$0.0
    & 15.3$\pm$2.1 & 22.4$\pm$3.4 & 100.0$\pm$0.0 & 0.1$\pm$0.0 \\
    
    \rowcolor[HTML]{FFFFFF} RAP & 1.7B (0.0\%)
    & 94.2$\pm$3.8 & 97.6$\pm$1.2 & 100.0$\pm$0.0 & 0.9$\pm$0.0
    & 18.3$\pm$1.7 & 29.0$\pm$2.6 & 100.0$\pm$0.0 & 0.1$\pm$0.0 \\
    
    % \midrule
    \rowcolor[HTML]{E9ECEF} CoT & 1.7B (0.0\%)
    & 95.0$\pm$4.5 & 98.5$\pm$2.5 & 100.0$\pm$0.0 & 1.0$\pm$0.0
    & 19.2$\pm$0.9 & 29.9$\pm$1.5 & 100.0$\pm$0.0 & 0.1$\pm$0.0 \\

    \rowcolor[HTML]{E9ECEF} ToT & 1.7B (0.0\%)
    & 95.0$\pm$6.3 & 99.5$\pm$1.2 & 100.0$\pm$0.0 & 1.0$\pm$0.1
    & 20.2$\pm$1.2 & 33.9$\pm$1.4 & 100.0$\pm$0.0 & 0.2$\pm$0.0 \\

    \rowcolor[HTML]{E9ECEF} GoT & 1.7B (0.0\%)
    & 95.8$\pm$4.9 & 100.0$\pm$0.0 & 100.0$\pm$0.0 & 1.0$\pm$0.0
    & 23.0$\pm$1.5 & 37.1$\pm$1.4 & 100.0$\pm$0.0 & 0.2$\pm$0.0 \\
    
    % \midrule
    \rowcolor[HTML]{FFFFFF} BUTLER & 1.2B (0.6\%)
    & 100.0$\pm$0.0 & 100.0$\pm$0.0 & 100.0$\pm$0.0 & 1.0$\pm$0.0
    & 56.5$\pm$2.0 & 69.8$\pm$2.9 & 100.0$\pm$0.0 & 0.5$\pm$0.0 \\

    \rowcolor[HTML]{FFFFFF} LongMem & 1.6B (24.6\%)
    & 100.0$\pm$0.0 & 100.0$\pm$0.0 & 100.0$\pm$0.0 & 1.0$\pm$0.0
    & 58.2$\pm$1.9 & 69.9$\pm$1.4 & 100.0$\pm$0.0 & 0.5$\pm$0.0 \\

    \rowcolor[HTML]{FFFFFF} LMM & 1.3B (6.3\%)
    & 100.0$\pm$0.0 & 100.0$\pm$0.0 & 100.0$\pm$0.0 & 1.0$\pm$0.0
    & 57.4$\pm$4.0 & 69.6$\pm$8.9 & 100.0$\pm$0.0 & 0.5$\pm$0.1 \\

    \rowcolor[HTML]{E9ECEF} $\ourmodel$ & 1.3B (6.3\%)
    & 100.0$\pm$0.0
    & 100.0$\pm$0.0
    & 100.0$\pm$0.0
    & 1.0$\pm$0.0
    & \textbf{73.5}$\pm$3.0
    & \textbf{80.8}$\pm$1.0
    & 100.0$\pm$0.0
    & \textbf{0.7}$\pm$ 0.0\\
    \midrule
    \multicolumn{7}{l}{\textbf{Domain}: Glibrearrangement} & \\
    \midrule
    \rowcolor[HTML]{FFFFFF} ZSP & 1.7B (0.0\%)
    & 68.0$\pm$2.9 & 75.1$\pm$2.9 & 100.0$\pm$0.0 & 0.7$\pm$0.0
    & 14.0$\pm$1.4 & 27.9$\pm$7.3 & 100.0$\pm$0.0 & 0.1$\pm$0.0 \\
    
    \rowcolor[HTML]{FFFFFF} RAP & 1.7B (0.0\%)
    & 77.8$\pm$2.9 & 89.2$\pm$3.8 & 100.0$\pm$0.0 & 0.8$\pm$0.0
    & 16.8$\pm$1.6 & 21.0$\pm$2.6 & 100.0$\pm$0.0 & 0.1$\pm$0.0 \\
    
    % \midrule
    \rowcolor[HTML]{E9ECEF} CoT & 1.7B (0.0\%)
    & 80.9$\pm$5.0 & 89.5$\pm$3.9 & 100.0$\pm$0.0 & 0.8$\pm$0.1
    & 17.8$\pm$1.6 & 21.4$\pm$1.6 & 100.0$\pm$0.0 & 0.1$\pm$0.0 \\

    \rowcolor[HTML]{E9ECEF} ToT & 1.7B (0.0\%)
    & 93.9$\pm$2.6 & 98.8$\pm$1.5 & 100.0$\pm$0.0 & 0.9$\pm$0.0
    & 19.7$\pm$0.5 & 28.2$\pm$1.3 & 100.0$\pm$0.0 & 0.1$\pm$0.0 \\

    \rowcolor[HTML]{E9ECEF} GoT & 1.7B (0.0\%)
    & 95.1$\pm$1.5 & 95.1$\pm$2.7 & 100.0$\pm$0.0 & 1.0$\pm$0.0
    & 20.5$\pm$1.4 & 28.4$\pm$1.9 & 100.0$\pm$0.0 & 0.2$\pm$0.0 \\
    
    % \midrule
    \rowcolor[HTML]{FFFFFF} BUTLER & 1.2B (0.6\%)
    & 99.2$\pm$1.3 & 99.7$\pm$0.5 & 100.0$\pm$0.0 & 1.0$\pm$0.0
    & 46.5$\pm$2.9 & 57.2$\pm$3.2 & 100.0$\pm$0.0 & 0.3$\pm$0.0 \\

    \rowcolor[HTML]{FFFFFF} LongMem & 1.6B (24.6\%)
    & 99.6$\pm$1.0 & 99.8$\pm$0.4 & 100.0$\pm$0.0 & 1.0$\pm$0.0
    & 47.7$\pm$5.3 & 53.6$\pm$3.4 & 100.0$\pm$0.0 & 0.3$\pm$0.0 \\

    \rowcolor[HTML]{FFFFFF} LMM & 1.3B (6.3\%)
    & 99.6$\pm$1.0 & 99.8$\pm$0.4 & 100.0$\pm$0.0 & 1.0$\pm$0.0
    & 48.3$\pm$1.9 & 60.3$\pm$0.0 & 100.0$\pm$0.0 & 0.3$\pm$0.0 \\

    \rowcolor[HTML]{E9ECEF} $\ourmodel$ & 1.3B (6.3\%)
    & \textbf{100.0}$\pm$0.0
    & \textbf{100.0}$\pm$0.0
    & 100.0$\pm$0.0
    & 1.0$\pm$0.0
    & \textbf{58.7}$\pm$1.8
    & \textbf{72.4}$\pm$1.2
    & 100.0$\pm$0.0
    & \textbf{0.5}$\pm$0.0 \\
    \bottomrule
\end{tabular}
\end{adjustbox}
\end{sc}
\end{center}
% \vspace{-20pt}
\end{table*}

\subsubsection{Open-loop Continual Embodied Tasks}
To evaluate the generalization performance of $\ourmodel$ on open-loop continual task planning, we conduct experiments in Table~\ref{app:tab:main:open} using multiple domains from PDDLGym.
In this setting, the agent generates a complete action sequence without intermediate observations and receives only binary task feedback (success or failure) before proceeding to the next task.
$\ourmodel$ consistently outperforms the strongest baseline in each domain, LongMem in Minecraft and Rearrangement, and LMM in Glibrearrangement. It achieves average improvements of 12.5\% in \text{CSR}, 11.9\% in \text{CGC}, and 0.16 in \text{SPL}, demonstrating superior structured and adaptive reasoning capabilities.
The single-step planning baselines such as \text{ZSP} and \text{RAP}, which rely on in-context retrieval-augmented generation~\cite{icrag}, show limited reasoning capacity on unseen tasks. The agentic workflow baselines such as \text{CoT}, \text{ToT}, and \text{GoT}, which use reasoning-guidance prompts crafted from the train set~\cite{llmagent:deder}, perform slightly better, but remain far from achieving reliable task success. The memory-augmented LMs such as \text{LongMem} and \text{LMM} outperform the fine-tuning baseline \text{BUTLER}, yet still show 13.2\% lower \text{CSR} on average compared to $\ourmodel$.

\begin{table*}[t]
\caption{
Detailed performance on closed-loop continual embodied tasks in \text{VirtualHome} and \text{ALFWorld}. These results correspond to Table 2 in the main paper (Section 4.2).
}
\label{app:tab:main:closed}
% \vspace{-6pt}
\begin{center}
\begin{sc}
\begin{subtable}[Performance across different evaluation categories in \text{VirtualHome}]{
\label{app:tab:main:closed:vh}
\adjustbox{max width=0.98\linewidth}{
    \begin{tabular}{l ccc ccc ccc}
    \toprule
    \multirow{2}{*}{Method}
    & \multicolumn{3}{c}{Train} & \multicolumn{3}{c}{Seen} & \multicolumn{3}{c}{Unseen} \\
     \cmidrule(rl){2-4} \cmidrule(rl){5-7} \cmidrule(rl){8-10}
     & CSR ($\uparrow$) & CGC ($\uparrow$) & SPL ($\uparrow$)  & CSR ($\uparrow$)  & CGC ($\uparrow$) & SPL ($\uparrow$)  & CSR ($\uparrow$) & CGC ($\uparrow$) & SPL ($\uparrow$) \\
    \midrule
    \multicolumn{7}{l}{\textbf{Configuration}: Behavior-Incremental} & \\
    \midrule
    \rowcolor[HTML]{FFFFFF} LLM-planner
    & 61.0$\pm$3.5 & 66.4$\pm$2.9 & 0.4$\pm$0.0
    & 45.5$\pm$1.9 & 47.4$\pm$2.0 & 0.3$\pm$0.0
    & 28.8$\pm$2.2 & 30.0$\pm$2.1 & 0.2$\pm$0.0  \\
    \rowcolor[HTML]{E9ECEF} ReAct
    & 63.6$\pm$1.1 & 70.0$\pm$0.6 & 0.4$\pm$0.0
    & 54.7$\pm$1.3 & 56.4$\pm$0.8 & 0.4$\pm$0.0
    & 32.7$\pm$3.5 & 34.3$\pm$3.6 & 0.3$\pm$0.0  \\
    \rowcolor[HTML]{E9ECEF} Reflexion
    & 60.8$\pm$5.9 & 68.8$\pm$5.5 & 0.4$\pm$0.0
    & 57.2$\pm$3.8 & 61.6$\pm$5.4 & 0.4$\pm$0.0
    & 33.7$\pm$2.5 & 35.3$\pm$2.1 & 0.3$\pm$0.0  \\
    \rowcolor[HTML]{FFFFFF} LongMem
    & 80.5$\pm$2.9 & 86.2$\pm$2.5 & 0.8$\pm$0.0
    & 63.3$\pm$5.6 & 68.3$\pm$6.3 & 0.6$\pm$0.1
    & 45.7$\pm$7.3 & 52.2$\pm$8.9 & 0.4$\pm$0.1  \\
    \rowcolor[HTML]{FFFFFF} LMM
    & 80.2$\pm$4.2 & 85.2$\pm$2.9 & 0.8$\pm$0.0
    & 53.6$\pm$5.9 & 57.6$\pm$5.2 & 0.5$\pm$0.1
    & 38.8$\pm$4.4 & 43.3$\pm$4.4 & 0.3$\pm$0.0  \\
    \rowcolor[HTML]{FFFFFF} DT-Mem
    & 77.3$\pm$3.8 & 80.8$\pm$4.8 & 0.7$\pm$0.0
    & 69.3$\pm$5.7 & 71.9$\pm$5.9 & 0.7$\pm$0.1
    & 48.7$\pm$5.9 & 52.6$\pm$6.0 & 0.5$\pm$0.1  \\
    \rowcolor[HTML]{FFFFFF} Optimus-2
    & 79.3$\pm$5.4 & 83.9$\pm$4.0 & 0.8$\pm$0.1
    & 70.4$\pm$4.4 & 74.0$\pm$3.4 & 0.7$\pm$0.1
    & 44.0$\pm$6.1 & 50.9$\pm$7.3 & 0.4$\pm$0.1  \\
    \rowcolor[HTML]{E9ECEF} $\ourmodel$
    & \textbf{89.8}$\pm$1.9 & \textbf{92.3}$\pm$1.1 & \textbf{0.9}$\pm$0.0
    & \textbf{78.9}$\pm$4.5 & \textbf{81.7}$\pm$2.5 & \textbf{0.8}$\pm$0.0
    & \textbf{61.1}$\pm$2.2 & \textbf{69.3}$\pm$2.6 & \textbf{0.6}$\pm$0.0  \\
    \midrule
    \multicolumn{7}{l}{\textbf{Configuration}: Environment-Incremental} \\
    \midrule
    \rowcolor[HTML]{FFFFFF} LLM-planner
    & 61.7$\pm$7.5 & 66.5$\pm$6.8 & 0.4$\pm$0.0
    & 45.5$\pm$1.9 & 47.7$\pm$2.6 & 0.3$\pm$0.0
    & 32.7$\pm$2.7 & 33.7$\pm$2.9 & 0.2$\pm$0.0  \\
    \rowcolor[HTML]{E9ECEF} ReAct
    & 64.3$\pm$3.4 & 69.9$\pm$2.5 & 0.4$\pm$0.0
    & 52.2$\pm$2.7 & 55.6$\pm$2.5 & 0.4$\pm$0.0
    & 35.1$\pm$1.0 & 37.1$\pm$1.6 & 0.3$\pm$0.0  \\
    \rowcolor[HTML]{E9ECEF} Reflexion
    & 66.2$\pm$4.4 & 72.1$\pm$3.6 & 0.5$\pm$0.0
    & 59.6$\pm$4.1 & 62.4$\pm$3.3 & 0.5$\pm$0.0
    & 34.1$\pm$1.8 & 35.5$\pm$2.4 & 0.3$\pm$0.0  \\
    \rowcolor[HTML]{FFFFFF} LongMem
    & 80.2$\pm$3.2 & 86.2$\pm$2.4 & 0.8$\pm$0.0
    & 68.9$\pm$5.9 & 72.9$\pm$6.3 & 0.7$\pm$0.1
    & 45.2$\pm$3.3 & 51.8$\pm$4.0 & 0.4$\pm$0.0  \\
    \rowcolor[HTML]{FFFFFF} LMM
    & 80.2$\pm$4.2 & 85.2$\pm$2.9 & 0.8$\pm$0.0
    & 53.6$\pm$5.9 & 57.7$\pm$5.2 & 0.5$\pm$0.1
    & 38.8$\pm$4.4 & 43.3$\pm$4.4 & 0.3$\pm$0.0  \\
    \rowcolor[HTML]{FFFFFF} DT-Mem
    & 77.5$\pm$4.7 & 81.0$\pm$6.1 & 0.7$\pm$0.0
    & 70.1$\pm$6.3 & 72.7$\pm$7.4 & 0.7$\pm$0.1
    & 51.6$\pm$6.0 & 56.6$\pm$6.9 & 0.5$\pm$0.1  \\
    \rowcolor[HTML]{FFFFFF} Optimus-2
    & 79.3$\pm$5.4 & 83.8$\pm$4.0 & 0.8$\pm$0.1
    & 70.4$\pm$4.4 & 74.0$\pm$3.4 & 0.7$\pm$0.0
    & 44.0$\pm$6.1 & 50.9$\pm$7.3 & 0.4$\pm$0.1  \\
    \rowcolor[HTML]{E9ECEF} $\ourmodel$
    & \textbf{90.1}$\pm$1.0 & \textbf{92.7}$\pm$0.6 & \textbf{0.9}$\pm$0.0
    & \textbf{77.0}$\pm$2.5 & \textbf{80.7}$\pm$2.4 & \textbf{0.8}$\pm$0.0
    & \textbf{58.2}$\pm$3.0 & \textbf{66.1}$\pm$3.8 & \textbf{0.6}$\pm$0.0  \\
    \bottomrule
    \end{tabular}
}}
\end{subtable}
\vskip -0.0cm
\begin{subtable}[Performance across different evaluation categories in \text{ALFWorld}]{
\label{app:tab:main:closed:alfw}
\adjustbox{max width=0.98\linewidth}{
    \begin{tabular}{l ccc ccc ccc}
    \toprule
    \multirow{2}{*}{Method}
    & \multicolumn{3}{c}{Train} & \multicolumn{3}{c}{Seen} & \multicolumn{3}{c}{Unseen} \\
     \cmidrule(rl){2-4} \cmidrule(rl){5-7} \cmidrule(rl){8-10}
    & CSR ($\uparrow$) & CGC ($\uparrow$) & SPL ($\uparrow$)  & CSR ($\uparrow$)  & CGC ($\uparrow$) & SPL ($\uparrow$)  & CSR ($\uparrow$) & CGC ($\uparrow$) & SPL ($\uparrow$) \\
    \midrule
    \multicolumn{7}{l}{\textbf{Configuration}: Behavior-Incremental} & \\
    \midrule
    \rowcolor[HTML]{FFFFFF} LLM-planner
    & 52.4$\pm$2.6 & 68.1$\pm$2.0 & 0.5$\pm$0.0
    & 14.0$\pm$0.8 & 21.7$\pm$0.7 & 0.1$\pm$0.0
    & 3.3$\pm$0.4 & 11.7$\pm$0.5 & 0.0$\pm$0.0  \\
    \rowcolor[HTML]{E9ECEF} ReAct
    & 46.3$\pm$1.5 & 65.3$\pm$1.1 & 0.4$\pm$0.0
    & 12.8$\pm$0.5 & 21.4$\pm$0.6 & 0.1$\pm$0.0
    & 2.9$\pm$0.2 & 11.6$\pm$0.2 & 0.0$\pm$0.0  \\
    \rowcolor[HTML]{E9ECEF} Reflexion
    & 44.3$\pm$0.7 & 64.1$\pm$0.7 & 0.4$\pm$0.0
    & 13.0$\pm$0.5 & 21.6$\pm$0.8 & 0.1$\pm$0.0
    & 2.9$\pm$0.4 & 11.8$\pm$0.4 & 0.0$\pm$0.0  \\
    \rowcolor[HTML]{FFFFFF} LongMem
    & 50.1$\pm$3.1 & 58.6$\pm$2.4 & 0.5$\pm$0.0
    & 48.6$\pm$0.6 & 57.3$\pm$0.7 & 0.5$\pm$0.0
    & 45.2$\pm$0.8 & 54.5$\pm$0.8 & 0.5$\pm$0.0  \\
    \rowcolor[HTML]{FFFFFF} LMM
    & 63.8$\pm$1.5 & 71.5$\pm$1.4 & 0.6$\pm$0.0
    & 42.6$\pm$2.4 & 46.9$\pm$2.0 & 0.4$\pm$0.0
    & 38.7$\pm$2.1 & 45.2$\pm$2.5 & 0.4$\pm$0.0  \\
    \rowcolor[HTML]{FFFFFF} DT-Mem
    & 57.7$\pm$2.2 & 61.7$\pm$2.7 & 0.6$\pm$0.0
    & 41.3$\pm$2.9 & 48.0$\pm$3.3 & 0.4$\pm$0.0
    & 38.0$\pm$3.8 & 44.1$\pm$4.2 & 0.4$\pm$0.0  \\
    \rowcolor[HTML]{FFFFFF} Optimus-2
    & 59.5$\pm$1.5 & 67.4$\pm$1.3 & 0.6$\pm$0.0
    & 52.8$\pm$0.9 & 61.6$\pm$0.7 & 0.5$\pm$0.0
    & 49.1$\pm$0.7 & 58.7$\pm$0.6 & 0.5$\pm$0.0  \\
    \rowcolor[HTML]{E9ECEF} $\ourmodel$
    & \textbf{69.6}$\pm$2.7 & \textbf{76.2}$\pm$2.1 & \textbf{0.7}$\pm$0.0
    & \textbf{61.1}$\pm$1.1 & \textbf{68.6}$\pm$1.3 & \textbf{0.6}$\pm$0.0
    & \textbf{59.7}$\pm$1.4 & \textbf{67.9}$\pm$1.3 & \textbf{0.6}$\pm$0.0  \\
    \midrule
    \multicolumn{7}{l}{\textbf{Configuration}: Environment-Incremental} \\
    \midrule
    \rowcolor[HTML]{FFFFFF} LLM-planner
    & 39.1$\pm$2.0 & 52.4$\pm$0.9 & 0.3$\pm$0.0
    & 8.8$\pm$0.4 & 17.0$\pm$0.3 & 0.1$\pm$0.0
    & 2.7$\pm$0.1 & 10.4$\pm$0.5 & 0.0$\pm$0.0  \\
    \rowcolor[HTML]{E9ECEF} ReAct
    & 33.0$\pm$1.6 & 49.2$\pm$0.7 & 0.3$\pm$0.0
    & 9.2$\pm$0.5 & 18.0$\pm$0.5 & 0.1$\pm$0.0
    & 2.6$\pm$0.6 & 10.5$\pm$0.8 & 0.0$\pm$0.0  \\
    \rowcolor[HTML]{E9ECEF} Reflexion
    & 31.3$\pm$0.9 & 48.2$\pm$0.5 & 0.3$\pm$0.0
    & 8.8$\pm$0.4 & 17.3$\pm$0.1 & 0.1$\pm$0.0
    & 2.6$\pm$0.4 & 10.5$\pm$0.1 & 0.0$\pm$0.0  \\
    % 임시
    \rowcolor[HTML]{FFFFFF} LongMem
    & 40.6$\pm$1.1 & 51.3$\pm$0.9 & 0.4$\pm$0.0
    & 39.4$\pm$2.4 & 51.6$\pm$1.9 & 0.4$\pm$0.0
    & 32.6$\pm$1.4 & 45.7$\pm$0.7 & 0.3$\pm$0.0  \\
    \rowcolor[HTML]{FFFFFF} LMM
    & 48.0$\pm$0.4 & 52.8$\pm$1.4 & 0.5$\pm$0.0
    & 32.0$\pm$2.1 & 41.4$\pm$0.5 & 0.3$\pm$0.0
    & 29.0$\pm$1.4 & 39.3$\pm$2.8 & 0.3$\pm$0.0  \\
    \rowcolor[HTML]{FFFFFF} DT-Mem
    & 42.4$\pm$1.8 & 52.9$\pm$1.9 & 0.4$\pm$0.0
    & 38.7$\pm$11.7 & 47.8$\pm$14.7 & 0.4$\pm$0.1
    & 30.0$\pm$8.2 & 40.5$\pm$10.1 & 0.3$\pm$0.1  \\
    \rowcolor[HTML]{FFFFFF} Optimus-2
    & 44.6$\pm$0.7 & 56.9$\pm$0.5 & 0.4$\pm$0.0
    & 42.7$\pm$0.9 & 55.2$\pm$1.2 & 0.4$\pm$0.0
    & 33.1$\pm$0.2 & 46.9$\pm$0.7 & 0.3$\pm$0.0  \\
    \rowcolor[HTML]{E9ECEF} $\ourmodel$
    & \textbf{52.2}$\pm$0.9 & \textbf{62.4}$\pm$0.9 & 0.5$\pm$0.0
    & \textbf{51.1}$\pm$2.1 & \textbf{63.5}$\pm$1.8 & \textbf{0.5}$\pm$0.0
    & \textbf{41.1}$\pm$0.7 & \textbf{55.3}$\pm$1.0 & \textbf{0.4}$\pm$0.0  \\
    \bottomrule
    \end{tabular}
}}
\end{subtable}
\end{sc}
\end{center}
% \vspace{-20pt}
\end{table*}

\newpage
\subsubsection{Closed-loop Continual Embodied Tasks}
To further evaluate the generalization performance of $\ourmodel$ alongside its adaptability in dynamic settings, we conduct experiments under a closed-loop continual task planning setup across VirtualHome and ALFWorld. The test sets are explicitly divided into seen and unseen sets, enabling a detailed assessment of the agent’s structured and adaptive reasoning capabilities, as shown in Table~\ref{app:tab:main:closed}.
Unlike open-loop settings, the agent selects actions sequentially in response to intermediate observations.
In VirtualHome experiment (Table~\ref{app:tab:main:closed:vh}), $\ourmodel$ similarly outperforms the strongest baseline, \text{DT-Mem}.
It shows average improvements of 12.6\% in \text{CSR} and 11.6\% in \text{CGC} on the train set, 8.3\% and 8.9\% on the seen set, and 9.5\% and 13.1\% on the unseen set, respectively.
In ALFWorld experiment (Table~\ref{app:tab:main:closed:alfw}), $\ourmodel$ outperforms the strongest baseline, \text{Optimus-2}, across all incremental settings.
It achieves average gains of 8.9\% in \text{CSR} and 7.2\% in \text{CGC} on the train set, 8.4\% and 7.7\% on the seen set, and 9.3\% and 8.8\% on the unseen set, respectively.
Across both benchmarks, the performance gains on the unseen set are consistently greater than those on the seen sets.
Combined with an average improvement of 0.12 in \text{SPL}, these results indicate that $\ourmodel$ performs effective symbolic reasoning.
% 세부 분석
Specifically, both \text{LLM-Planner} and the agentic workflows such as \text{ReAct} and \text{Reflexion} show limited capabilities for symbolic reasoning across all evaluation scenarios.
While \text{Reflexion} leverages past experiences via feedback, it appears to lack the robust reasoning capabilities required in dynamic and complex tasks.
Memory-augmented models tailored for embodied agents, such as \text{DT-Mem} and \text{Optimus-2}, outperform general-purpose variants like \text{LongMem} and \text{LMM}.
However, $\ourmodel$ achieves higher performance, surpassing both \text{DT-Mem} and \text{Optimus-2} by an average of 11.2\% in \text{CSR} and 11.6\% in \text{CGC}, underscoring the effectiveness of neurosymbolic proceduralization.

\begin{table*}[ht]
\caption{
Detailed analysis of proceduralization, corresponding to Table~3 in the main paper (Section~4.3). \textsc{Latency} denotes the agent's plan generation time in seconds. \textsc{In/Out tokens} refer to the number of input and generated tokens, respectively. 
}
% \vspace{-5pt}
\begin{center}
\begin{sc}
\begin{adjustbox}{width=0.98\linewidth}
\label{app:tab:ana:proc}
\centering
\begin{tabular}{ll ccc ccc c}
    \toprule
    \multirow{2}{*}{Method}
    & \multirow{2}{*}{LM}
    & \multicolumn{3}{c}{Task performance} & \multicolumn{3}{c}{Reasoning load} & \multirow{2}{*}{FLOPs}  \\ 
    \cmidrule(rl){3-5} \cmidrule(rl){6-8}
    & & CSR ($\uparrow$) & CGC ($\uparrow$) & SPL ($\uparrow$) & Latency ($\downarrow$) & In tokens ($\downarrow$) & Out tokens ($\downarrow$) &  \\
    \midrule
    \multirow{3}{*}{BoT} & \cellcolor[HTML]{E9ECEF}LLaMa3.1-8B
    & 53.0$\pm$0.5 & 63.5$\pm$0.4 & 0.3$\pm$0.0 & 59.5$\pm$1.9 & 8007.9$\pm$103.9 & 1315.4$\pm$28.1 & 86.8$+\alpha$ TFLOPs \\
    & LLaMa3.1-70B      
    & 81.9$\pm$0.4 & 85.1$\pm$0.3 & 0.6$\pm$0.0 & 75.1$\pm$3.8 & 7651.0$\pm$127.7 & 794.1$\pm$33.4 & -  \\
    & GPT4.1
    & 92.1$\pm$0.3 & 93.6$\pm$0.2 & 0.7$\pm$0.0 & 22.2$\pm$2.8 & 7986.1$\pm$144.2 & 1202.2$\pm$197.2 & -  \\
    \midrule
    \multirow{3}{*}{LRM} & \cellcolor[HTML]{E9ECEF}DeepSeek-R1-8B
    & 11.5$\pm$0.3 & 15.6$\pm$0.3 & 0.1$\pm$0.0 & 111.0$\pm$3.3 & 3198.5$\pm$15.8 & 2187.6$\pm$69.0 & 55.4 TFLOPs  \\
    & DeepSeek-R1-70B
    & 26.5$\pm$0.4 & 27.5$\pm$0.4 & 0.2$\pm$0.0 & 209.4$\pm$9.2 & 3198.5$\pm$15.8 & 1679.3$\pm$87.5 & -  \\
    & o3-mini
    & 78.9$\pm$0.4 & 80.8$\pm$0.4 & 0.5$\pm$0.0 & 18.6$\pm$1.7 & 3214.6$\pm$15.7 & 2113.9$\pm$63.2 & -  \\
    \midrule
    \multirow{3}{*}{PlaSma} & LLaMa3.2-1B
    & 67.4$\pm$0.5 & 71.9$\pm$0.4 & 0.7$\pm$0.0    & 2.7$\pm$0.5 & 3221.8$\pm$45.3 & 32.7$\pm$4.6 & -  \\
    & LLaMa3.2-3B
    & 70.7$\pm$0.4 & 75.7$\pm$0.3 & 0.7$\pm$0.0 & 7.2$\pm$0.7 & 3247.7$\pm$17.2 & 29.5$\pm$5.5 & -  \\
    & \cellcolor[HTML]{E9ECEF}LLaMa3.1-8B
    & 80.5$\pm$0.5 & 89.2$\pm$2.3 & 0.8$\pm$0.0 & 18.4$\pm$5.8 & 3371.0$\pm$13.5 & 122.4$\pm$41.6 & 236.9 TFLOPs  \\
    \midrule
    \multirow{3}{*}{$\ourmodel$} & LLaMa3.2-1B
    & 73.2$\pm$0.4 & 76.0$\pm$0.4 & 0.7$\pm$0.0 & 1.2$\pm$0.3 & 3168.5$\pm$0.0 & 30.1$\pm$5.3 & -  \\
    & LLaMa3.2-3B
    & 83.6$\pm$2.0 & 88.8$\pm$2.0 & 0.8$\pm$0.0 & 3.5$\pm$0.3 & 3169.5$\pm$0.0 & 43.6$\pm$5.3 & -  \\
    & \cellcolor[HTML]{E9ECEF}LLaMa3.1-8B
    & 89.0$\pm$2.0 & 93.5$\pm$1.8 & 0.9$\pm$0.0 & 5.2$\pm$0.7 & 3155.5$\pm$0.0 & 41.9$\pm$6.0 & 98.3 TFLOPs  \\
    \bottomrule
\end{tabular}
\end{adjustbox}
\end{sc}
\end{center}
% \vspace{-15pt}
\end{table*}

\subsubsection{Analysis on Proceduralization}
Table~\ref{app:tab:ana:proc} presents a comparative analysis of our neurosymbolic proceduralization method to existing proceduralization methods, evaluated in terms of task performance and reasoning efficiency, with a particular focus on enabling timely reasoning through single-step inference.
To ensure a fair comparison, we additionally include a unified setting in which all methods are evaluated under identical inference conditions using the same LLaMA3.1-8B backbone, highlighted in gray background in the table.
% Under these conditions, $\ourmodel$ achieves the lowest average plan generation latency, the highest task success rate, and the minimal input and output token usage.
Under these conditions, $\ourmodel$ achieves the lowest average plan generation latency of 5.2 seconds, the highest task success rate of 89.0\%, and the fewest input and output tokens—averaging 3,155.5 and 41.9, respectively.
\text{BoT} and \text{LRM} exhibit latencies that are 54.3 and 105.8 seconds longer than $\ourmodel$, respectively, along with 6,125.9 and 2,188.7 more total tokens consumed, due to their reliance on multi-step reasoning.
\text{PlaSma}, which distills procedural knowledge from larger to smaller LMs, achieves competitive results with efficient inference, reaching a CSR of 80.5\% using an 8B LM. Yet, $\ourmodel$ outperforms it with a higher CSR of 83.6\% while operating with only a 3B LM.
While $\ourmodel$ incurs slightly higher FLOPs than simpler baselines such as LRM, this overhead stems from its memory-augmented module and contrastive planning mechanisms, which are designed to enhance robustness and correctness in reasoning. 
It is important to note that in our implementation, the problem distillation step of BoT is performed by GPT-4.1 regardless of the base LM. Consequently, this additional step is excluded from FLOPs measurement, and we denote it as ``$+\alpha$'' to reflect the unaccounted cost—potentially underestimating the actual computational burden of BoT. In contrast to prompt-based or distillation-based approaches, $\ourmodel$ composes and validates procedures during inference, trading off minimal compute overhead for significantly higher planning accuracy and consistency.
Despite this moderate FLOPs usage, $\ourmodel$ achieves the best trade-off across success rate, latency, and token efficiency. This demonstrates that structured reasoning can remain performant without relying on excessively large models or token-intensive decoding.

\begin{figure*}[ht]
% \vspace{-5pt}
\begin{center}
\centerline{\includegraphics[width=0.99\linewidth]{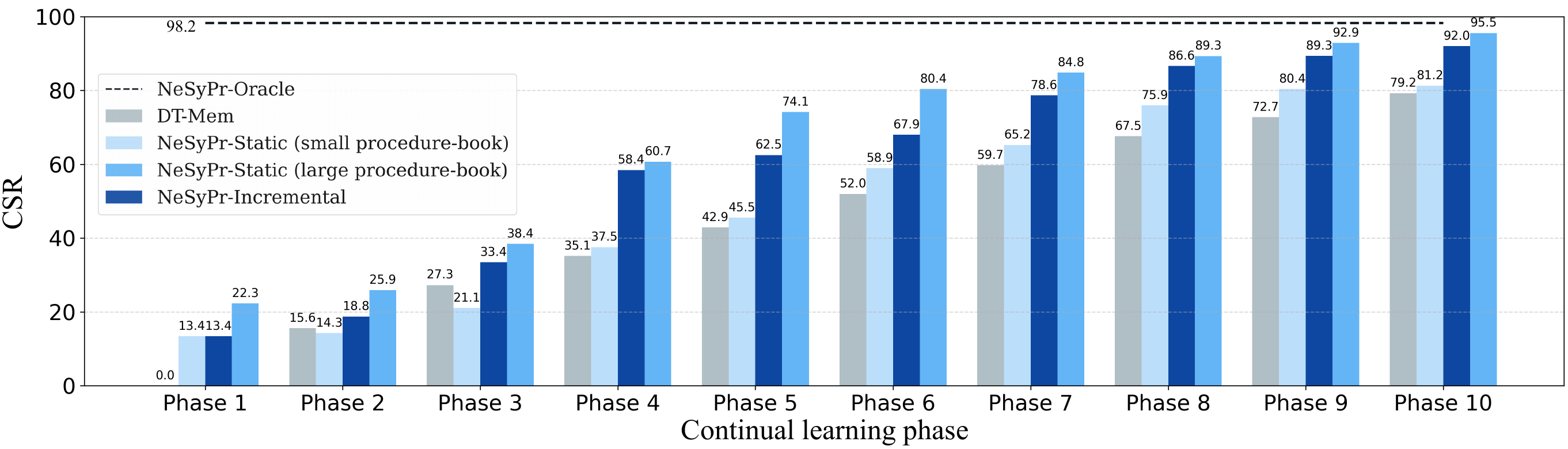}}
% \vspace{-5pt}
\caption{
Analysis on continual embodied task learning scenarios
}
\label{app:tab:ana:cl}
\end{center}
% \vspace{-15pt}
\end{figure*}

\subsubsection{Analysis on Continual Embodied Task Learning}
Figure~\ref{app:tab:ana:cl} shows how \text{CSR} evolves over 10 continual learning phases, where new tasks are introduced at each phase and evaluation is consistently performed on the full task set defined in Phase 10~\cite{nesyagent:nesyc, lee2024incremental, liu2023libero}.  
We test whether effective continual learning can be achieved by incrementally expanding the procedure-book as new procedural knowledge is acquired. This setting is denoted as $\ourmodel$-Incremental.
We compare four baselines. $\ourmodel$-Oracle represents the upper-bound performance, trained on the entire task set with a sufficiently large procedure-book in a single learning phase.  
DT-Mem is the memory-augmented LM baseline explained in Section~\ref{baselines}, adapted to the continual learning setting.  
$\ourmodel$-Static reuses a fixed-size procedure-book throughout training; we include both small and large variants to observe the effect of capacity.
$\ourmodel$-Incremental increases the procedure-book size at each phase while continuing neurosymbolic proceduralization on the new train data.
It shows steady improvement in \text{CSR} and reaches performance comparable to $\ourmodel$-Static with a large procedure-book.
This indicates that the agent can effectively accumulate procedural knowledge over time by gradually expanding the memory without needing to retrain from scratch.

\begin{table}[h]
\caption{Evaluation of inference efficiency and task performance of $\ourmodel$ across a range of devices, including an embedded device. Latency indicates the time taken for the agent to generate a plan, and memory usage captures the minimum and maximum values observed during continuous logging throughout inference. On GPU-based devices, this reflects GPU memory only, while on Jetson Orin, which lacks a discrete GPU, it reflects overall system memory usage.}
\label{app:tab:resource}
\begin{center}
\begin{adjustbox}{width=0.95\linewidth}
\begin{tabular}{l ccc cc ccc ccc ccc}
\toprule
\multirow{2}{*}{Device}
& \multicolumn{3}{c}{Latency (sec)} & \multicolumn{2}{c}{Memory Usage (GB)} & \multicolumn{3}{c}{Train} & \multicolumn{3}{c}{Seen} & \multicolumn{3}{c}{Unseen} \\
\cmidrule(rl){2-4} \cmidrule(rl){5-6} \cmidrule(rl){7-9} \cmidrule(rl){10-12} \cmidrule(rl){13-15}
& Min. & Max. & Avg.  & Min. & Max.  & SR & GC & SPL  & SR & GC & SPL  & SR & GC & SPL \\
\midrule
RTX A6000
& $0.5$ & $5.1$ & $1.5$ & $0.0$ & $1.9$ & $97.4$ & $97.7$ & $1.0$ & $88.4$ & $89.6$ & $0.9$ & $61.5$ & $67.9$ & $0.6$ \\

RTX 4090
& $0.5$ & $4.5$ & $1.0$ & $0.0$ & $1.9$ & $96.1$ & $98.7$ & $1.0$ & $88.4$ & $89.7$ & $0.9$ & $61.5$ & $68.5$ & $0.6$ \\

RTX 3090
& $0.5$ & $4.8$ & $1.2$ & $0.0$ & $1.9$ & $92.2$ & $94.6$ & $0.9$ & $86.6$ & $87.8$ & $0.9$ & $61.5$ & $68.5$ & $0.6$ \\

RTX 3050
& $0.6$ & $7.7$ & $1.6$ & $0.0$ & $1.9$ & $94.8$ & $96.9$ & $0.9$ & $84.8$ & $86.5$ & $0.8$ & $61.5$ & $67.9$ & $0.6$ \\

Jetson Orin
& 2.0 & 11.4 & $5.0$ & 4.3 & 8.6 & $93.5$ & $96.1$ & $0.9$ & $87.5$ & $88.7$ & $0.9$ & $59.6$ & $66.6$ & $0.6$ \\

\bottomrule
\end{tabular}
\end{adjustbox}
\end{center}
\vskip -0.1in
\end{table}

\subsubsection{Evaluation on Inference Efficiency Across Diverse Devices.}
Table~\ref{app:tab:resource} presents the inference efficiency and task performance of $\ourmodel$ across five different devices, ranging from high-end (RTX A6000, 4090) to off-the-shelf GPUs (RTX 3090, 3050) and embedded (Jetson Orin) devices.
Despite substantial differences in computational resources, $\ourmodel$ demonstrates stable performance across the train, seen, and unseen sets.
Across all tested devices, the task performance of $\ourmodel$ remains relatively stable, exhibiting less than a 5.5\% absolute variation in SR on the train set.
High-end GPUs such as the RTX A6000 and 4090 achieve peak SRs exceeding 96\%, while even lower-tier devices like the RTX 3090 maintain performance above 92\%, demonstrating consistent task execution across diverse hardware configurations.
Although the agent stores and reuses procedures during inference, it incurs minimal overhead in both latency and memory usage.  
On Jetson Orin, the average inference time is approximately 5 seconds—still reasonably fast—demonstrating the potential for deployment on embedded devices.
However, the considerable variance between minimum and maximum latency indicates that further optimization may be necessary to ensure more stable performance under highly resource-constrained conditions.

\newpage
\printbibliography
% \end{document}

\end{document}